\documentclass{article}
\pdfoutput=1
\usepackage{arxiv}
\usepackage[utf8]{inputenc}
\usepackage[T1]{fontenc}
\usepackage{hyperref}
\usepackage{url}
\usepackage{booktabs}
\usepackage{amsfonts}
\usepackage{nicefrac}
\usepackage{microtype}
\usepackage{lipsum}
\usepackage{graphicx}
\usepackage{natbib}
\usepackage{doi}
\usepackage{amsmath}
\usepackage{amssymb}
\usepackage{algorithm}
\usepackage{algpseudocode}
\usepackage{subcaption}
\usepackage{textcomp}
\usepackage{float}

\DeclareMathOperator{\tr}{tr}

\title{Calibrating constitutive models with full-field data via physics informed neural networks}

\author{Craig M.~Hamel\thanks{Corresponding Author: \texttt{chamel@sandia.gov}}\\
		Materials and Failure Modeling\\
		Sandia National Laboratories\\
		Albuquerque, NM 87123 \\
		\And
		Kevin N.~Long \\
		Materials and Failure Modeling\\
		Sandia National Laboratories\\
		Albuquerque, NM 87123 \\
		\And
		Sharlotte L.B.~Kramer \\
		Experimental Solid Mechanics\\
		Sandia National Laboratories\\
		Albuquerque, NM 87123 \\
}

\renewcommand{\shorttitle}{Calibrating constitutive models with full-field data via physics informed neural networks}

\hypersetup{
pdftitle={Calibrating constitutive models with full-field data via physics informed neural networks},
pdfauthor={Craig M.~Hamel, Kevin N.~Long, Sharlotte L.B.~Kramer},
pdfkeywords={Physics Informed Neural Networks, Continuum Mechanics, Constitutive Modeling, Inverse Modeling},
}

\graphicspath{{./figures/}}

\begin{document}
\maketitle

\begin{abstract}
The calibration of solid constitutive models with full-field experimental data is a long-standing challenge, especially in materials which undergo large deformation. In this paper, we propose a physics-informed deep-learning framework for the discovery of constitutive model parameterizations given full-field displacement data and global force-displacement data. Contrary to the majority of recent literature in this field, we work with the weak form of the governing equations rather than the strong form to impose physical constraints upon the neural network predictions. The approach presented in this paper is computationally efficient, suitable for irregular geometric domains, and readily ingests displacement data without the need for interpolation onto a computational grid. A selection of canonical hyperelastic materials models suitable for different material classes is considered including the Neo-Hookean, Gent, and Blatz-Ko constitutive models as exemplars for general hyperelastic behavior, polymer behavior with lock-up, and compressible foam behavior respectively. We demonstrate that physics informed machine learning is an enabling technology and may shift the paradigm of how full-field experimental data is utilized to calibrate constitutive models under finite deformations.
\end{abstract}


\section{Introduction} \label{sec:introduction}
The maturity of full-field measurement techniques, such as digital image correlation~\citet{Sutton2009} (DIC) and digital volume correlation~\citet{Buljac2018} (DVC), has enabled a paradigm shift in the way experimental mechanics can be conducted, especially for the purpose of constitutive model calibration. Although these methods cannot probe the local forces within mechanical specimens, the information associated with the local kinematics in conjunction with the global force response can offer modelers much richer data than in the past. Beyond experimental diagnostics, researchers have tried to exploit this wealth of new information over the past several decades in two major veins. These are finite element model updating (FEMU) techniques and the Virtual Fields Method (VFM)~\citet{avril_overview_2008}.

Using FEMU for constitutive model calibration consists of iteratively updating material properties using a representative finite element model of an experimental specimen. The strengths of FEMU is that it uses the mature technologies of FEM and optimization algorithms to arrive at material calibrations while minimizing the error between the measured full-field surface displacements and the global force. In this method either the experimentally measured displacement data needs to be interpolated onto the finite element mesh or the two need to be transformed onto a common basis for calculating the observed error~\citet{Salloum2020}. Since the FEM solve is typically decoupled from the optimization algorithm, researchers typically rely on either numerically approximated gradients in gradient-based optimization methods or gradient-free optimization techniques, such as genetic algorithms, which can be slow and prone to converging on local minima. 

VFM is an inverse modeling technique based on the Principle of Virtual Work, or the Principle of Virtual Power depending on the formulation. After first being developed for isotropic linear elasticity~\citet{pierron2012virtual}, the method has seen widespread application in various regimes of constitutive modeling such as anisotropic materials~\citet{Gre2003,AVRIL20102978}, composites~\citet{Pierron2007A,Pierron2007B}, and inelastic materials~\citet{Yoon2015}. The typical approach for VFM is to reconstruct the stress field (based on a prescribed constitutive model) from the measured full-field displacements and strains by applying the measured global force as a traction boundary condition. The material properties are then fit by optimizing a cost function that is based on the error between the internal and external work. However, this method is typically only utilized for specimens where certain assumptions can be made to reduce the dimensions of the problem since internal displacements are not generally measurable but are needed to calculate internal work. Typical assumptions include plane-stress for thin specimens or beam bending problems. With such assumptions, utilizing thick or fully three-dimensional specimens is difficult. 

Recently there has been an explosion of research in physics informed neural networks (PINNs) as a new approach for physics-based inverse modeling. Although the first example of using a neural network to solve differential equations in the literature dates back to the 1990s~\citet{Lagaris712178}, this idea has only recently been revisited~\citet{RAISSI2019686} due to advancements in GPU hardware, optimizers~\citet{kingma2017adam}, and libraries such as TensorFlow~\citet{tensorflow2015-whitepaper},  PyTorch~\citet{NEURIPS2019_9015}, and JAX~\citet{jax2018github}. According to the universal approximation theorem, any continuous function can be approximated with a sufficiently complex neural network. In the sense of solving forward problems, PINNs exploit this concept by approximately enforcing physical constraints, such as boundary conditions and conservation laws, on the output of a neural network. In essence, enforcing the neural network to approximate the solution of a partial differential equation (PDE). The trainable parameters of the PINN then act as the``free degrees of freedom'' in the solution process. This approach can also be extended to approximating model parameters~\citet{RAISSI2019686} (such as material properties), design parameters, or even mathematical operators by adding additional trainable parameters to the neural network and proper loss function definition. An overwhelming majority of the current literature is on collocation based methods, although there are a few examples using variational~\citet{2021Kharazmi} or finite volume approaches~\citet{PATEL2022110754}. 

Although PINNs have seen a great deal of success for problems in an Eulerian setting, i.e. fluid mechanics~\citet{Cai2022}, there has been limited work considering Lagrangian problems such as those typically seen in large-deformation solid-mechanics problems. One of the first works involving PINNs in solid mechanics was presented in~\citet{HAGHIGHAT2021113741} where simple 2D linear elastic static problems with known analytic solutions were considered for the purpose of discovering the elastic constants or creating surrogate models for fast predictions after training. This burdensome approach required a separate neural network for each displacement component, strain component, and stress component. Additional penalty terms were also necessary in the loss function for enforcing both essential and natural boundary conditions as well as enforcing the outputs of the stress neural networks to be compatible with the constitutive equation. Furthermore, the approach required data representative of the entire displacement, strain, and stress field making it an infeasible candidate for ingesting realistic experimental data. This approach to PINNs that has many penalty terms can prove to be problematic in practice due to no known optimal way to weight the different terms as discussed in~\citet{wang2020understanding,WANG2022110768,rohrhofer2021pareto}. There are also multiple works considering PINNs as a viable forward problem solution technique based on collocation methods~\citet{2021Abueidda}, and energy based variational methods~\citet{NGUYENTHANH2020103874,2022Fuhg}. 

However, the true power of PINNs are unlocked in solving inverse problems since the current training times for forward problems cannot compete with traditional approaches in scientific computing. Other than the simple cases considered in ~\citet{HAGHIGHAT2021113741}, there are only three publications known to the authors that consider solving inverse problems in solid mechanics with PINNs. The first is a design problem ~\citet{ZHANG2021100220} that used an energy-based variational approach to find optimal distributions of material properties for digital design in a 2D linear elastic framework. The second is a 2D inverse-modeling problem presented in~\citet{zhang2020physicsinformed} where an additional neural network was used to approximate the spatially dependent shear modulus for applications in biomedical imaging. The final example is a recent paper that uses PINNs in an inverse problem where the goal is to determine defect or inclusion geometries along with constituent material properties for simple 2D problems~\citet{2022_zhang}. This method was also collocation-based and thus suffered from the need to have many collocation points to accurately resolve the deformation gradients, which will lead to scaling issues when extended to 3D. These last two examples are the only examples of inverse modeling with PINNs for large-deformation solid mechanics known to the authors at the time of submission.

In this paper we propose a slightly different approach to PINNs that is comparable to VFM. Rather than using collocation points as inputs to the network, we use the nodal coordinates of a FEM mesh. The notion of a mesh is not known to the neural network but is used downstream to numerically approximate the integrals present in variational descriptions of solid mechanics. In this way we have reduced the number of necessary derivatives to take to enforce physics constraints on the loss function, and we can use the mature methods of how to perform Gauss quadrature on finite elements from FEM. We further reduce the burden of gradient representations on the neural network by calculating displacement gradients using standard FEM shape functions that allow us to use relatively coarse computational representations of the geometry. The variational setting also reduces the number of penalty terms in the loss function due to traction-free boundary conditions being satisfied by construction. We additionally use distance functions, similar to the approach used in~\citet{NGUYENTHANH2020103874,rao2021physics}, to satisfy displacement boundary conditions (BCs) by construction. In the case of forward problems, our method's loss function is only composed of physical quantities corresponding to a total global energy and the internal force vector residual. For inverse problems we simply use the same architecture for forward problems where additional parameters are added to the neural network optimizer to serve as the unknown material parameters and two penalty terms in the loss function that correspond to errors in (the experimentally input) surface full-field displacements and global reaction forces.

Our paper is outlined as follows. We present the general underlying theory for both the relevant equations of continuum mechanics and the neural network architectures that we will use in our PINN in section~\ref{sec:theory}. A detailed description of the numerical implementation for both forward and inverse problems is then given in section~\ref{sec:implementation}. Results are then shown in section~\ref{sec:results} for both forward problem boundary value problems (BVPs) as a form of validation and verification (V\&V) and inverse problems for some simple and unrealistic yet representative geometries of DIC specimens with heterogeneous strain fields. We close the paper with some concluding remarks in section~\ref{sec:conclusion}.

\section{Theory} \label{sec:theory}
In this section we introduce some preliminary notation and relationships that are standard continuum mechanics~\citet{holzapfel_nonlinear_2000,gurtin_mechanics_2010} and neural network material. In what follows in the next section of our neural network approximation of the problem, we will be working in the Lagrangian framework, and we will thus only list the relevant equations of motion in this setting for brevity. Relevant constitutive models that are utilized later on in the paper will also be presented and briefly reviewed. Finally we conclude this section with a brief review of multi-layer perceptron neural (MLP) networks for the uninitiated readers.
\subsection{Kinematics} \label{sec:kinematics}
We will denote the continuous region of space occupied by a material in the reference configuration as $\mathcal{B}_0$. The vectors corresponding to the coordinates of the material points that reside in this space we denote by $\mathbf{X}\in\mathcal{B}_0$. At an arbitrary time $t > 0$ we will use $\mathcal{B}_t$ to denote the continuous region of space that the material occupies, and $\mathbf{x}\in\mathcal{B}_t$ denotes the spatial coordinates. There exists a motion $\mathbf{\chi}$ that maps material points to spatial points and is defined by
\begin{equation}
	\mathbf{x} = \mathbf{\chi}\left(\mathbf{X}, t\right).
	\label{eq:motion}
\end{equation}
The deformation gradient is defined as
\begin{equation}
	\mathbf{F} = \frac{\partial \mathbf{x}}{\partial \mathbf{X}},
	\label{eq:deformation_gradient}
\end{equation}
and the Jacobian is simply the determinant of Equation~\ref{eq:deformation_gradient}, i.e.
\begin{equation}
	J = \det\mathbf{F}.
	\label{eq:jacobian}
\end{equation}
The right Cauchy-Green tensor is given by $\mathbf{C} = \mathbf{F}^T\mathbf{F}$ and is both. The three invariants of the right Cauchy-Green tensor are $I_1 = \tr{\mathbf{C}}$, $I_2 = \frac{1}{2}\left[\left(\tr\mathbf{C}\right)^2 - \tr\mathbf{C}^2\right]$, and $I_3 = \det\mathbf{C}$. The Green-Lagrange strain is defined in terms of the right Cauchy-Green tensor as
\begin{equation}
	\mathbf{E} = \frac{1}{2}\left(\mathbf{C} - \mathbf{I}\right),
\end{equation}
which we will use to present the underlying theory even though this quantity is never calculated in our PINN implementation.

In order to make boundary value problems well-posed, we must also impose appropriate boundary conditions upon the body. We only consider cases where surfaces on the boundary may only have prescribed displacements, \emph{or} prescribed tractions, not both. We denote the surface of the arbitrary body in the reference configuration by $\partial\mathcal{B}_0$, and we split the surface into two components. The first part of the surface is where displacements are prescribed and will be denoted by $\partial\mathcal{B}_0^{\mathbf{u}}$, and the remaining parts of the surface are where tractions are prescribed and will be denoted by $\partial\mathcal{B}_0^{\mathbf{t}}$. The combination of the two surfaces is the entire surface, i.e. $\partial\mathcal{B}_0 = \partial\mathcal{B}_0^{\mathbf{u}} \cup \partial\mathcal{B}_0^{\mathbf{t}},$
and their intersection is the empty set, $\partial\mathcal{B}_0^{\mathbf{u}}\cap\partial\mathcal{B}_0^{\mathbf{t}} = \varnothing$.


\subsection{Energy Functional}
The energy functional for a hyperelastic solid under static loading is given by
\begin{equation}
	\Pi = \int_{\mathcal{B}_0}\psi\left(\mathbf{E}\right)dv - \int_{\mathcal{B}_0}\mathbf{b}\cdot\mathbf{u}dv - \int_{\partial\mathcal{B}_0^{\mathbf{t}}}\tilde{\mathbf{t}}\cdot\mathbf{u}da,
	\label{eq:energy_functional}
\end{equation}
where $\mathcal{B}_0$ is an arbitrary volume, $\psi$ is the strain energy density (written here in terms of the Green-Lagrange strain for convenience), $dv$ is an infinitesimal volume element, $\mathbf{b}$ is the body force vector, and $\partial\mathcal{B}_{\tilde{\mathbf{t}}}$
is an arbitrary surface that the traction vector $\tilde{\mathbf{t}}$ acts on. In general for non-linear theories, solutions may be obtained that are non-unique, and only a stationary value of Equation~\ref{eq:energy_functional} may be obtained~\citet{wriggers_nonlinear_2008}. Finding a stationary value for Equation~\ref{eq:energy_functional} becomes a minimization problem of finding a displacement field $\mathbf{u}$ such that $\mathbf{u} = \tilde{\mathbf{u}}$ on $\partial\mathcal{B}_0^{\mathbf{u}}$ and $\Pi\left(\mathbf{u}\right)$ is minimum. This may be stated as 
\begin{equation}
	\min_{\mathbf{u}\in H^1\left(\mathcal{B}_0\right)} \Pi\left(\mathbf{u}\right),
	\label{eq:minimization_problem}
\end{equation}
where $H^1$ denotes the Sobolev space of degree one. This will be a key feature in our neural network formulation of the problem that will follow. The minimization problem given in equation~\ref{eq:minimization_problem} can be recast into $\delta\Pi = 0$, i.e.
\begin{equation}
	\delta\Pi = \int_{\mathcal{B}_0}\delta\psi\left(\mathbf{E}\right)dv - \int_{\mathcal{B}_0}\mathbf{b}\cdot\delta\mathbf{u}dv - \int_{\partial\mathcal{B}_0^{\mathbf{t}}}\tilde{\mathbf{t}}\cdot\delta\mathbf{u}da = 0,
	\label{eq:first_variation}
\end{equation}
where it has been assumed that the body force and traction vector are independent of the displacement $\mathbf{u}$. In equation~\ref{eq:first_variation}, $\delta\psi$ is given by 
\begin{equation}
	\delta\psi = \frac{\partial\psi\left(\mathbf{E}\right)}{\partial\mathbf{E}}:\delta\mathbf{E} = \mathbf{S}:\delta\mathbf{E},
	\label{eq:strain_energy_variation}
\end{equation} 
where $\mathbf{S} = \frac{\partial\psi\left(\mathbf{E}\right)}{\partial\mathbf{E}}$ is the definition of the second Piola-Kirchoff stress and $\delta\mathbf{E}$ is the first variation of the Green-Lagrange strain which is given by
\begin{equation}
	\delta\mathbf{E} = \frac{1}{2}\left[\left(\nabla_\mathbf{X}^T\delta\mathbf{u}\right)\mathbf{F} + \mathbf{F}^T\nabla_\mathbf{X}\delta\mathbf{u}\right].
	\label{eq:green_lagrange_strain_variation}
\end{equation}
Substituting equation~\ref{eq:strain_energy_variation} into~\ref{eq:first_variation} gives
\begin{equation}
	\delta\Pi = \int_{\mathcal{B}_0}\mathbf{S}:\delta\mathbf{E}dv - \int_{\mathcal{B}_0}\mathbf{b}\cdot\delta\mathbf{u}dv - \int_{\partial\mathcal{B}_0^{\mathbf{t}}}\tilde{\mathbf{t}}\cdot\delta\mathbf{u}da = 0,
	\label{eq:first_variation_with_pk2_stress}
\end{equation}
which is the \emph{Principle of Virtual Work} under static assumptions. By choosing to work with the second Piola-Kirchoff stress, $\mathbf{S}$, the balance of angular momentum is already satisfied due to the symmetry of the stress tensor, i.e.
\begin{equation}
	\mathbf{S} = \mathbf{S}^T.
	\label{eq:balance_of_angular_momentum}
\end{equation}
Now consider the term $\mathbf{S}:\delta\mathbf{E}$, which can be rewritten with the help of equation~\ref{eq:green_lagrange_strain_variation} as
\begin{equation}
	\mathbf{S}:\delta\mathbf{E} = \mathbf{F}\mathbf{S}:\delta\nabla_\mathbf{X}\mathbf{u}.
	\label{eq:pk1_stress_work_energy}
\end{equation}
Equation~\ref{eq:pk1_stress_work_energy} can be expanded into the following two terms:
\begin{equation}
	\mathbf{F}\mathbf{S}:\delta\nabla_\mathbf{X}\mathbf{u} = \nabla_\mathbf{X}\cdot\left[\left(\mathbf{F}\mathbf{S}\right)^T\delta\mathbf{u}\right] - \left[\nabla_\mathbf{X}\cdot\left(\mathbf{F}\mathbf{S}\right)\right]\cdot\delta\mathbf{u},
	\label{eq:pk1_stress_work_energy_expanded}
\end{equation}
which can now be substituted back into equation~\ref{eq:first_variation_with_pk2_stress} to obtain
\begin{equation}
	\delta\Pi = \int_{\mathcal{B}_0}\nabla_\mathbf{X}\cdot\left[\left(\mathbf{F}\mathbf{S}\right)^T\delta\mathbf{u}\right]dv - \int_{\mathcal{B}_0} \left[\nabla_\mathbf{X}\cdot\left(\mathbf{F}\mathbf{S}\right)\right]\cdot\delta\mathbf{u} dv - \int_{\mathcal{B}_0}\mathbf{b}\cdot\delta\mathbf{u}dv - \int_{\partial\mathcal{B}_0^{\mathbf{t}}}\tilde{\mathbf{t}}\cdot\delta\mathbf{u}da = 0.
	\label{eq:first_variation_manipulation_1}
\end{equation}
The first term in the above equation can be rewritten with the help of the divergence theorem as
\begin{equation}
	\int_{\mathcal{B}_0}\nabla_\mathbf{X}\cdot\left[\left(\mathbf{F}\mathbf{S}\right)^T\delta\mathbf{u}\right]dv = \int_{\partial\mathcal{B}_0} \left(\mathbf{F}\mathbf{S}\right)^T\delta\mathbf{u}\cdot\mathbf{n}da = \int_{\partial\mathcal{B}_0^{\mathbf{t}}} \left(\mathbf{F}\mathbf{S}\cdot\mathbf{n}\right)\cdot\delta\mathbf{u} da,
\end{equation}
where $\mathbf{n}$ is a unit surface normal in the reference configuration and the second equality comes from the fact that $\delta\mathbf{u} = 0$ on $\partial\mathcal{B}_0^\mathbf{u}$. It can be readily seen the the term $\mathbf{F}\mathbf{S}\cdot\mathbf{n}$ is the traction vector acting on the surface $\partial\mathcal{B}_0^{\tilde{\mathbf{t}}}$. 


Substituting the manipulated terms back into equation~\ref{eq:first_variation_manipulation_1} and collecting terms with common integration domains gives
\begin{equation}
	\delta\Pi = - \int_{\mathcal{B}_0} \left[\nabla_\mathbf{X}\cdot\left(\mathbf{F}\mathbf{S}\right) + \mathbf{b}\right]\cdot\delta\mathbf{u} dv + \int_{\mathcal{B}_0^{\mathbf{t}}} \left(\mathbf{F}\mathbf{S}\cdot\mathbf{n} - \tilde{\mathbf{t}}\right)\cdot\delta\mathbf{u} da = 0.
\end{equation}
According to standard results from the calculus of variations we obtain that $\nabla_\mathbf{X}\cdot\left(\mathbf{F}\mathbf{S}\right) + \mathbf{b} = \mathbf{0}$ in the interior of the body $\mathcal{B}_0$, which corresponds to the strong form of the conservation of linear momentum, and $\mathbf{F}\mathbf{S}\cdot\mathbf{n} = \tilde{\mathbf{t}}$ on the surface $\partial\mathcal{B}_0^{\tilde{\mathbf{t}}}$, which corresponds to the natural boundary conditions. Combining these results with the essential boundary condition of $\mathbf{u} = \tilde{\mathbf{u}}$ on the surface $\partial\mathcal{B}_0^{\mathbf{u}}$ gives the strong of the BVP and is thus
\begin{eqnarray}
	\nabla_\mathbf{X}\cdot\left(\mathbf{F}\mathbf{S}\right) + \mathbf{b} &=& \mathbf{0}   \quad\text{for }\mathbf{X}\in\mathcal{B}_0, \nonumber \\
	\mathbf{u}                               	   &=& \tilde{\mathbf{u}} \quad\text{for 			}\mathbf{X}\in\partial\mathcal{B}_0^\mathbf{u},\\
	\mathbf{t}                                     &=& \tilde{\mathbf{t}} \quad\text{for }\mathbf{X}\in\partial\mathcal{B}_0^\mathbf{t} \nonumber.
\end{eqnarray}

\subsection{Constitutive Models} \label{sec:constitutive_models}
In this manuscript only constitutive models that can be derived from a strain energy density, i.e. hyperelastic constitutive models, will be considered. We will denote the strain energy density by $\psi$ as done in the previous section and will consider three models in this paper. These models are the Neo-Hookean model, which is suitable for general hyperelastic material behavior for moderate strain levels; the Gent model, which is suitable for large deformation elastomeric behavior with chain lock up; and the Blatz-Ko model, which is a phenomenological model for flexible foam behavior suitable for moderate levels of compression. Our method is not limited to just these three models, but they were chosen to serve as exemplars for different types of material behavior. We will only review the forms of the strain energy of the different models since this is the only necessary constitutive ingredient for our method.

\subsubsection{Neo-Hookean} \label{sec:neohookean}
There are many different models referred to by the name "Neo-Hookean" but we will specifically utilize a model form which decouples the volumetric and isochoric material response given in~\cite{2022_lame_user_guide}. The strain energy for this particular Neo-Hookean model is given by
\begin{equation}
	\psi\left(\mathbf{C}\right) = \frac{1}{2}K\left[\frac{1}{2}\left(J^2 - 1\right) - \ln J\right] + \frac{1}{2}\mu \left(\bar{I}_1 - 3\right),
	\label{eq:neohookean_free_energy}
\end{equation}
where $K$ is the bulk modulus, $\mu$ is the shear modulus, and $\bar{I}_1$ is the first invariant of the distortional right Cauchy-Green tensor, which is defined by
\begin{equation}
	\bar{I}_1 = \tr\bar{\mathbf{C}} = J^{-2/3}\tr{\mathbf{C}}.
\end{equation}
It should be noted the the bulk modulus $K$ and the shear modulus $\mu$ are consistent with the shear and bulk modulus of the linear theory.

\subsubsection{Gent} \label{sec:gent}
The free energy for the Gent model~\cite{1996Gent} is similar to the form of the Neo-Hookean model presented in the previous section but differs in the form of the isochoric response. The strain energy density is given by
\begin{equation}
	\psi\left(\mathbf{C}\right) = \frac{1}{2}K\left[\frac{1}{2}\left(J^2 - 1\right) - \ln J\right] - \frac{1}{2}\mu J_m\ln\left(1 - \frac{\bar{I}_1 - 3}{J_m}\right),
\end{equation}
where $K$ is the bulk modulus, $\mu$ is the shear modulus, and $J_m$ is a parameter, which accounts for limited chain inextensibility. 

\subsubsection{Blatz-Ko Model} \label{sec:blatz_ko}
The free energy for the Blatz-Ko model~\cite{1962BlatzKo} is a specialized case of the Generalized Ogden model~\cite{STORAKERS1986125,ogden1984non} with the number of terms truncated to $N = 1$, $\nu = 0.25$, and $\alpha = -2.0$ which reduces to
\begin{equation}
	\psi = \frac{\mu}{2}\left(\frac{I_2}{I_3} + 2\sqrt{I_3} - 5\right).
\end{equation}
This is a single parameter model that can act as a simplified model for the description of elastomeric foams at moderate deformations, but does have a fixed Poisson response.

\subsection{Neural Networks} \label{sec:neural_networks}
Neural networks (NN) are one of the most commonly utilized machine learning (ML) algorithms currently due to recent advances in training algorithms and graphics processing unit (GPU) availability and speed. One of the simplest NN architectures is the feed-forward fully-connected neural network. This type of network is constructed by building layers of neurons and connecting neurons from previous layers to subsequent layers through a non-linear activation function. The first layer is typically called the \emph{input layer} and the last layer is usually referred to as the \emph{output layer}. The layers in between are known as \emph{hidden layers}. Hidden layers with a large number of neurons are referred to as \emph{wide} neural networks. If a network has few hidden layers it is called \emph{shallow}, and if a network has many hidden layers it is referred to as \emph{deep}, which is where the term deep learning is derived. The number of hidden layers, choice of activation function, and the number of neurons in each hidden layer can be viewed as hyperparameters, and there is no optimal set of these for all possible problems. If a feed-forward neural network has $n$ inputs and $m$ outputs, we can denote the mapping of the neural network by
\begin{equation}
	\mathcal{N}:\mathbb{R}^n\rightarrow\mathbb{R}^m,
\end{equation}
and we will denote a generic non-linear activation function by the mapping
\begin{equation}
	\sigma:\mathbb{R}\rightarrow\mathbb{R}.
\end{equation}
We denote the set of weights and biases for a given layer $l$ of the NN with $\mathcal{B}^l\in\mathbb{R}$ and $\mathcal{B}^l$ respectively. For simplicity we will assume that each hidden layer has $p$ neurons. Then $\mathcal{W}^1\in\mathbb{R}^{n\times p}$, $\mathcal{W}^{N_l}\in\mathbb{R}^{p\times m}$, and for a general hidden layer $\mathcal{W}^{l}\in\mathbb{R}^{p\times p}$. Similarly, $\mathcal{B}^1\in\mathbb{R}^n$, $\mathcal{B}^{N_l}\in\mathbb{R}^m$, and for a general hidden layer $\mathcal{B}^l\in\mathbb{R}^p$. We will denote the collection of all weights as $\mathcal{W} = \left\{\mathcal{W}_1, ..., \mathcal{W}_{N_l}\right\}$ and the collection of all biases as $\mathcal{B} = \left\{\mathcal{B}_1, ..., \mathcal{B}_{N_l}\right\}$. The general affine transformation of a given layer $l$ can be then written as
\begin{equation}
	\mathcal{T}^l\left(\cdot\right) = \mathcal{W}^l\cdot\left(\cdot\right) + \mathcal{B}^l,
	\label{eq:layer_affine_transformation}
\end{equation}
where $\cdot$ is meant to denote a generic input to this mapping. Using equation~\ref{eq:layer_affine_transformation} we can write the mapping $\mathcal{N}$ in terms of the following composition of functions:
\begin{equation}
	\mathcal{N}\left(\cdot\right) = \mathcal{T}^{N_L}\circ\sigma\circ\mathcal{T}^{N_L - 1}\circ ... \circ\sigma\circ\mathcal{T}^1\left(\cdot\right).
\end{equation}
The training of a neural network requires the definition of a loss function (to be defined for our framework later), which is typically an error measurement between NN predictions and data. Training proceeds in an iterative fashion where an optimizer updates the weights and biases of each layer typically through a variation of stochastic gradient descent (SGD). A general network parameter $w$ is updated with SGD in the following way:
\begin{equation}
	w_{n+1} = w_{n} - \eta\nabla_{w_n}\mathcal{L},
\end{equation}
where $\eta$ represents the \emph{learning rate} (another tunable hyperparameter), $\nabla_w$ is the gradient operator for $w$, and $\mathcal{L}$ is used to represent a generic loss function. This is done for each network parameter during each training iteration. The general minimization problem of neural network training can be written as follows
\begin{equation}
	\min_{\left(\mathcal{W}, \mathcal{B}\right)}\mathcal{L}.
\end{equation}

\section{Implementation} \label{sec:implementation}
In this section we will present the implementation details for how we solve solid mechanics BVPs via NNs and modifications to accommodate for inverse problems. Theoretically speaking, we take an approach that is total Lagrangian, i.e. all calculations are performed with respect to the initial configuration. We take this avenue rather then an updated-Lagrangian approach for efficiency reasons to eliminate mesh remap steps and recalculation of shape functions during training. There may be instances where an updated-Lagrangian approach is more appropriate such as multiphysics problems where one of the fields is more naturally calculated in the current configuration. According to the universal approximation theorem, neural networks can represent any continuous functions if enough layers and neurons are utilized. In that sense, utilizing a neural network to find solutions to governing equations of solid mechanics is no different in essence to utilizing isoparametric shape functions in Galerkin finite element methods. The mathematical structure is very similar, but the implementation is where the differences are seen. We implemented our method in the library JAX~\cite{jax2018github} to leverage its robust and easy-to-use interfaces for automatic differentiation, GPU vectorization, and just-in-time compilation. 
\begin{figure}
	\centering
	\includegraphics[width=0.8\textwidth]{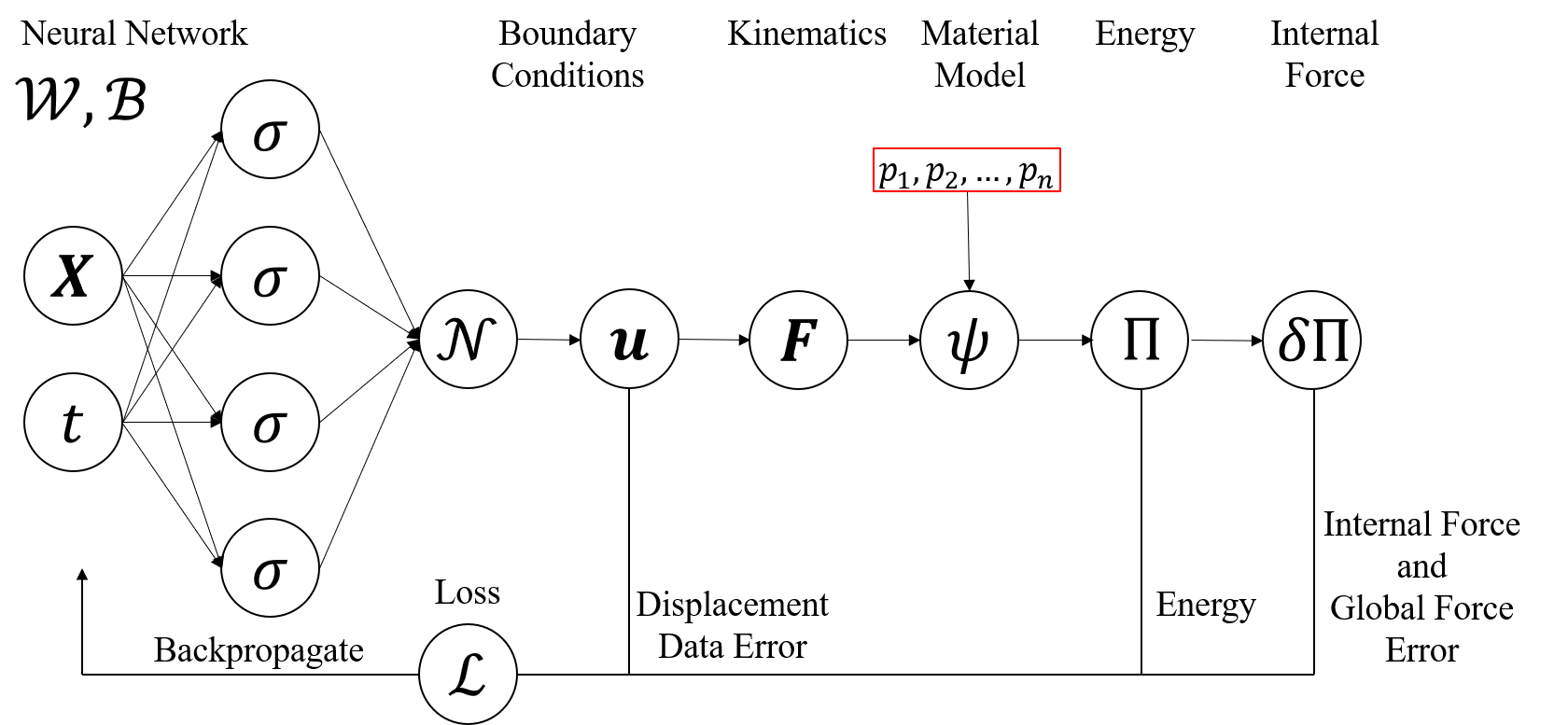}
	\caption{Schematic showing the implementation of our PINN architecture.}
\end{figure}

\subsection{Calculation of Kinematic Quantities} \label{sec:calculation_of_kinematic_quantities}
\label{sec:caluclation_of_kinematic_quantities}
The first difference between FEM and our method is the following. Rather than approximating the displacement field with shape functions, as is done in FEM, we approximate the displacement field with a NN in the following way:
\begin{equation}
	\mathbf{u}_\mathcal{N}\left(\mathbf{X}, t\right) \approx \tilde{\mathbf{u}}\left(\mathbf{X}, t\right) + g\left(\mathbf{X}\right)\mathcal{N}\left(\mathbf{X}, t\right),
	\label{eq:ansatz}
\end{equation}
where $\mathbf{u}_\mathcal{N}$ is the neural network approximation of the displacement field, $\tilde{\mathbf{u}}$ is the known displacement on the boundary $\partial\mathcal{B}^\mathbf{u}_0$, $g$ is a function that vanishes on the boundary $\partial\mathcal{B}^\mathbf{u}_0$, and $\mathcal{N}\left(\mathbf{X}, t\right)$ is a neural network that takes the reference configuration coordinates $\mathbf{X}$ (which are read in from the nodal coordinates of a mesh file) and the time $t$ as inputs. Utilizing this construction ensures that the displacement field on the boundary is exact, unlike in classical PINN formulations where BCs are enforced through penalties in the loss function. Small deviations in essential boundary conditions can lead to training issues with classical PINNs for even the simplest equations of mathematical physics~\cite{wang2020understanding}, let alone non-linear solid mechanics where small deviations in the displacement BCs can greatly change the solution. For simple BVPs on non-complex geometries, such as uniaxial tension on a cube, an analytic form of $g$ can be found. For more complicated geometries and BVPs, $g$ can be approximated via another NN that is pre-trained to be zero on the displacement boundaries as was done for linear elastodynamics in~\cite{rao2021physics}.

Now that the displacement field has been approximated via equation~\ref{eq:ansatz}, the deformation gradient can be calculated in one of two ways. The typical approach for PINNs would be to utilize automatic differentiation to calculate the following:
\begin{equation}
	\mathbf{F}_\mathcal{N} = \frac{\partial}{\partial \mathbf{X}}\left(\mathbf{X} + \mathbf{u}_\mathcal{N}\right) = \mathbf{I} + \frac{\partial\mathbf{u}_\mathcal{N}}{\partial\mathbf{X}},
\end{equation}
which would normally be calculated at collocation points. This approach is straightforward to implement in any modern NN library such as TensorFlow or PyTorch. The other advantageous feature of this approach is that it can be calculated all at once and does not necessitate the definition of shape functions, which can reduce the continuity of the deformation gradient field. The downside of this approach, however, is it increases computational cost since automatic differentiation will essentially double the depth of the network, making training more difficult, and also necessitates a large number of collocation points such that the spatial derivatives can be accurately captured.

The other approach for calculating the deformation gradient, and the approach we will utilize in this paper, is to use the same approach used in FEM. We first start by constructing the displacement gradient at the element level in the following way:
\begin{equation}
	\nabla_\mathbf{X}\mathbf{u}^{e,q}_{\mathcal{N}} = \sum_{I = 1}^{N_{n}}\mathbf{u}^I_{\mathcal{N}}\otimes\nabla_\mathbf{X} N^{I,q},
\end{equation}
where $\nabla_\mathbf{X}\mathbf{u}^{e,q}_{\mathcal{N}}$ is the displacement gradient at the element level for element $e$ at quadrature point $q$, $N_n$ is the number of nodes for a generic finite element, $\mathbf{u}^I_{\mathcal{N}}$ is the value of the NN approximated displacement field at node $I$ of element $e$, and $N^{I,q}$ is the value of the $I^{th}$ shape function calculated at quadrature point $q$. The deformation gradient at the element level can then be calculated as 
\begin{equation}
	\mathbf{F}^{e,q}_{\mathcal{N}} = \mathbf{I} + \nabla_\mathbf{X}\mathbf{u}^{e,q}_{\mathcal{N}},
\end{equation}
where $\mathbf{I}$ is the second order identity tensor. Utilizing this shape-function-based approach offers us the benefit of the wealth of FEM literature that can be consulted to tailor element types to different problem types, such as incompressibility or fracture. This approach also does not add great computational cost since the operations can be readily vectorized on GPUs. Regardless of which method above is utilized to calculate the deformation gradient, any other kinematic quantities, which may be needed for the constitutive model such as the invariants of the right Cauchy-Green tensor, can be calculated from this fundamental quantity.

\subsection{Calculation of Energy Functional and Internal Force Vector}
Moving forward we will only discuss the calculation of mechanical quantities using the shape-function-based displacement gradient approximation outlined in section~\ref{sec:caluclation_of_kinematic_quantities}. Using this approach we can calculate the NN approximated energy functional given in equation~\ref{eq:energy_functional}. However, the integral in equation~\ref{eq:energy_functional} needs to be calculated numerically. For this we will use standard Gauss quadrature. The NN approximated energy functional at a time step $n$ can be calculated in the following way:
\begin{equation}
	\Pi_\mathcal{N}^n = \sum_{e = 1}^{N_e}\sum_{q = 1}^{N_q}w_q\left(\det\mathbf{J}^{e,q}\right)\psi^{e,q}\left(\mathbf{F}_\mathcal{N}^{e,q}\right),
	\label{eq:energy_functional_nn_single_time_step}
\end{equation}
where $w_q$ is the quadrature weight for quadrature point $q$, $\det\mathbf{J}^{e,q}$ is the mapping Jacobian between the reference element and the physical element calculated at quadrature point $q$, and $\psi^{e,q}$ is the element-level strain energy from element $e$ at quadrature point $q$. We can then add the energy functionals for each time step in the BVP and define this to be our total neural network approximated energy functional, i.e.
\begin{equation}
	\Pi_\mathcal{N} = \sum_{n = 1}^{N_t} \Pi_\mathcal{N}^n.
	\label{eq:energy_functional_nn}
\end{equation} 
The internal force vector can then be calculated as follows:
\begin{equation}
	\mathbf{f}_\mathcal{N} = \delta\Pi_\mathcal{N} = \frac{\partial\Pi_\mathcal{N}}{\partial\mathbf{u}_\mathcal{N}},
	\label{eq:internal_force_nn}
\end{equation}
with the aid of automatic differentiation, which is present in any standard neural network library. It should be noted that the internal force vector $\mathbf{f}_\mathcal{N}$ will only be zero on the nodes with no essential boundary conditions when the energy is minimized. The value of the internal force vector on the nodes and degrees of freedom with essential boundary conditions correspond to reaction forces.

\subsection{Algorithmic Procedure for Forward Problems}
Although our main goal is not to use PINNs as forward solvers, we present the approach using this technique as we validate our forward solution procedure against standard FEM solutions in subsequent sections as a form of V\&V. If we define our neural network loss function to be equation~\ref{eq:energy_functional_nn}, then this collapses to a quasi-static formulation (although this quasi-static approach has not been discussed in the literature) of the deep energy method (DEM)~\cite{NGUYENTHANH2020103874} for forward problems, i.e. $\mathcal{L}_{DEM} = \Pi_\mathcal{N}$ with corresponding minimization problem
\begin{equation}
	\min_{\left(\mathcal{W}, \mathcal{B}\right)}\mathcal{L}_{DEM}.
	\label{eq:minimization_dem}
\end{equation}
However, in the course of this work we found this physics-informed loss function to be unsuitable for solving inverse problems with sparse displacement data (as is the case for realistic DIC data). Furthermore, this particular approach has been recently shown to be inadequate for resolving stress concentrations~\cite{2022Fuhg}. For these reasons we propose a slight modification to the loss function $\mathcal{L}_{DEM}$ that adds an additional term corresponding to the residual of the first variation of the energy functional given by equation~\ref{eq:internal_force_nn}. This loss function takes the following form:
\begin{equation}
	\mathcal{L}_r = \Pi_\mathcal{N} + \alpha\|\delta\Pi_\mathcal{N}\|^2_{free},
\end{equation}
where $\alpha$ is a hyperparameter to be tuned and $\|\delta\Pi_\mathcal{N}\|^2_{free}$ is the $L_2$ norm of the internal force vector calculated only at free degrees of freedom. It should be noted that there is no optimal way to choose the value of $\alpha$ due to a Pareto front~\citet{rohrhofer2021pareto}. For our purposes a value of $\alpha = 1$ was sufficient for forward problems and a value of $\alpha = 250.0$ was sufficient for inverse problems. The minimization problem is analogous to equation~\ref{eq:minimization_dem} but with a different loss function to minimize. This modification does not violate any physical laws since $\delta\Pi_\mathcal{N} = 0$ for all free degrees of freedom when $\Pi_\mathcal{N}$ is at a minimum. This additional term also helps enforce first order optimality conditions. A procedural algorithm, which does not mimic the actual implementation for simplicity of presentation, is given in algorithms~\ref{alg:forward_problem_algorithm_training} and~\ref{alg:forward_problem_algorithm}.
\begin{algorithm}
	\caption{Algorithm for training neural networks to solve forward problems}
	\label{alg:forward_problem_algorithm_training}
	\begin{algorithmic}
		\Procedure{ForwardProblem}{}
		\State Read in the mesh nodal coordinates and connectivity
		\State Pre-calculate quantities for computing integrals $w_q, \nabla_\mathbf{X}N^{I,q}, \det\mathbf{J}^{e,q}$
		\State Initialize network weights and biases $\mathcal{W},\mathcal{B}$
		\While{$epoch < N_{epoch}$ or $\mathcal{L}_r < tolerance$}
		\State $\mathcal{L}_r \gets $ PhysicsLoss$\left(\right)$
		\For{$w \in\left(\mathcal{W},\mathcal{B}\right)$}
		\State $w \gets w - \eta\nabla_w\mathcal{L}_r$
		\EndFor
		\EndWhile \\
		\Return $\left(\mathcal{W}, \mathcal{B}\right)$
		\EndProcedure
	\end{algorithmic}
\end{algorithm}
\begin{algorithm}
	\caption{Algorithm for calculating the physics based loss function}
	\label{alg:forward_problem_algorithm}
	\begin{algorithmic}
		\Procedure{PhysicsLoss}{}
		
		\State $\mathcal{L}_r \gets 0$
		\State $\mathbf{u}_\mathcal{N}\left(\mathbf{X}, t\right) \gets \tilde{\mathbf{u}}\left(\mathbf{X}, t\right) + f\left(\mathbf{X}\right)\mathcal{N}\left(\mathbf{X}, t\right)$
		\For{$n < N_t$}
		\State $\Pi^n \gets 0$
		\For{$e < N_e$}
		\For{$q < N_q$}
		\State $\nabla_\mathbf{X}\mathbf{u}_\mathcal{N}^{e,q} \gets \sum_{I = 1}^{N_n}\mathbf{u}_\mathcal{N}^I\otimes\nabla_\mathbf{X} N^{I,q}\left(\mathbf{\xi}_q\right)$
		\State $\mathbf{F}_\mathcal{N}^{e,q} \gets \mathbf{I} + \nabla_\mathbf{X}\mathbf{u}_\mathcal{N}^{e,q}$
		\State $\Pi^n \gets \Pi^n + w_q^e\left(\det\mathbf{J}^e\right)\psi^{e,q}\left(\mathbf{F}_\mathcal{N}^{e,q}\right)$
		\EndFor
		\EndFor
		\State $\mathcal{L}_r \gets \mathcal{L}_r + \Pi$
		\EndFor
		\State $\mathcal{L}_r \gets \mathcal{L}_r + \frac{\partial\mathcal{L}_r}{\partial\mathbf{u}_\mathcal{N}}$ \\
		\Return $\mathcal{L}_r$
		\EndProcedure
	\end{algorithmic}
\end{algorithm}
The details that have been omitted for brevity relate to vectorization and other GPU details, which are beyond the scope of the present work. The main takeaway on the implementation of the physics-informed loss function given in algorithm~\ref{alg:forward_problem_algorithm} is that the internal force vector is not assembled as is typically done in FEM, but rather automatic differentiation is exploited to calculate this quantity. This particular implementation choice is the most efficient approach that was considered and far surpasses the speed of standard FEM assembly on GPUs and is easily parallelizable.

\subsection{Inverse Problems}
The primary goal of this work is not to utilize PINNs as a replacement for forward problem FEM solves but rather to utilize PINNs for solving inverse solid mechanics problems for known constitutive models given some full-field displacement data and a global force-displacement response. To incorporate this data into our PINN architecture, we modify the loss function when solving inverse problems by adding additional penalty terms that aim to minimize the error between PINN predictions and the observed data. Specifically, two terms need to be added. These are an error term for the displacement data, which is given by
\begin{equation}
	\mathcal{L}_\mathbf{u} = \frac{1}{N_\mathbf{u}}\sum_{i=1}^{N_\mathbf{u}}\|\mathbf{u}_\mathcal{N}\left(\mathbf{X}^*_i, t^*_i\right) - \mathbf{u}^*_i\left(\mathbf{X}^*_i, t^*_i\right)\|^2,
\end{equation}
where $N_\mathbf{u}$ is the number of available displacement data points, $\mathbf{X}^*_i$ represents the coordinates of a particular measurement (these do not have to align with the computational mesh), $t^*_i$ represents the time a particular measurement was taken (again this does not have to line up with the computational time of the physics-informed loss function), $\mathbf{u}_\mathcal{N}$ corresponds to the neural network approximation of the displacement at coordinates $\mathbf{X}^*_i$ and time $t^*_i$, and $\mathbf{u}^*_i$ represents the measured displacement at $\mathbf{X}^*_i$ and $t^*_i$.

The second additional term is a penalty term corresponding to the error between the PINN predicted global force and the measured global force. We denote this as follows:
\begin{equation}
	\mathcal{L}_f = \frac{1}{N_t}\sum_{n=1}^{N_t}\|f_{net}\left(t_n\right) - f_{net}^*\left(t_n\right)\|^2,
\end{equation}
where $N_t$ is the number of time steps here assumed to be equal to the number of global force measurements for simplicity, $t_n$ is the time at which a particular global force measurement was made, $f_\mathcal{N}$ is the PINN predicted global force, and $f^*_n$ is the global force measurement at time $t_n$. For a BVP with a surface with unit normal $\mathbf{n}\left(\mathbf{X}\right)$, the global net force acting on that surface can be calculated as
\begin{equation}
	f_{net}\left(t_n\right) = \sum_{I=1}^{N_n^s} \mathbf{f}\left(\mathbf{X}_I, t_n\right)\cdot\mathbf{n}\left(\mathbf{X}_I\right),
\end{equation}
where $N_n^s$ is the number of nodes on that surface, $\mathbf{f}\left(\mathbf{X}_I, t_n\right)$ is the internal force at node $I$ and time $t_n$, and $\mathbf{n}\left(\mathbf{X}_I\right)$ is the unit surface normal at node $I$.
The total loss function for an inverse problem with full-field displacement and global force data can then be formulated as 
\begin{equation}
	\mathcal{L} = \beta\mathcal{L}_r + \gamma\mathcal{L}_\mathbf{u} + \delta\mathcal{L}_f,
\end{equation}
where $\beta$, $\gamma$, and $\delta$ are additional hyperparameters to balance the contributions of the different terms. In general there is no known way to pick optimal values for $\beta$, $\gamma$, and $\delta$ a priori due to a Pareto front as discussed in~\cite{wang2020understanding,rohrhofer2021pareto}. However, through changes in existing architectures we have greatly reduced the number of hyperparameters that appear in the total loss function for inverse problems in solid mechanics. Only four hyperparameters are necessary to tune to solve inverse problems unlike other approaches in the literature. Specifically the values for the hyperparameters that we utilized for inverse problems $\alpha = 250$, $\beta = 1$, $\gamma = 1$, and $\delta = 1$.

The last ingredient for solving inverse problems is to add additional parameters to the neural network optimizer that correspond to unknown material properties such as Young's modulus and Poisson's ratio in a linear elastic theory. We will denote these parameters by $p_1$, $p_2$, ..., $p_{N_p}$ where $N_p$ corresponds to the total number of unknown material properties. For the models presented earlier this would correspond to values of 1, 2, and 3 for Blatz-Ko, Neo-Hookean, and Gent respectively. For convenience we will denote the entire collection of unknown material parameters by $\mathcal{P}$. These material parameters are optimized in an analogous fashion to weights and biases of the neural network with SGD. A procedural algorithm for solving inverse problems with our approach is given in~\ref{alg:inverse_problem_algorithm}.
\begin{algorithm}
	\caption{Algorithm for solving inverse problems with full-field displacement and net force}
	\label{alg:inverse_problem_algorithm}
	\begin{algorithmic}
		\Procedure{InverseProblem}{}
		\State Read in the mesh nodal coordinates and connectivity
		\State Pre-calculate quantities for computing integrals $w_q, \nabla_\mathbf{X}N^{I,q}, \det\mathbf{J}^{e,q}$
		\State Read in data for displacement measurement coordinates and values $\mathbf{X}^*$ and $\mathbf{u}^*$
		\State Read in force data $f_{net}^*$
		\State Initialize network weights and biases $\mathcal{W},\mathcal{B}$
		\While{$epoch < N_{epoch}$ or $\mathcal{L} < tolerance$}
		\State $\mathcal{L}_r \gets $ PhysicsLoss$\left(\right)$
		\State $\mathcal{L}_\mathbf{u} \gets \frac{1}{N_\mathbf{u}^*}\sum_{n = 1}^{N_\mathbf{u}^*}\|\mathbf{u}^*\left(\mathbf{X}^*, t^*\right) - \mathbf{u}_\mathcal{N}\left(\mathbf{X}^*, t^*\right)\|^2$
		\State $\mathcal{L}_f \gets \frac{1}{N_f^*}\sum_{n = 1}^{N_f^*}\|f^*\left(t^*\right) - f_\mathcal{N}\left(t^*\right)\|^2$
		\State $\mathcal{L} \gets \alpha\mathcal{L}_r + \beta \mathcal{L}_\mathbf{u} + \gamma\mathcal{L}_f$
		\For{$w \in\left(\mathcal{W},\mathcal{B}\right)$}
		\State $w \gets w - \eta\nabla_w\mathcal{L}$
		\EndFor
		\For{$p \in\mathcal{P}$}
		\State $p \gets p - \eta\nabla_p\mathcal{L}$
		\EndFor
		\EndWhile
		\State \Return $\left(\mathcal{W}, \mathcal{B}, \mathcal{P}\right)$
		\EndProcedure
	\end{algorithmic}
\end{algorithm}

\section{Results} \label{sec:results}
In this section we present results of both forward problems (for the purposes of method validation) and inverse problems, which is our real interest. The geometrical dimensions are not meant to be realistic but chosen for convenience and simplicity. Specifically in this work we utilized a multi-layer perceptron with 5 hidden layers and 50 neurons per hidden layer with ReLU activation functions. For training, we utilized the Adam optimizer~\cite{kingma2017adam} with a learning rate or $\eta = 0.001$. Our PINNs were trained on a Nvidia V100 (Volta) with 32 GB of graphics memory and 2 petaFLOPS of processing power. Our meshes were generated with the meshing software Cubit~\cite{2022_CubitOnlineManual_0} and our FEM/PINN 3D visualizations were performed using Paraview.

To benchmark our PINN's ability to solve forward problems, we solve identical BVPs via FEM. Here we used the SIERRA Solid Mechanics~\cite{2022_sierrasm_userguide}, developed at Sandia National Laboratories, as our benchmark. Specifically fully-integrated Hex8 elements were utilized. We utilize similar simulations to generate synthetic DIC experimental data as ground truth data for material property discovery in inverse problems. For all cases considered, only displacement-controlled boundary conditions will be utilized such that there are no applied tractions or body forces, i.e. $\mathbf{b} = \mathbf{0}$ and $\tilde{\mathbf{t}} = \mathbf{0}$. All BVPs considered will have a uniaxial motion such that the neural network approximation of the displacement field can be written for this special case as
\begin{equation}
	\mathbf{u}\left(\mathbf{X}, t\right) = \left(\frac{X_1}{L}\right)\begin{bmatrix}
		\hat{u}\left(t\right) \\ 
		0 \\ 
		0
	\end{bmatrix} + tX_1\left(X_1 - L\right)\mathcal{N}\left(\mathbf{X}, t\right),
\end{equation}
where we use $X_1$ to be the direction of stretching or compressing, $L$ is the total length of the specimen in this direction, and $\hat{u}$ is the global displacement in the direction of stretching.

\subsection{Forward Problems} \label{sec:forward_problems}
Here we explore the ability of our method to solve forward solid mechanics problems and compare them to solutions from a finite element code for each of the models listed in section~\ref{sec:constitutive_models}. Our goal here is to show that for these models with varying degrees of non-linearities, our method is reasonably accurate enough for predicting displacement fields while balancing forces in a quasi-static sense. To ensure an ideal one-to-one comparison between our PINN and FEM solutions, we utilized an identical element formulation in the PINN implementation as was done for FEM solves.

\begin{figure}
	\centering
	\begin{subfigure}{0.4\textwidth}
		\centering
		\includegraphics[width=\textwidth]{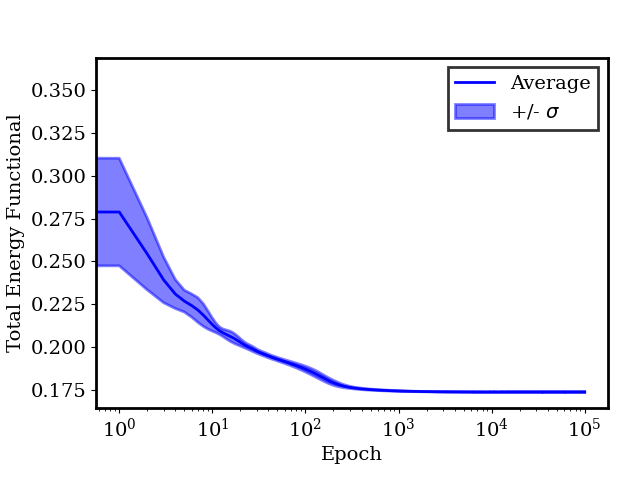}
		\caption{}
	\end{subfigure}
	\begin{subfigure}{0.4\textwidth}
		\centering
		\includegraphics[width=\textwidth]{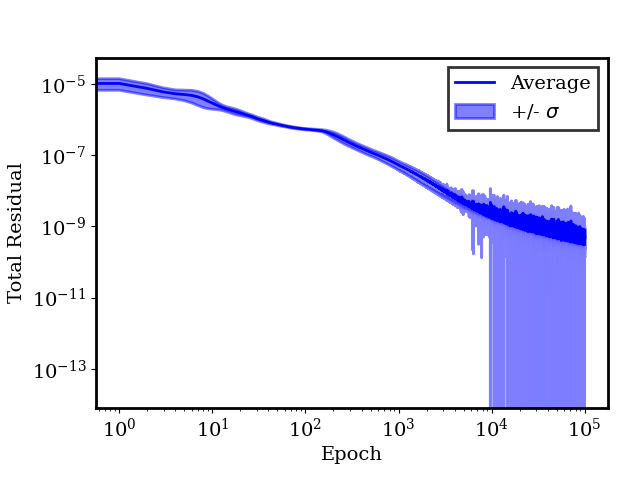}
		\caption{}
	\end{subfigure}
	\begin{subfigure}{0.4\textwidth}
		\centering
		\includegraphics[width=\textwidth]{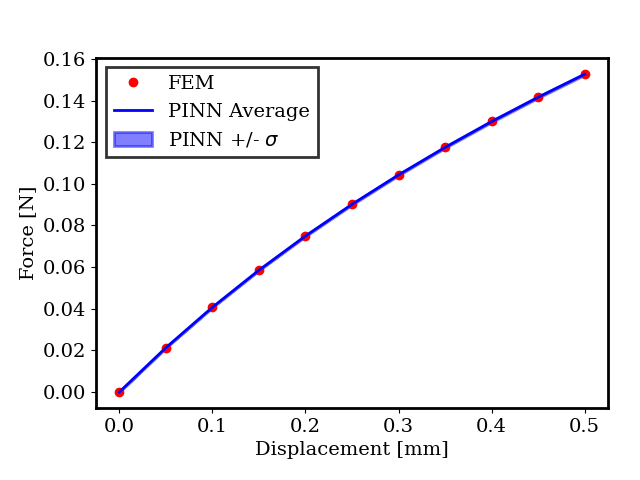}
		\caption{}
	\end{subfigure}
	\caption{Results for 8 separate training runs for a Neo-Hookean model with a Poisson's ratio of $\nu = 0.3$. (a) Total energy functional convergence over epochs. (b) Total residual convergence over epochs. (c) Comparison of global force-displacement curve between FEM results and the trained PINN after 100,000 epochs.}
\label{fig:neohookean_poissons_ratio_0_3_global_results}
\end{figure}

\begin{figure}
	\centering
	\begin{subfigure}{0.3\textwidth}
		\centering
		\includegraphics[width=\textwidth]{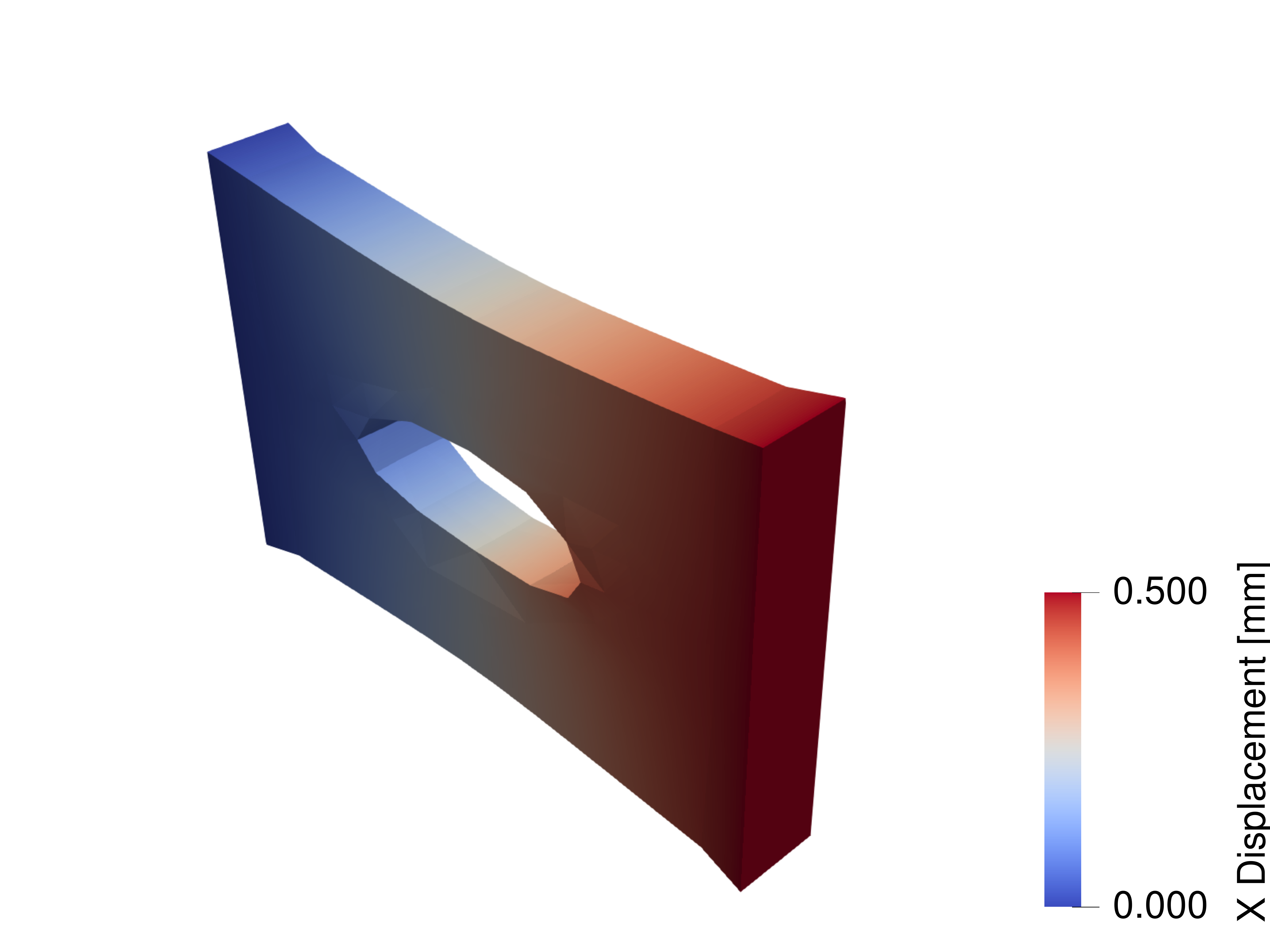}
		\caption{FEM $u_x$}
	\end{subfigure}
	\begin{subfigure}{0.3\textwidth}
		\centering
		\includegraphics[width=\textwidth]{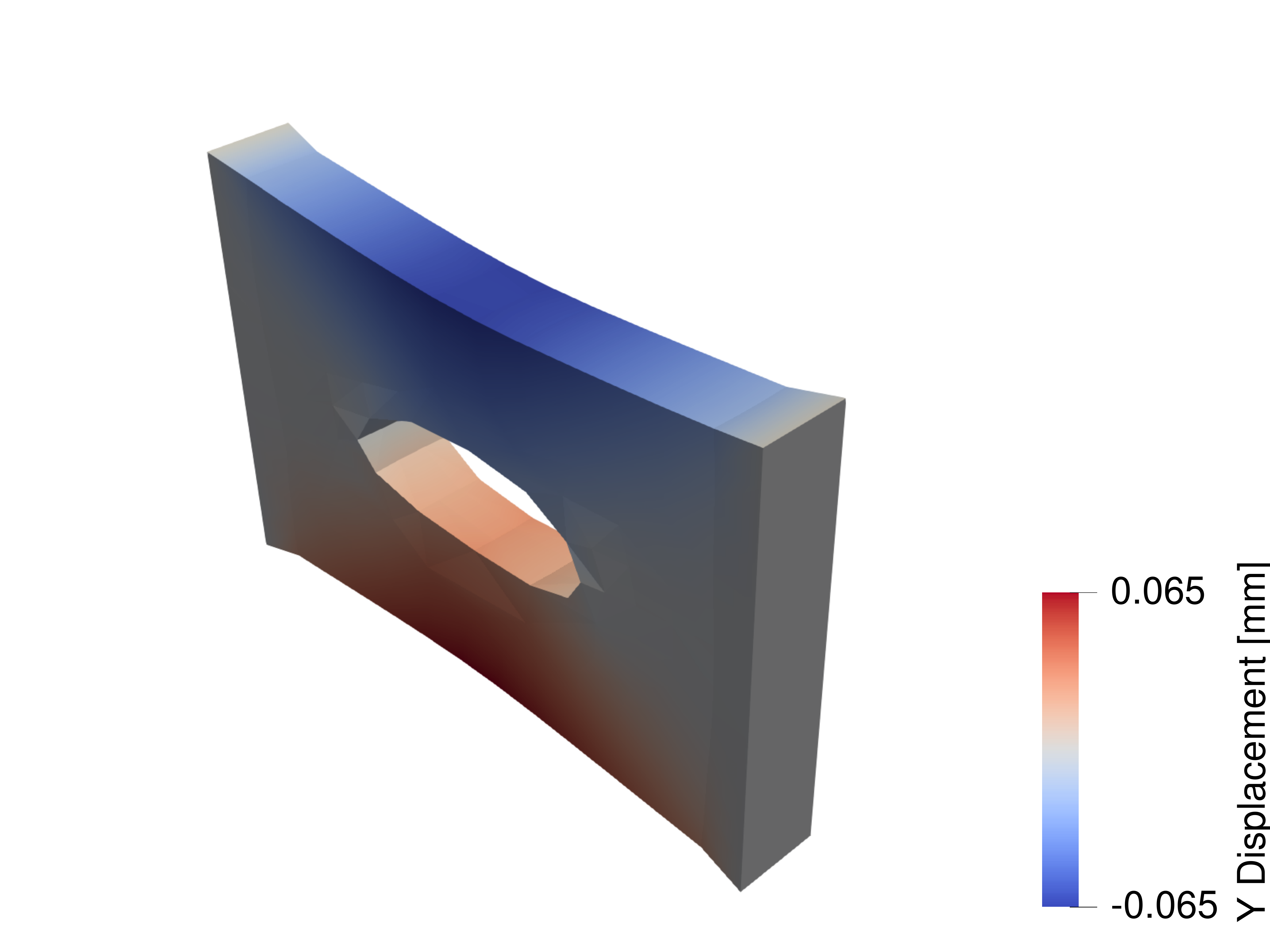}
		\caption{FEM $u_y$}
	\end{subfigure}
	\begin{subfigure}{0.3\textwidth}
		\centering
		\includegraphics[width=\textwidth]{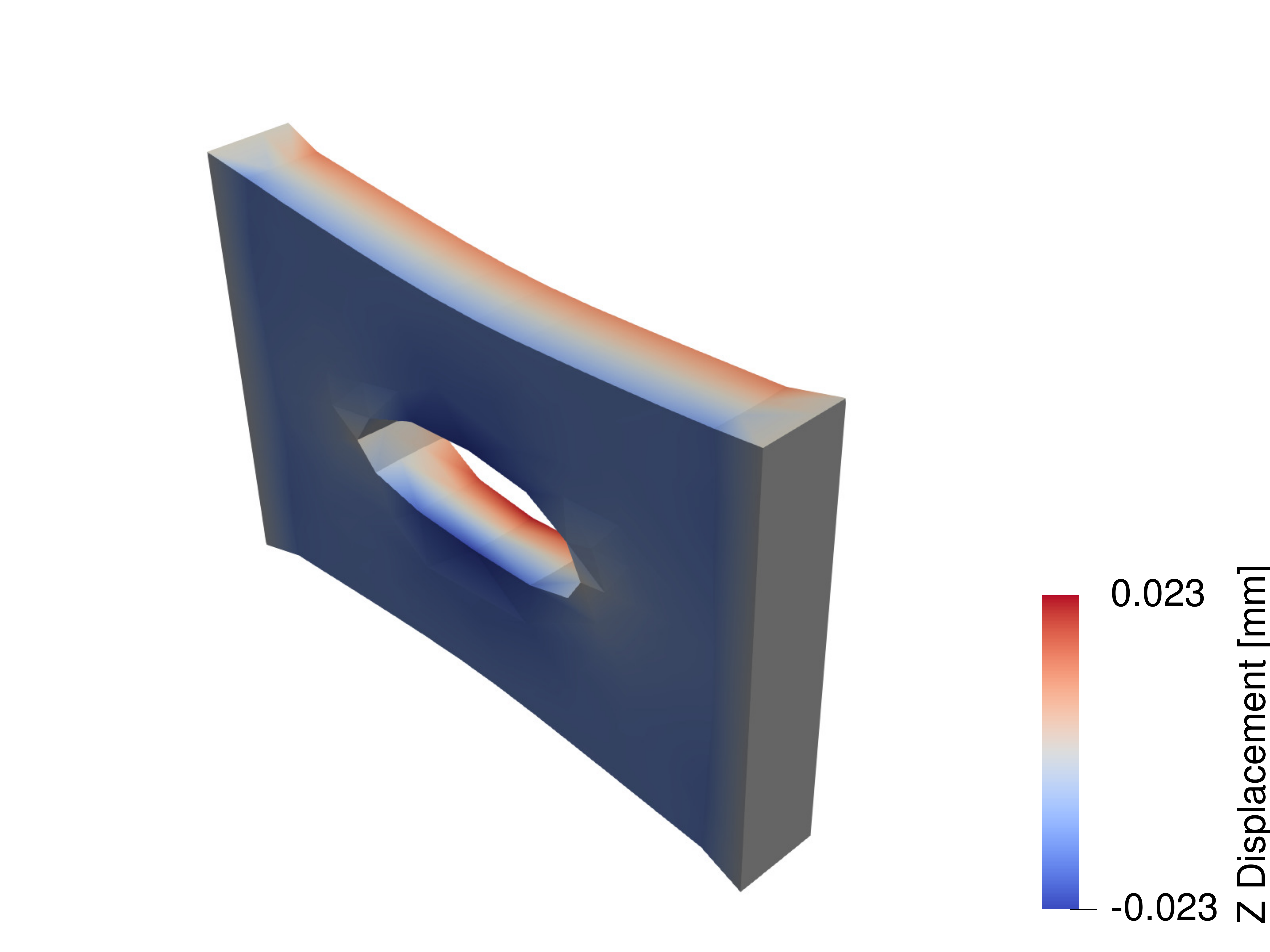}
		\caption{FEM $u_z$}
	\end{subfigure}
	\begin{subfigure}{0.3\textwidth}
		\centering
		\includegraphics[width=\textwidth]{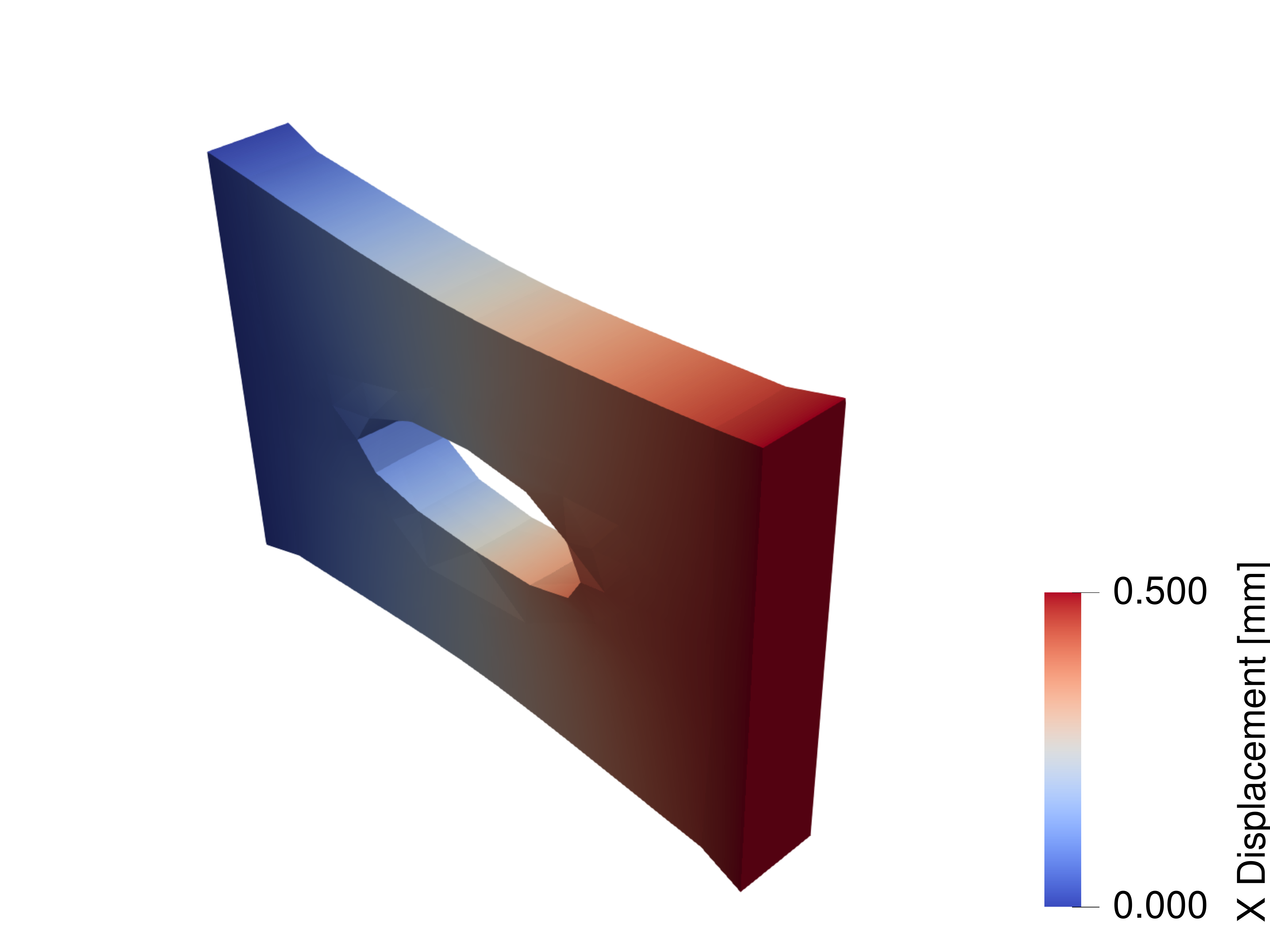}
		\caption{PINN $u_x$}
	\end{subfigure}
	\begin{subfigure}{0.3\textwidth}
		\centering
		\includegraphics[width=\textwidth]{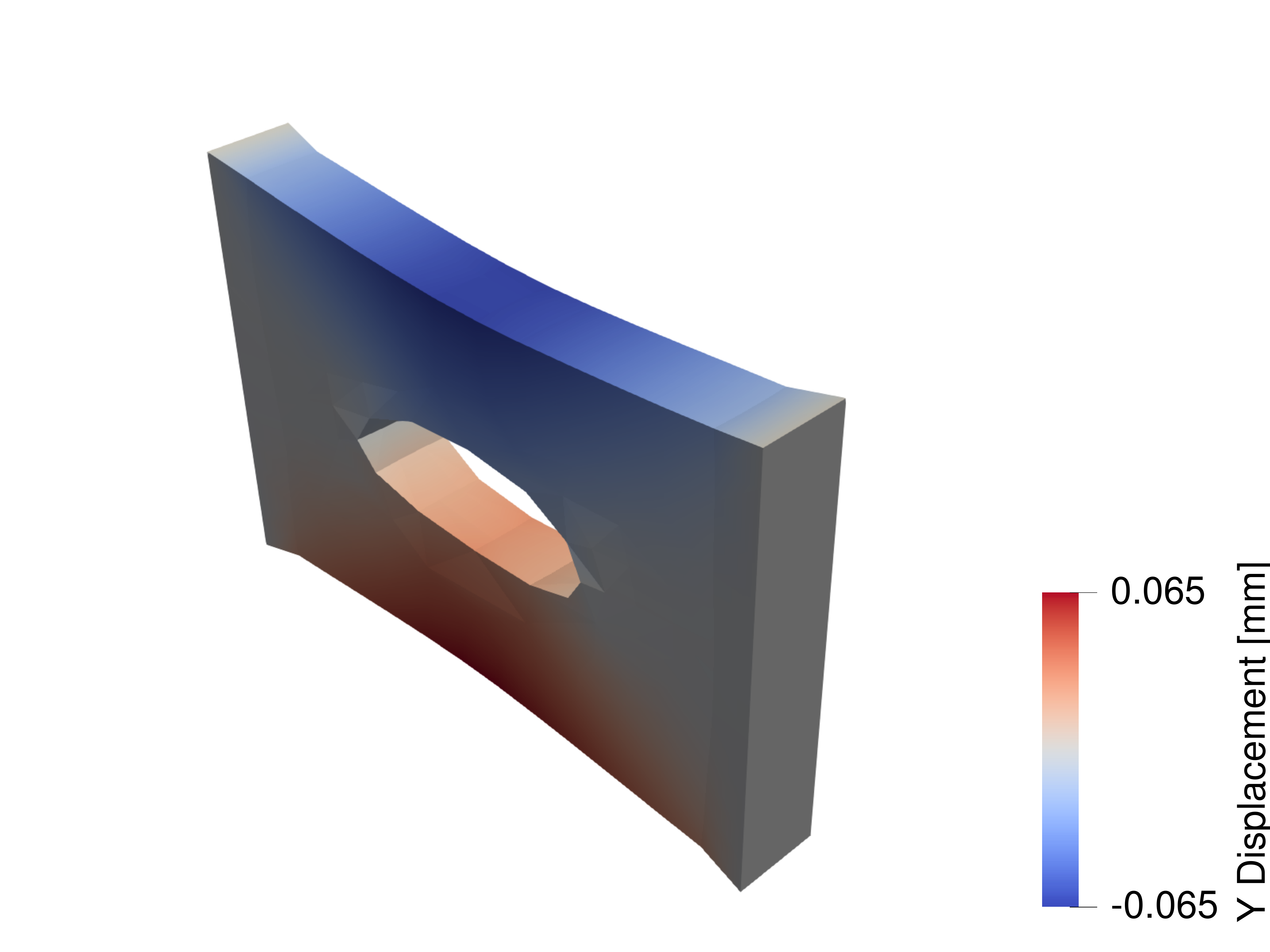}
		\caption{PINN $u_y$}
	\end{subfigure}
	\begin{subfigure}{0.3\textwidth}
		\centering
		\includegraphics[width=\textwidth]{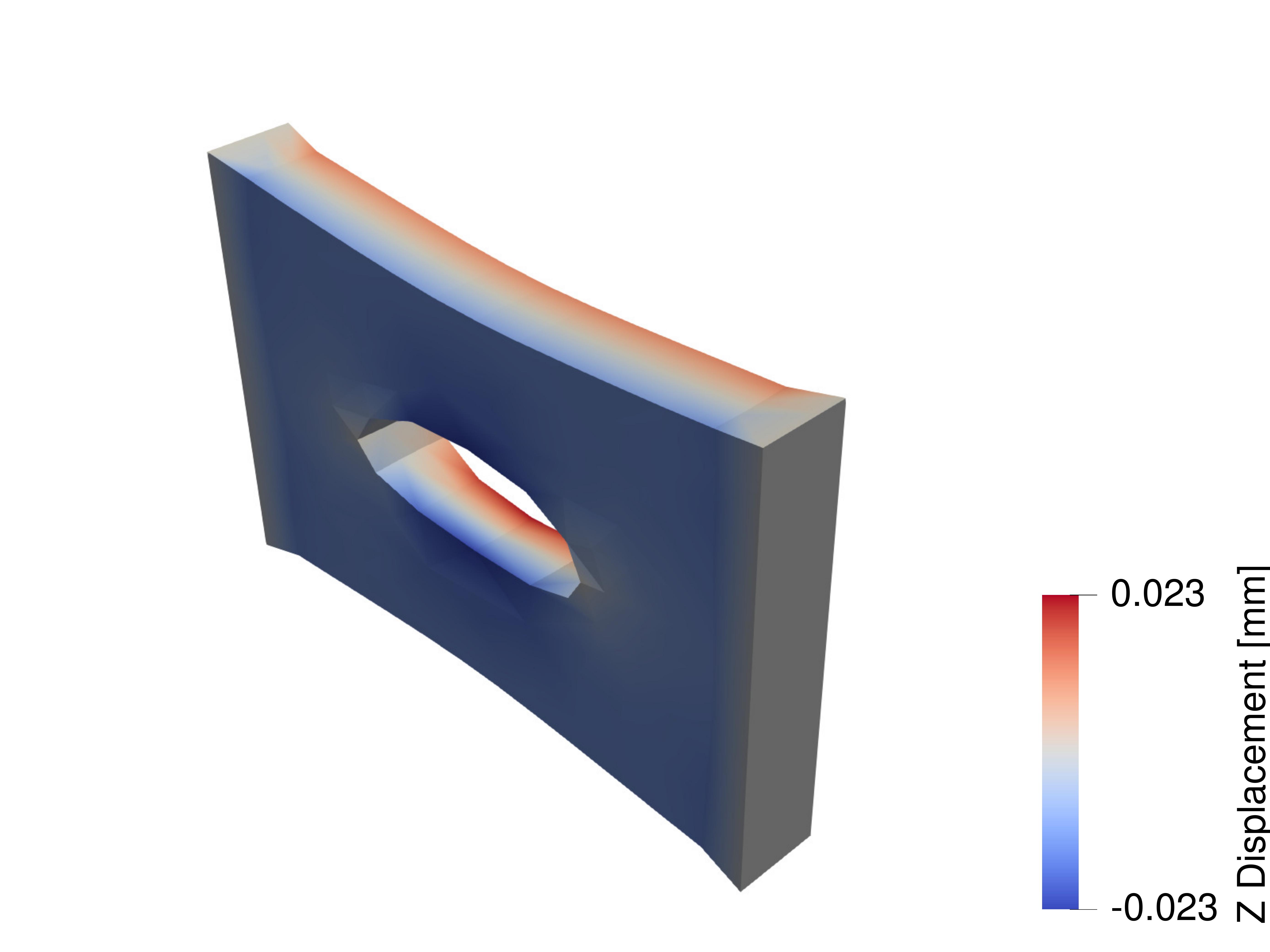}
		\caption{PINN $u_z$}
	\end{subfigure}
	\caption{Comparison of displacement field components between FEM simulations and PINN training for analogous BVPs for a Neo-Hookean model with a Poisson's ratio of $\nu = 0.3$.}
\label{fig:neohookean_poissons_ratio_0_3_displacement}
\end{figure}

\begin{figure}
	\centering
	\begin{subfigure}{0.3\textwidth}
		\centering
		\includegraphics[width=\textwidth]{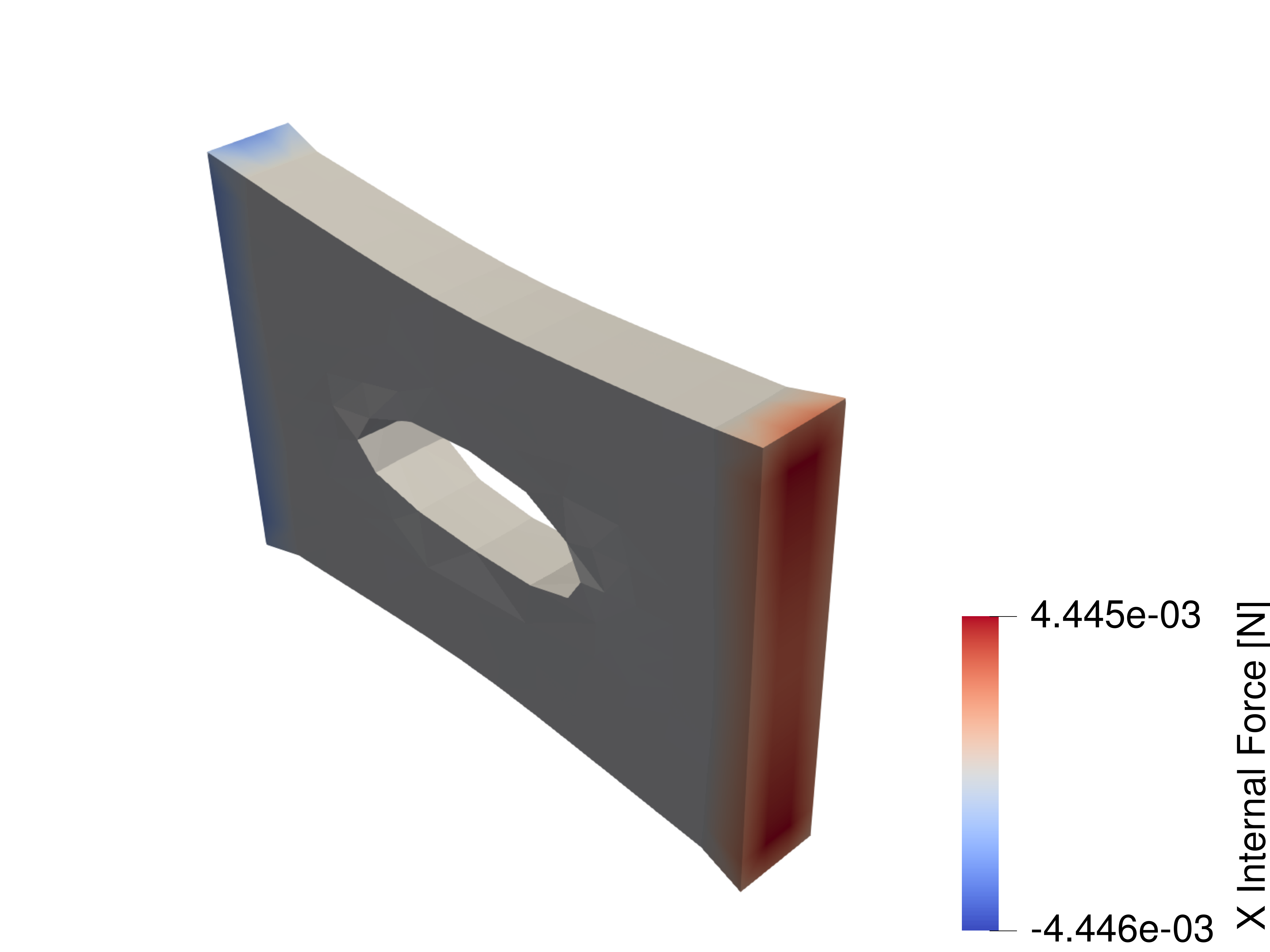}
		\caption{FEM $f_x$}
	\end{subfigure}
	\begin{subfigure}{0.3\textwidth}
		\centering
		\includegraphics[width=\textwidth]{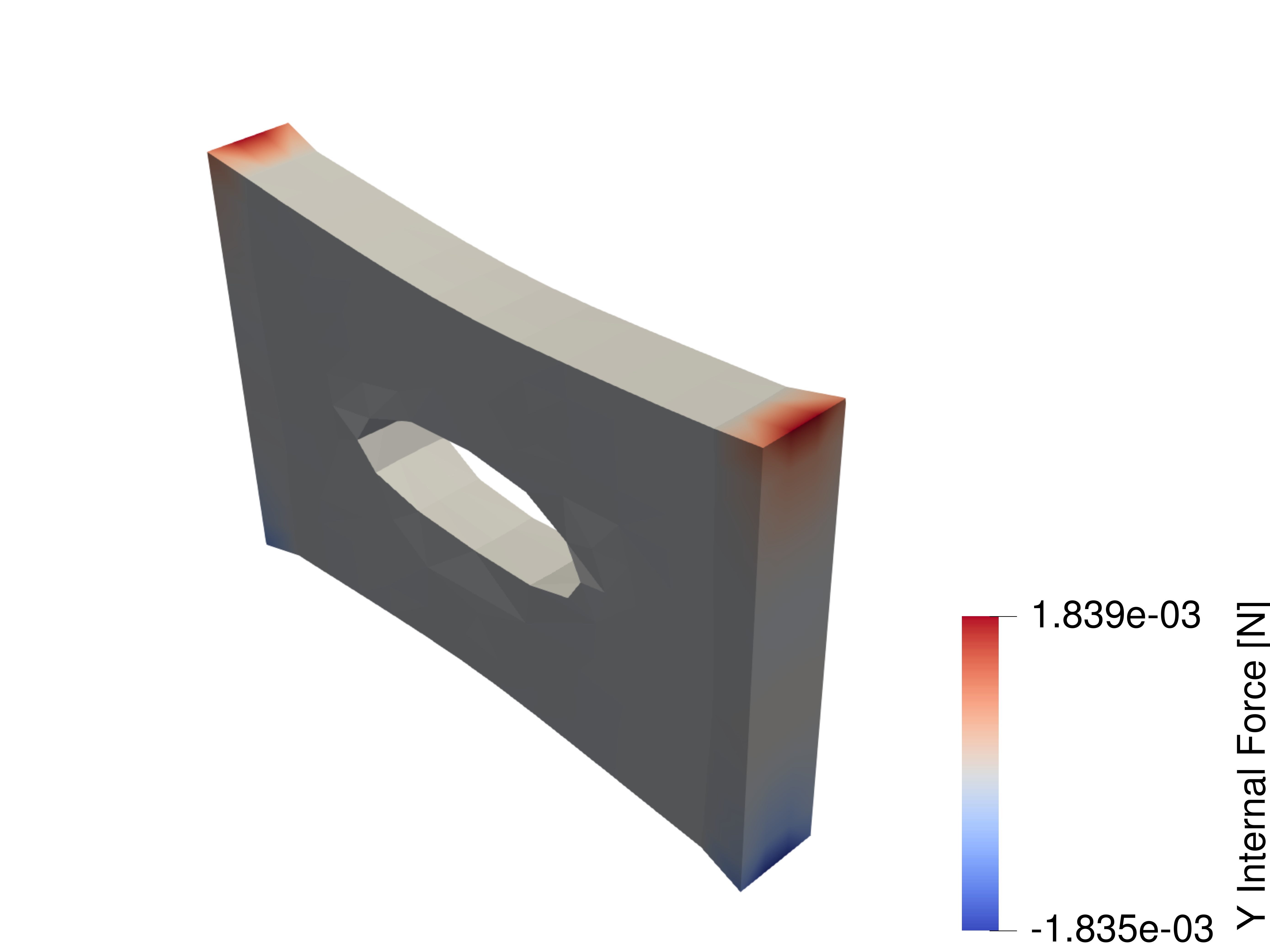}
		\caption{FEM $f_y$}
	\end{subfigure}
	\begin{subfigure}{0.3\textwidth}
		\centering
		\includegraphics[width=\textwidth]{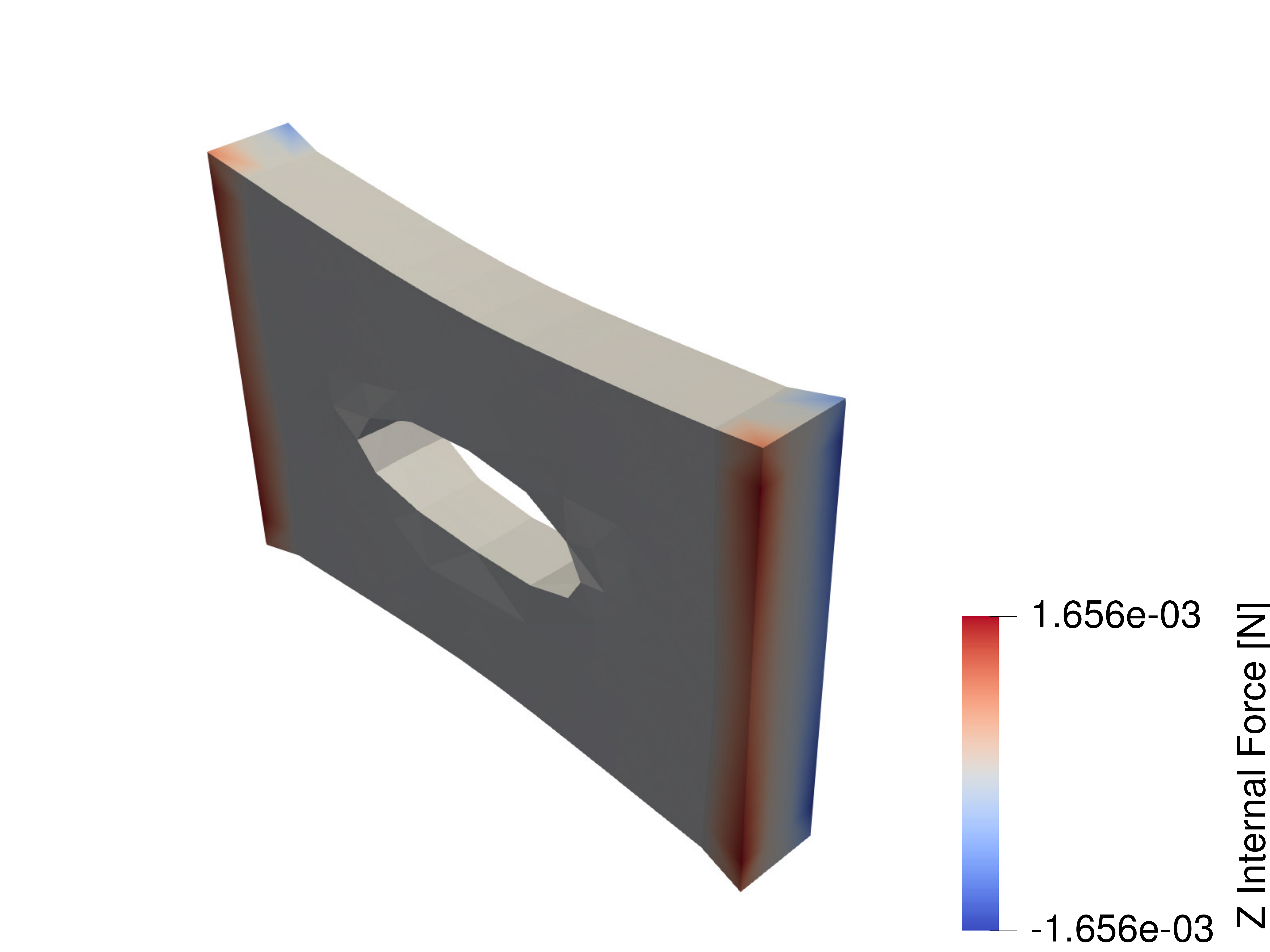}
		\caption{FEM $f_z$}
	\end{subfigure}
	\begin{subfigure}{0.3\textwidth}
		\centering
		\includegraphics[width=\textwidth]{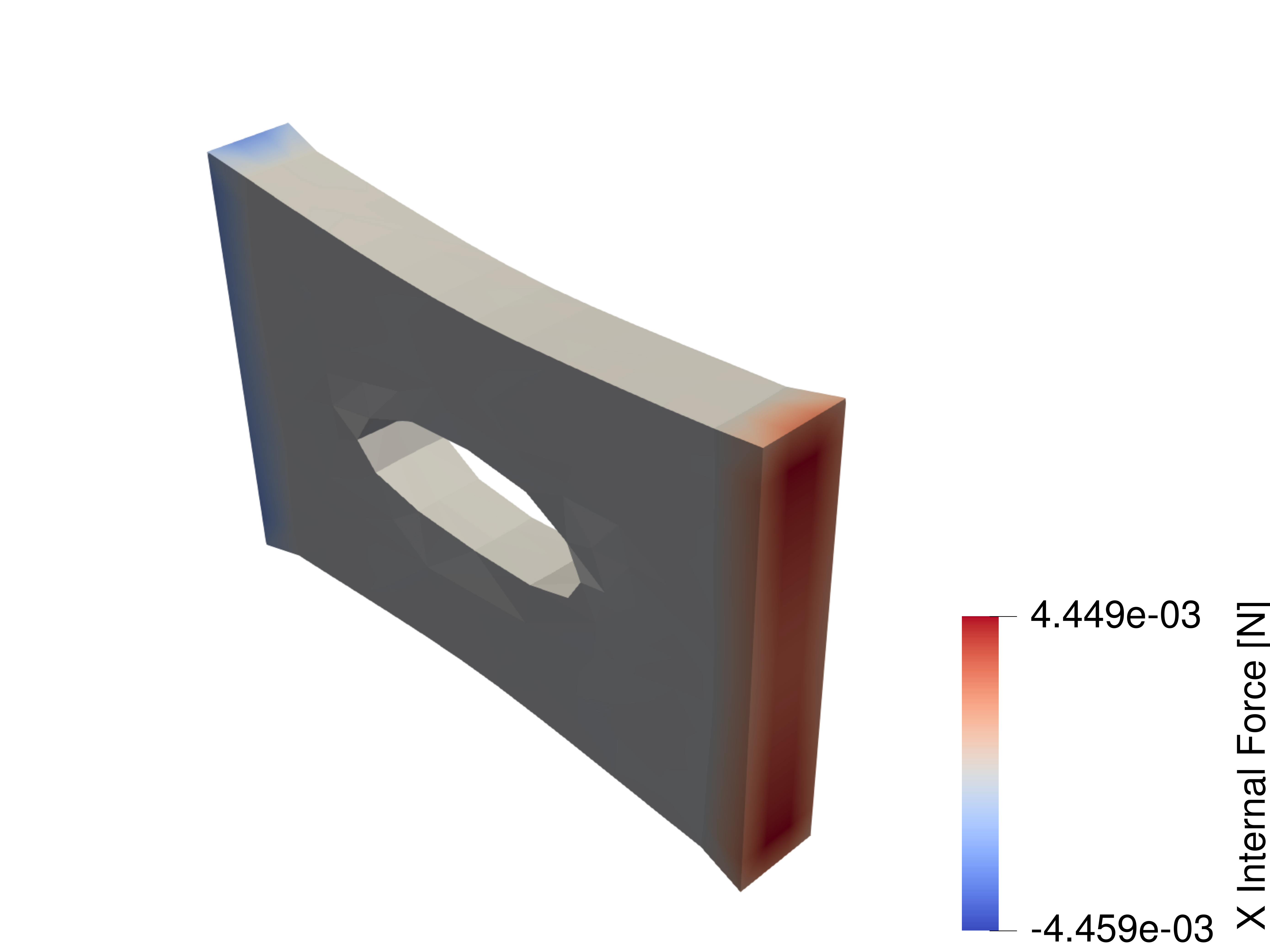}
		\caption{PINN $f_x$}
	\end{subfigure}
	\begin{subfigure}{0.3\textwidth}
		\centering
		\includegraphics[width=\textwidth]{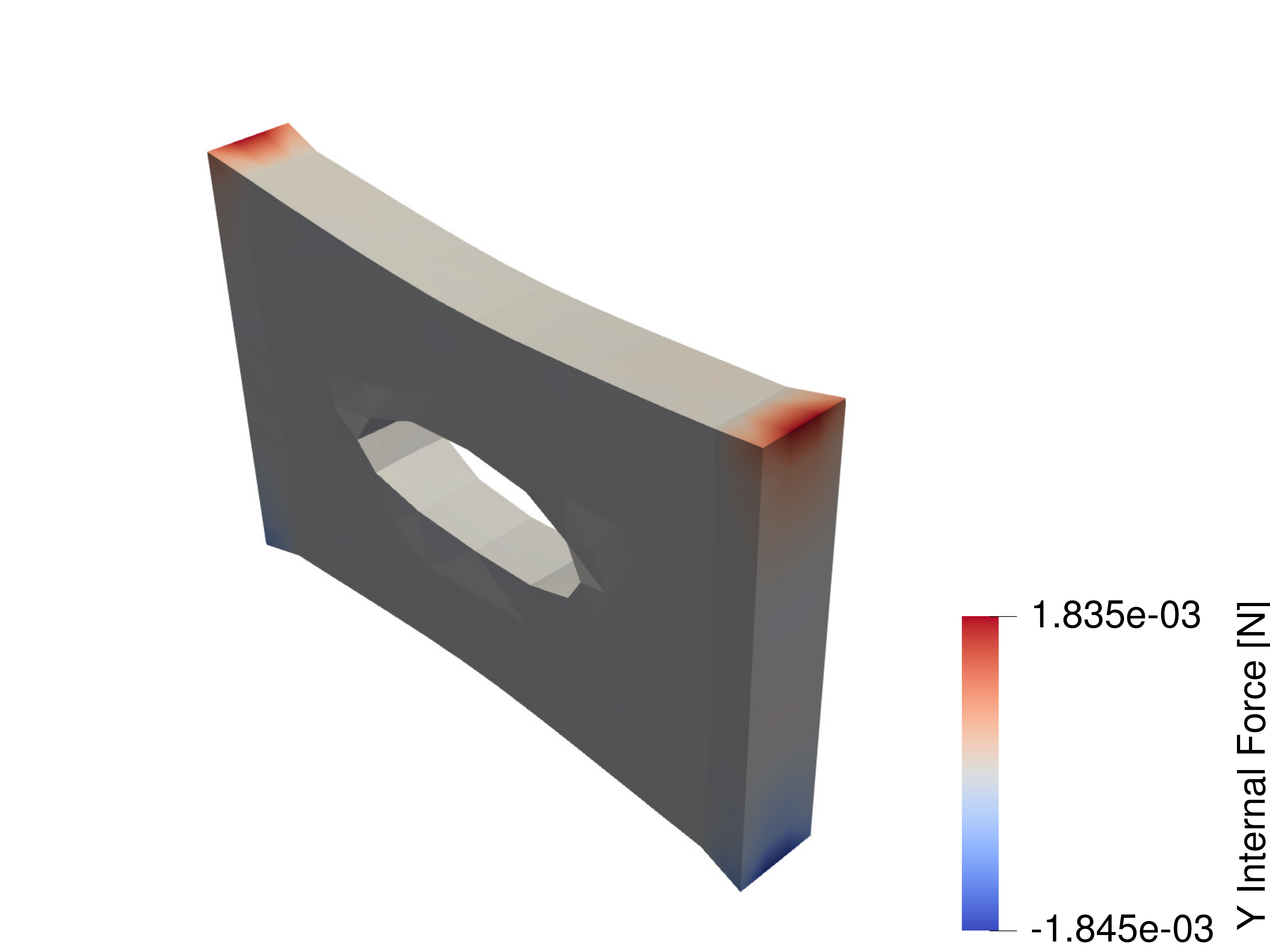}
		\caption{PINN $f_y$}
	\end{subfigure}
	\begin{subfigure}{0.3\textwidth}
		\centering
		\includegraphics[width=\textwidth]{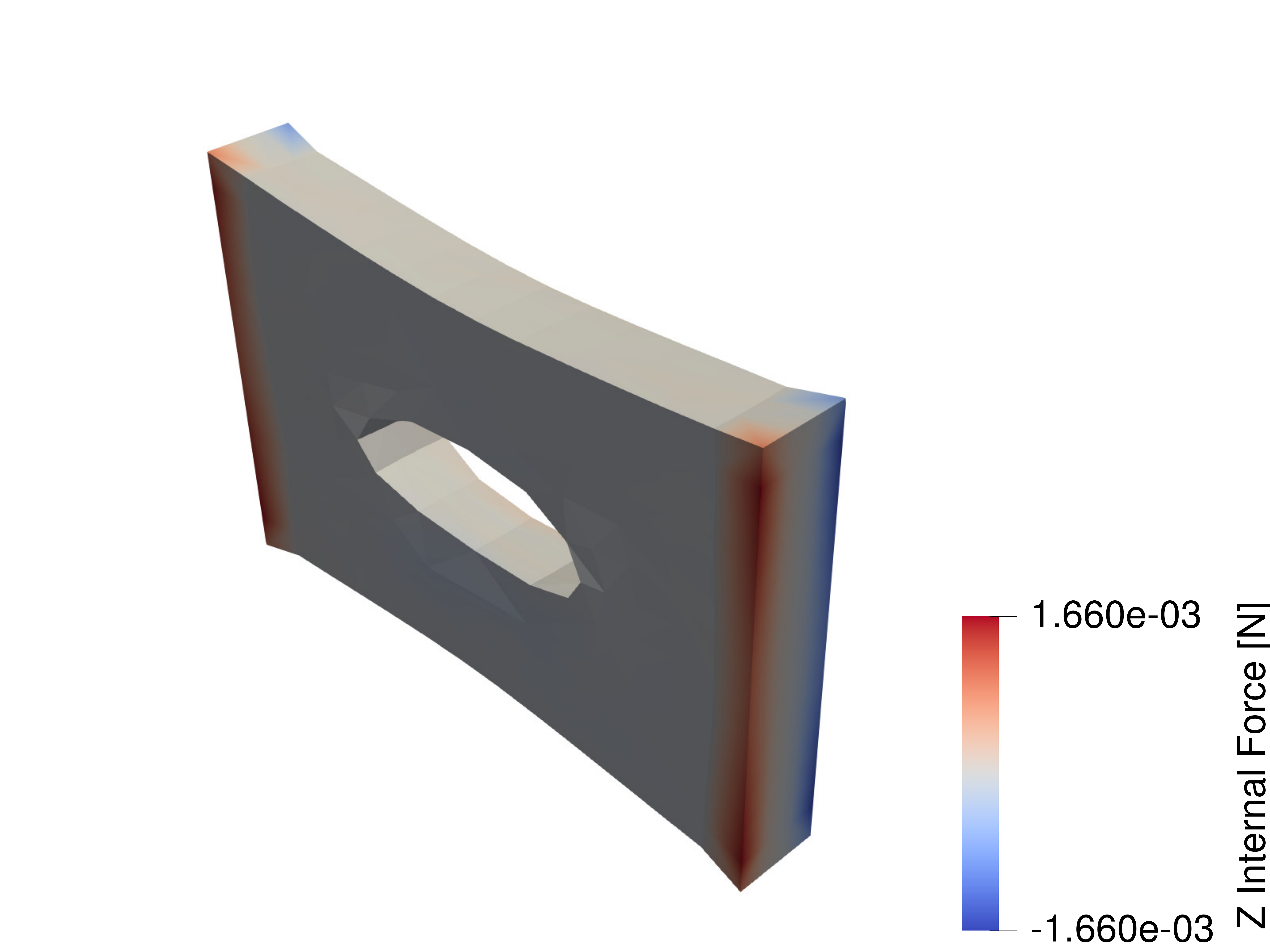}
		\caption{PINN $f_z$}
	\end{subfigure}
	\caption{Comparison of internal force field components between FEM simulations and PINN training for analogous BVPs for a Neo-Hookean model with a Poisson's ratio of $\nu = 0.3$.}
\label{fig:neohookean_poissons_ratio_0_3_internal_force}
\end{figure}

Our first benchmark problem involves a simple square slab of material with a hole punched through it pulled in uniaxial tension. The dimensions of the specimen are 1mm x 1mm x 0.2mm with a hole radius of 0.15mm punched through the center. The constitutive model for this case is a Neo-Hookean model with material properties of $K = 2.167$ MPa and $\mu = 1$ MPa, which corresponds to a material with a Poisson's ratio of $\nu = 0.3$. The left face of the structure had all three components of the displacement fixed, while the right face had the $y$ and $z$ components fixed with the $x$ displacement prescribed to have a final displacement of 0.5mm or 50\% nominal strain. Ten time steps were utilized both in the FEM solve and the PINN training to mimic a quasi-static response. The geometry was meshed with 438 Hex8 finite elements, which corresponds to 35,040 quadrature point evaluations per epoch of training. Eight separate PINNs with identical architectures but different initial values for the weights and biases were trained to show the repeatability of the method due to the stochasticity of the NN optimizer, and each PINN was trained for 100,000 epochs. The evolution of the PINN energy and residual during training and the comparison of the PINN global force-displacement to FEM are shown in Figure~\ref{fig:neohookean_poissons_ratio_0_3_global_results}.
It can be seen that the method is very repeatable in terms of energy convergence, residual convergence, and also in terms of the global force-displacement curve when compared to FEM. The energy remains relatively unchanged after about 500-1000 epochs, which is typically where a DEM approach would be deemed have converged. However, the energy from 1,000 epochs onward reduces another two decades on the log scale to properly balance the internal force vector to values reasonably close to zero. The three components of the displacement field at maximum global strain for both the FEM solution and one of the trained PINNs after 100,000 epochs are shown in Figure~\ref{fig:neohookean_poissons_ratio_0_3_displacement}.
As can be seen from the comparisons of the displacement fields, our PINN does a reasonably accurate job of predicting the full-field displacement field when compared to FEM. However, a reasonable prediction of the displacement field does not necessarily mean that physics is being properly obeyed throughout the domain. To show this, we compare the three components of the internal force field of the FEM solve and one of the trained PINNs after 100,000 epochs at maximum global strain in Figure~\ref{fig:neohookean_poissons_ratio_0_3_internal_force}.
Although very small deviations of the reaction forces are seen between the trained PINN and the FEM solution, the PINN is doing a quite reasonable job of predicting the reactions forces up to a few decimal places of the FEM solution as well as ensuring the forces are near zero in the interior of the domain.

\begin{figure}
	\centering
	\begin{subfigure}{0.4\textwidth}
		\centering
		\includegraphics[width=\textwidth]{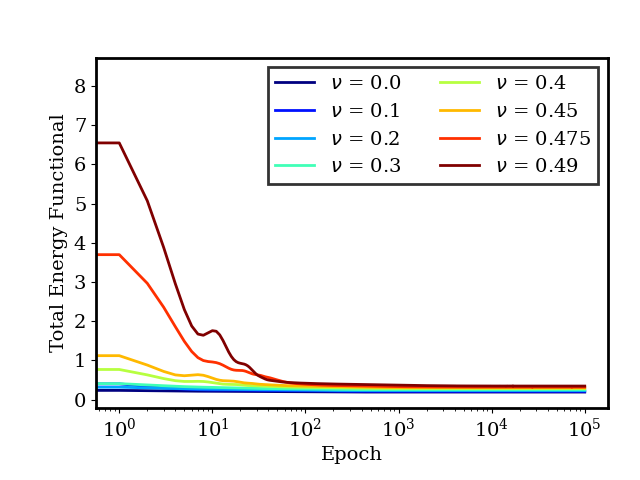}
		\caption{}
		\label{fig:neohookean_different_poissons_ratios_energy}
	\end{subfigure}
	\begin{subfigure}{0.4\textwidth}
		\centering
		\includegraphics[width=\textwidth]{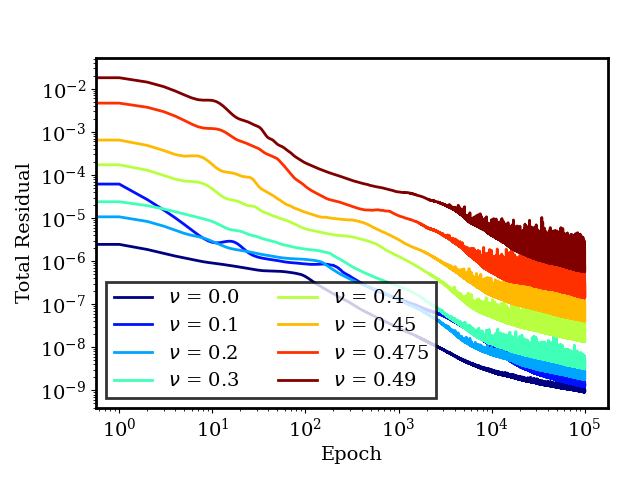}
		\caption{}
		\label{fig:neohookean_different_poissons_ratios_residual}
	\end{subfigure}
	\begin{subfigure}{0.4\textwidth}
		\centering
		\includegraphics[width=\textwidth]{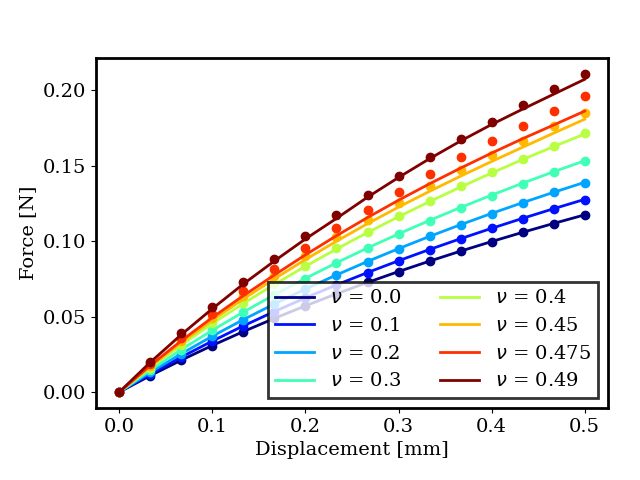}
		\caption{}
		\label{fig:neohookean_different_poissons_ratios_force_displacement}
	\end{subfigure}
	\begin{subfigure}{0.4\textwidth}
		\centering
		\includegraphics[width=\textwidth]{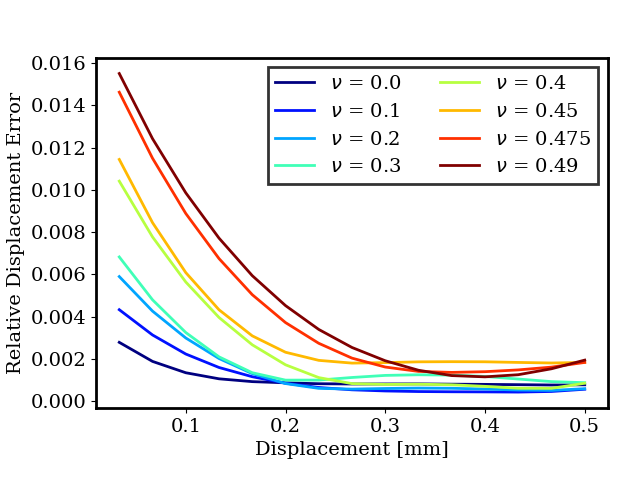}
		\caption{}
		\label{fig:neohookean_different_poissons_ratios_displacement_relative_error}
	\end{subfigure}
	\begin{subfigure}{0.4\textwidth}
		\centering
		\includegraphics[width=\textwidth]{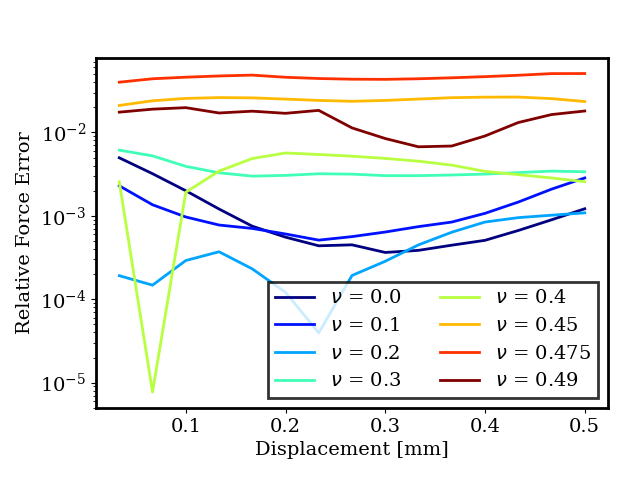}
		\caption{}
		\label{fig:neohookean_different_poissons_ratios_force_realtive_error}
	\end{subfigure}
	\caption{PINN training results for a Neo-Hookean model with various Poisson's ratios. (a) The total energy functional of the various Poisson ratio runs during training. (b) The total residuals of the various Poisson ratio runs during training. (b) Comparison of FEM and PINN force displacement curves after 100,000 epochs. (c) Relative error between the FEM and PINN displacement fields after 100,000 epochs. (d) The relative error between the PINN and FEM global force after 100,000 epochs.}
\label{fig:neohookean_different_poissons_ratios}
\end{figure}

To further investigate our methods ability to solve forward problems with the Neo-Hookean constitutive model, we consider the same BVP but consider the effect of varying the Poisson's ratio on the accuracy of the global force-displacement response and the local displacement error. Specifically, we fixed the shear modulus to be $\mu = 1$ MPa and calculated the accompanying bulk modulus $K$ for Poisson's ratios of $\nu = 0.0$, $0.1$, $0.2$, $0.3$, $0.4$, $0.45$, $0.475$, and $0.49$ respectively. Figure~\ref{fig:neohookean_different_poissons_ratios} shows the differences in the residual convergence, global force-displacement curve, relative full-field displacement error, and relative error in the global force-displacement curve.

As can be seen from the figures, our method is relatively insensitive to the value of Poisson's ratio except at values that approach $\nu = 0.5$, which should be expected since our element formulation does not account for incompressibility which can be considered in future work. For all values of $\nu$ considered, the energy plateaus after about 1,000 epochs as in the previous example, but the residual of the internal force vector continues to decrease several decades to balance forces. Except for values of $\nu \ge 0.45$, our method generally predicts the displacement field and global force-displacement response to roughly 1-2\% or less relative error as can bee seen in Figures~\ref{fig:neohookean_different_poissons_ratios_displacement_relative_error} and~\ref{fig:neohookean_different_poissons_ratios_force_realtive_error}. Interestingly, the displacement relative error is higher for lower levels of global displacement, which could be an indication that time steps with large global displacements may be dominating the loss function.

\begin{figure}
	\centering
	\begin{subfigure}{0.4\textwidth}
		\centering
		\includegraphics[width=\textwidth]{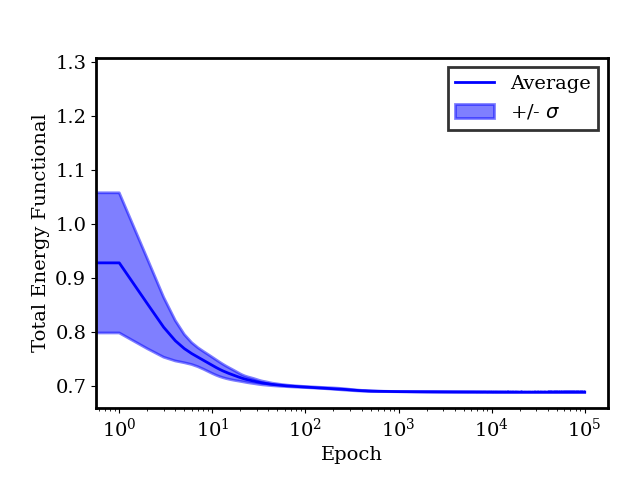}
		\caption{}
	\end{subfigure}
	\begin{subfigure}{0.4\textwidth}
		\centering
		\includegraphics[width=\textwidth]{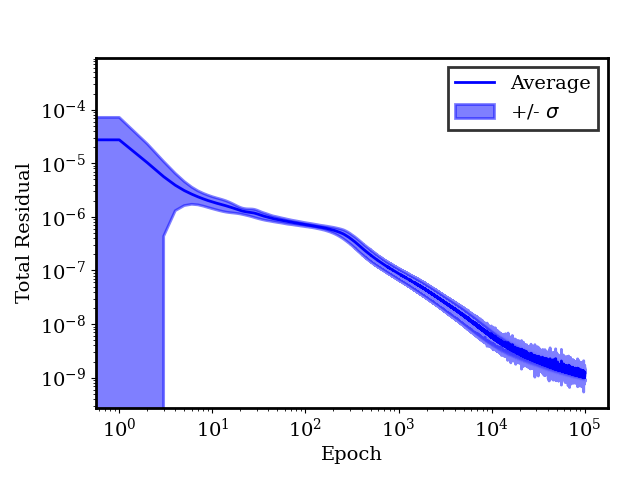}
		\caption{}
	\end{subfigure}
	\begin{subfigure}{0.4\textwidth}
		\centering
		\includegraphics[width=\textwidth]{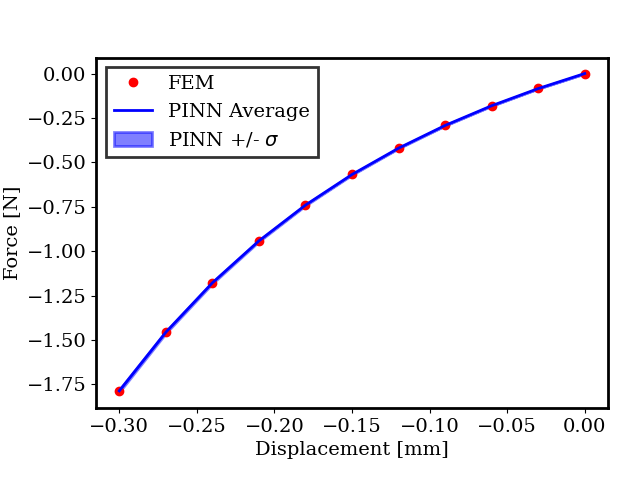}
		\caption{}
	\end{subfigure}
	\caption{Results for 8 separate training runs for a Blatz-Ko model. (a) Total energy functional convergence over epochs. (b) Total residual convergence over epochs. (c) Comparison of global force-displacement curve between FEM results and the trained PINN after 100,000 epochs.}
\label{fig:blatz_ko_global_results}
\end{figure}

\begin{figure}
	\centering
	\begin{subfigure}{0.3\textwidth}
		\centering
		\includegraphics[width=\textwidth]{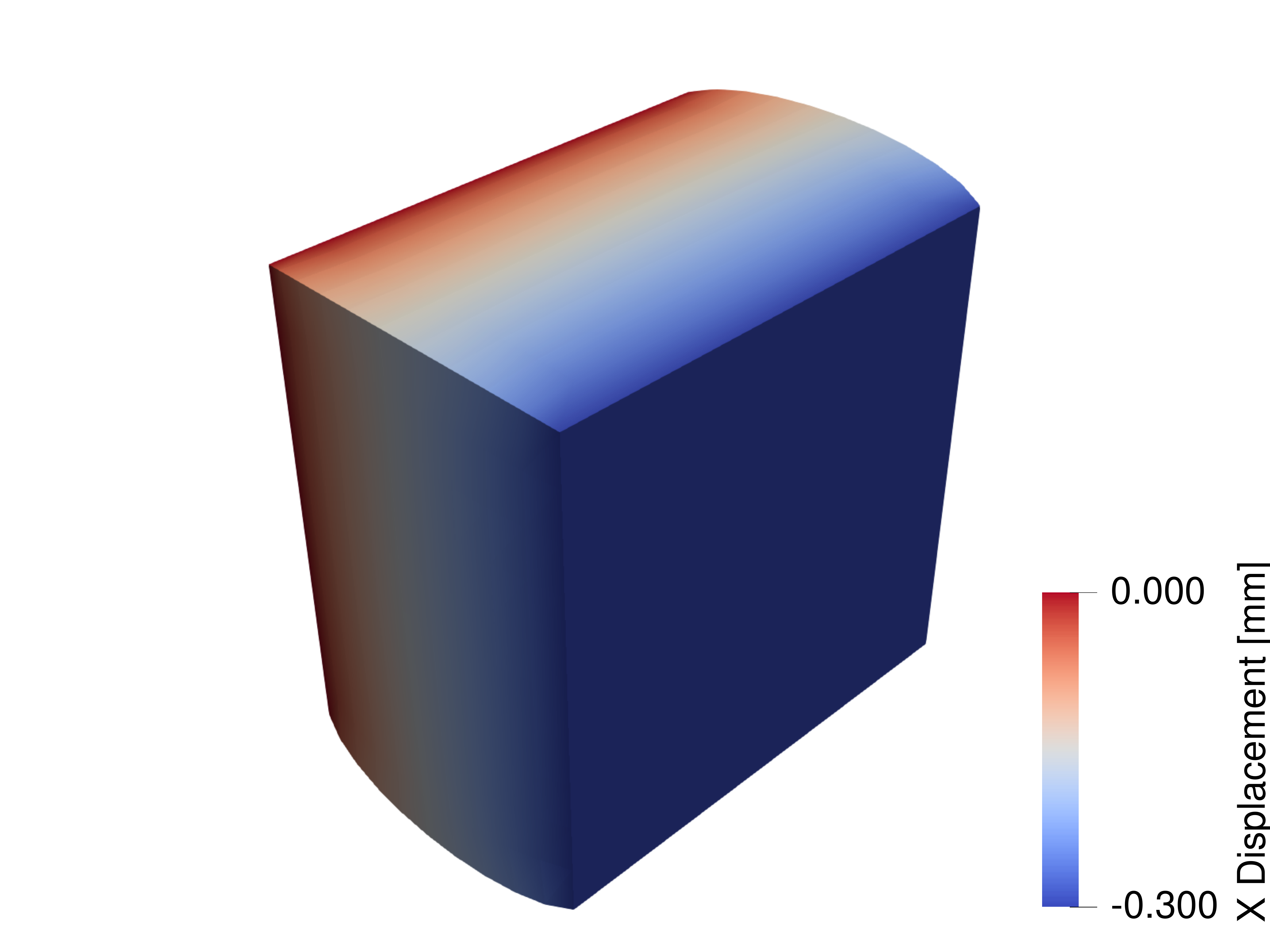}
		\caption{FEM $u_x$}
	\end{subfigure}
	\begin{subfigure}{0.3\textwidth}
		\centering
		\includegraphics[width=\textwidth]{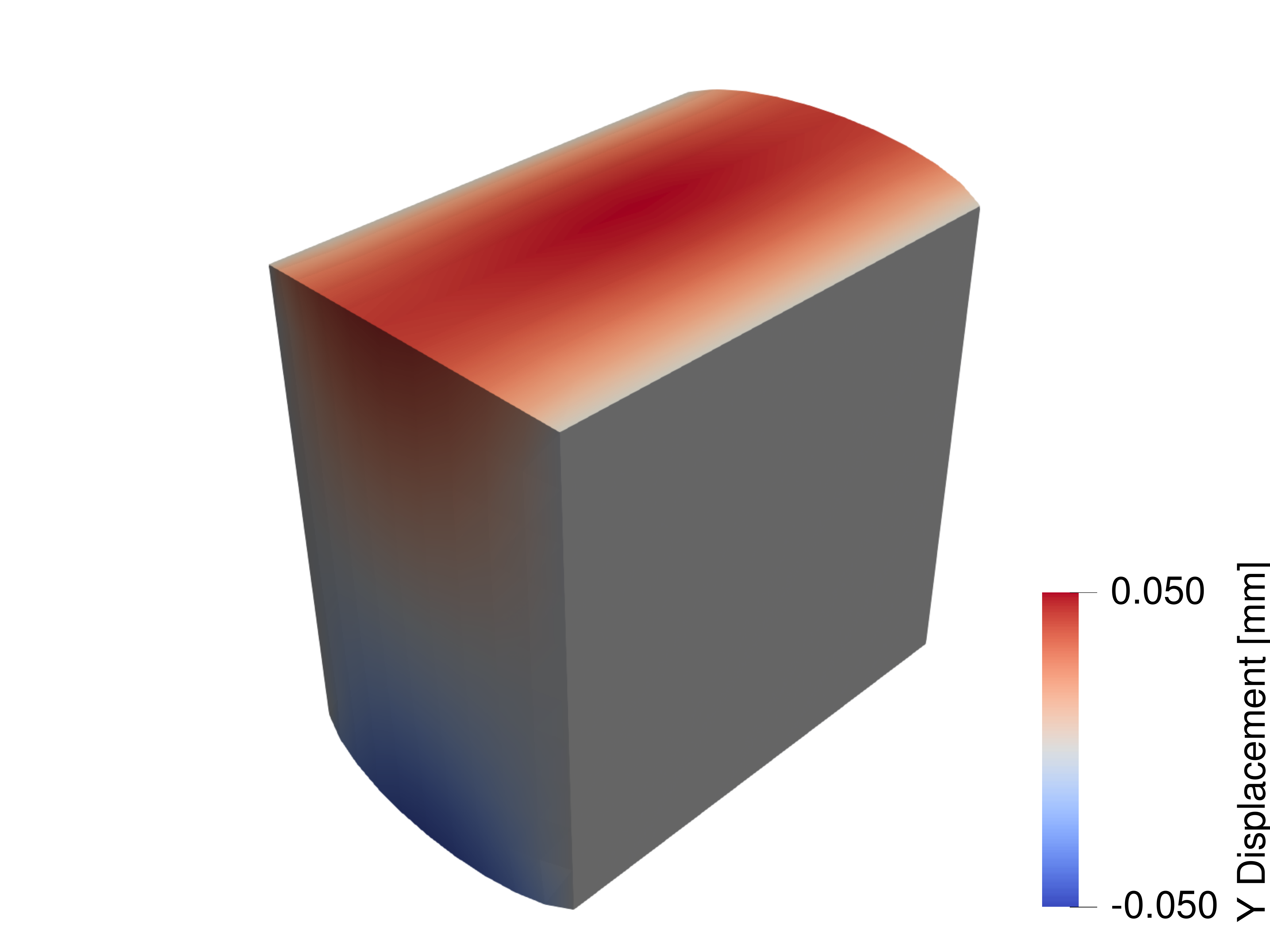}
		\caption{FEM $u_y$}
	\end{subfigure}
	\begin{subfigure}{0.3\textwidth}
		\centering
		\includegraphics[width=\textwidth]{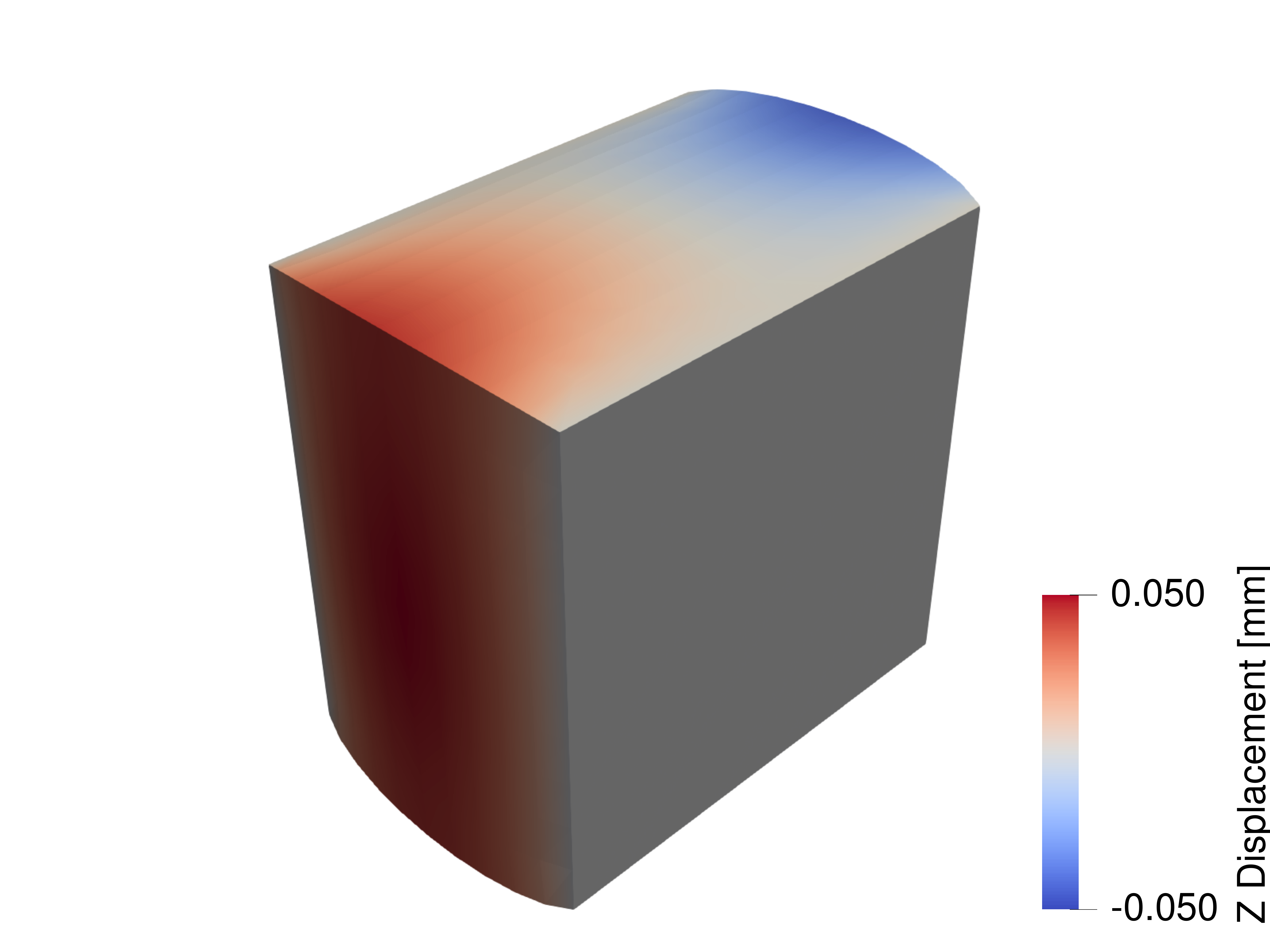}
		\caption{FEM $u_z$}
	\end{subfigure}
	\begin{subfigure}{0.3\textwidth}
		\centering
		\includegraphics[width=\textwidth]{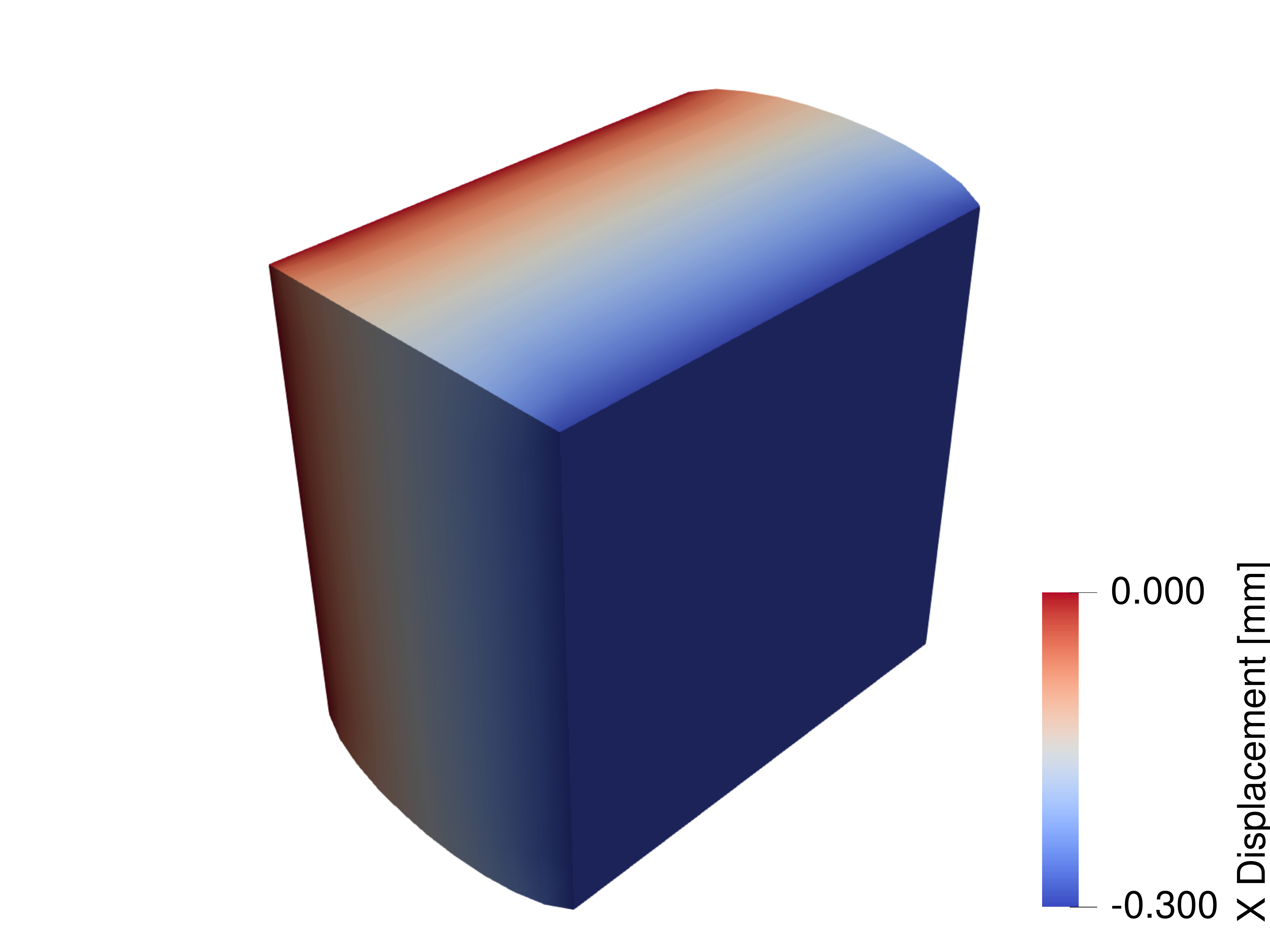}
		\caption{PINN $u_x$}
	\end{subfigure}
	\begin{subfigure}{0.3\textwidth}
		\centering
		\includegraphics[width=\textwidth]{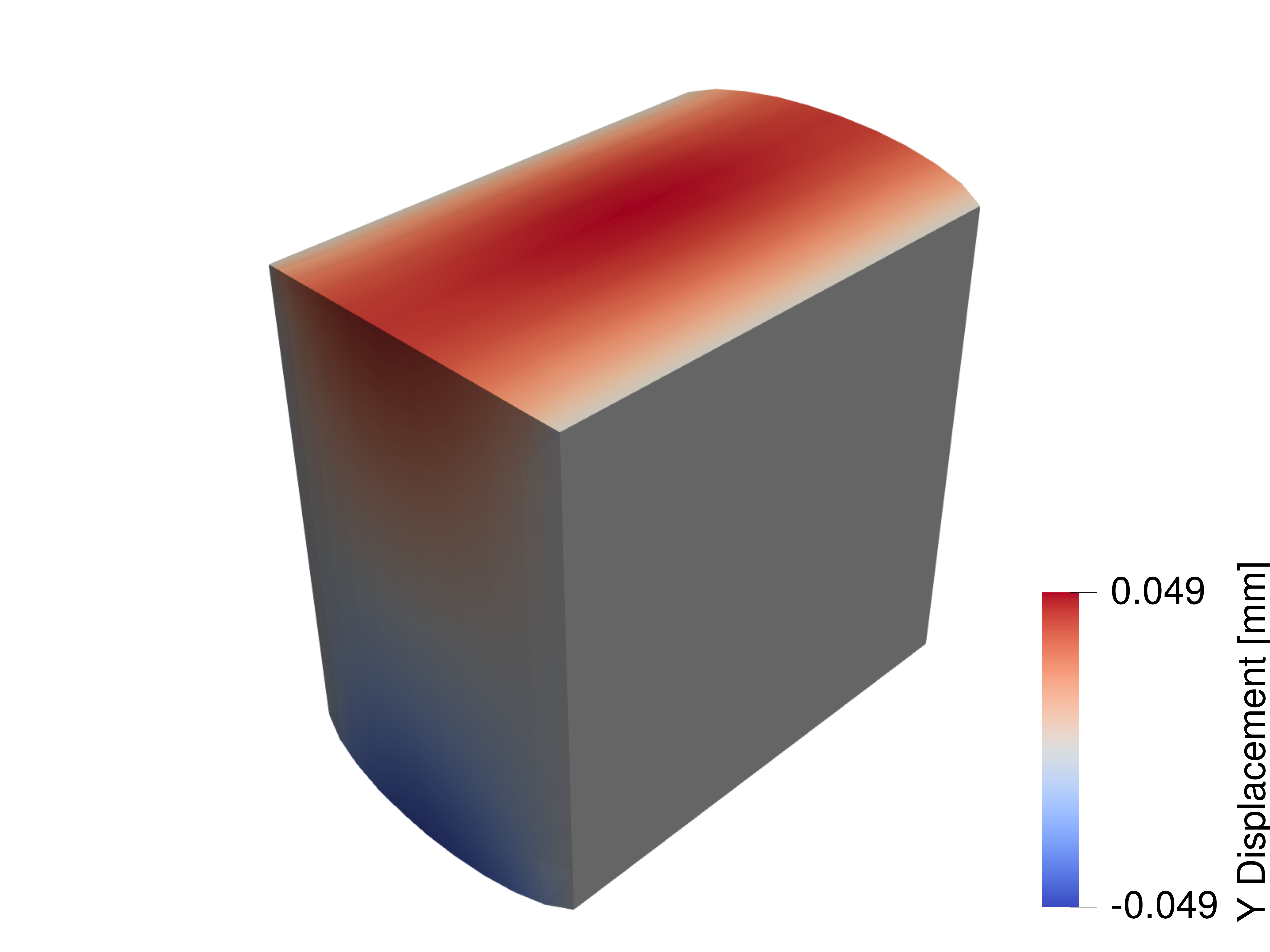}
		\caption{PINN $u_y$}
	\end{subfigure}
	\begin{subfigure}{0.3\textwidth}
		\centering
		\includegraphics[width=\textwidth]{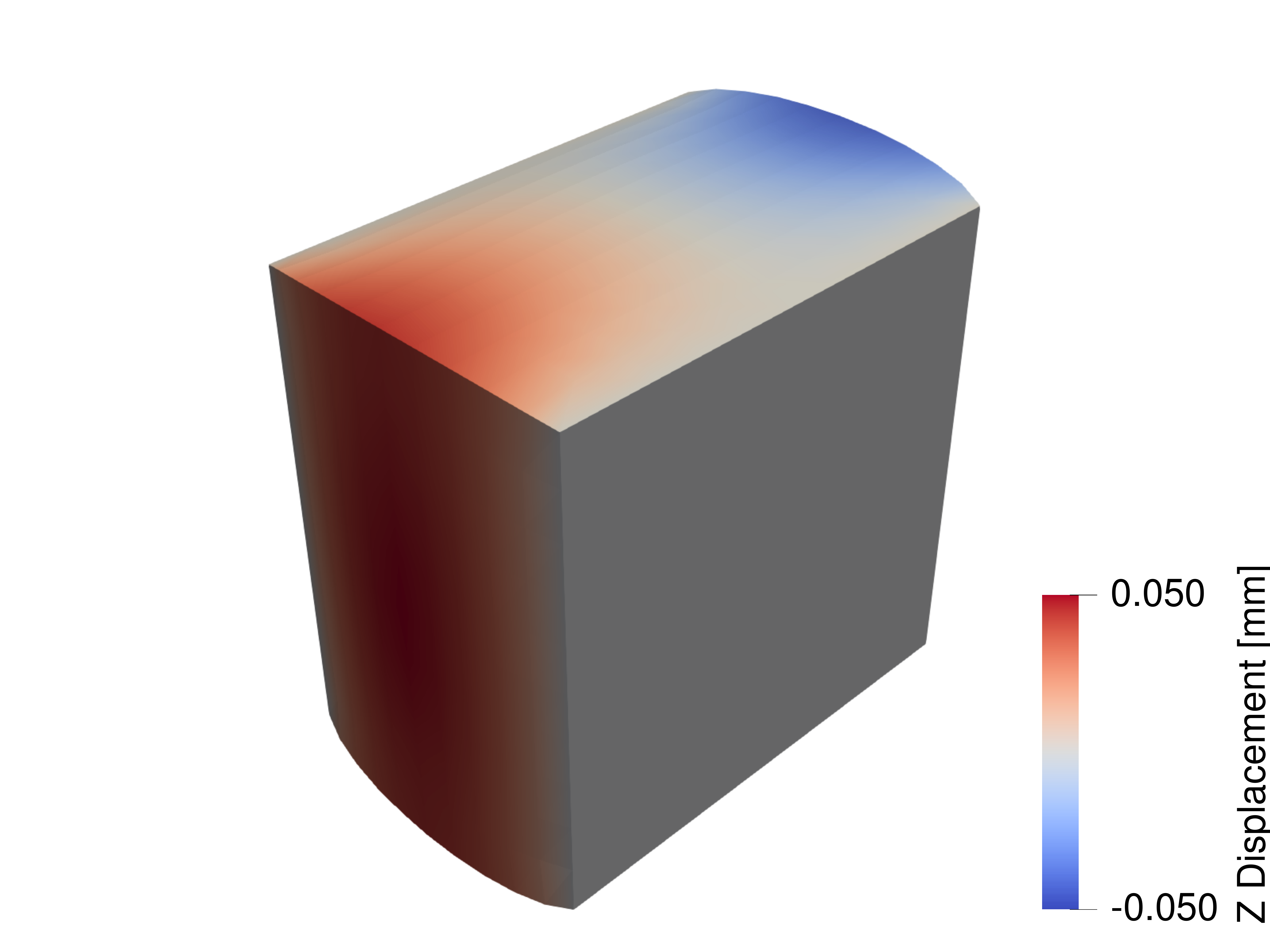}
		\caption{PINN $u_z$}
	\end{subfigure}
	\caption{Comparison of displacement field components between FEM simulations and PINN training for analogous BVPs for a Blatz-Ko model with a shear modulus of $\mu = 1.0$MPa}
\label{fig:blatz_ko_displacement}
\end{figure}

\begin{figure}
	\centering
	\begin{subfigure}{0.3\textwidth}
		\centering
		\includegraphics[width=\textwidth]{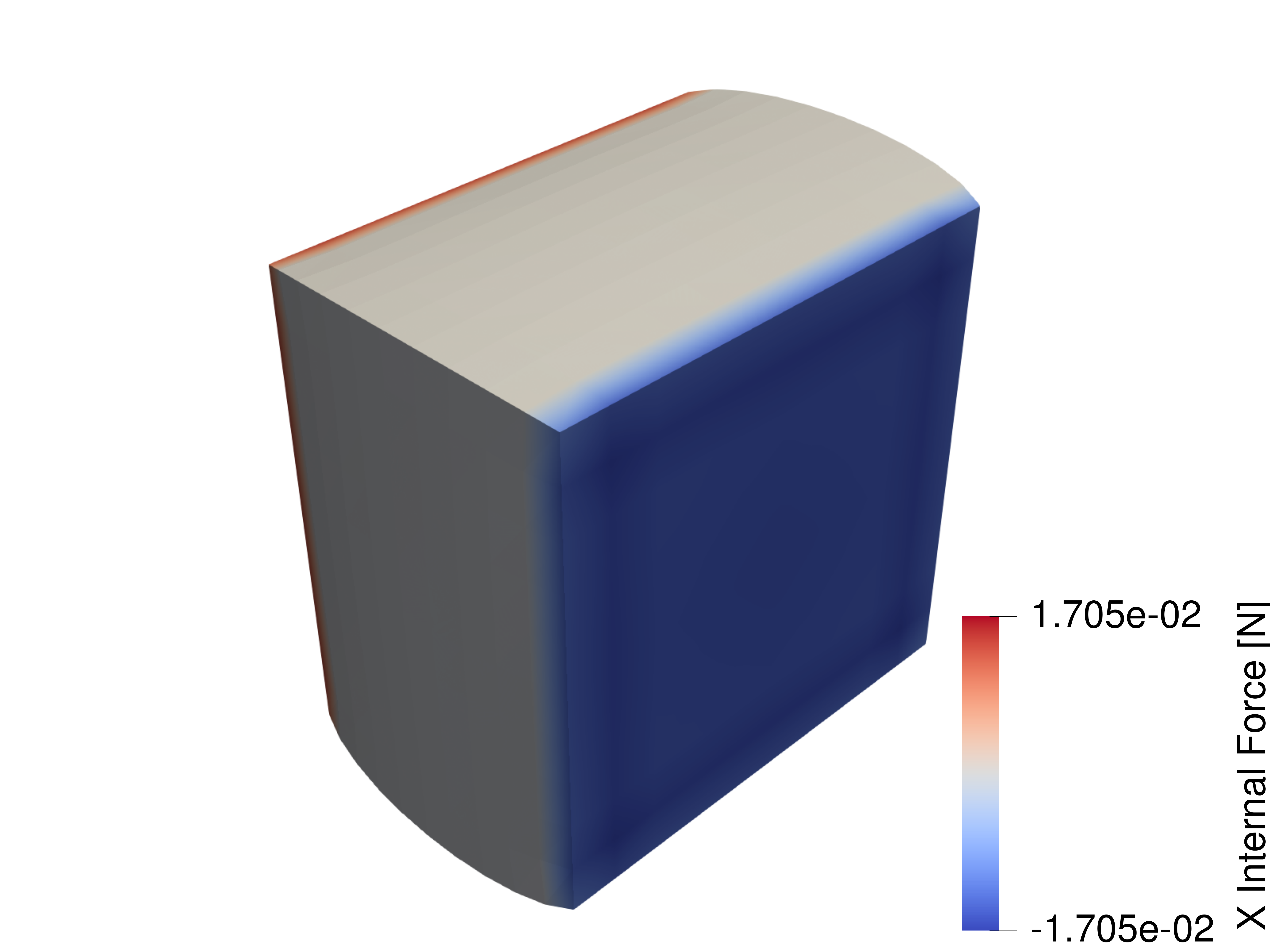}
		\caption{FEM $f_x$}
	\end{subfigure}
	\begin{subfigure}{0.3\textwidth}
		\centering
		\includegraphics[width=\textwidth]{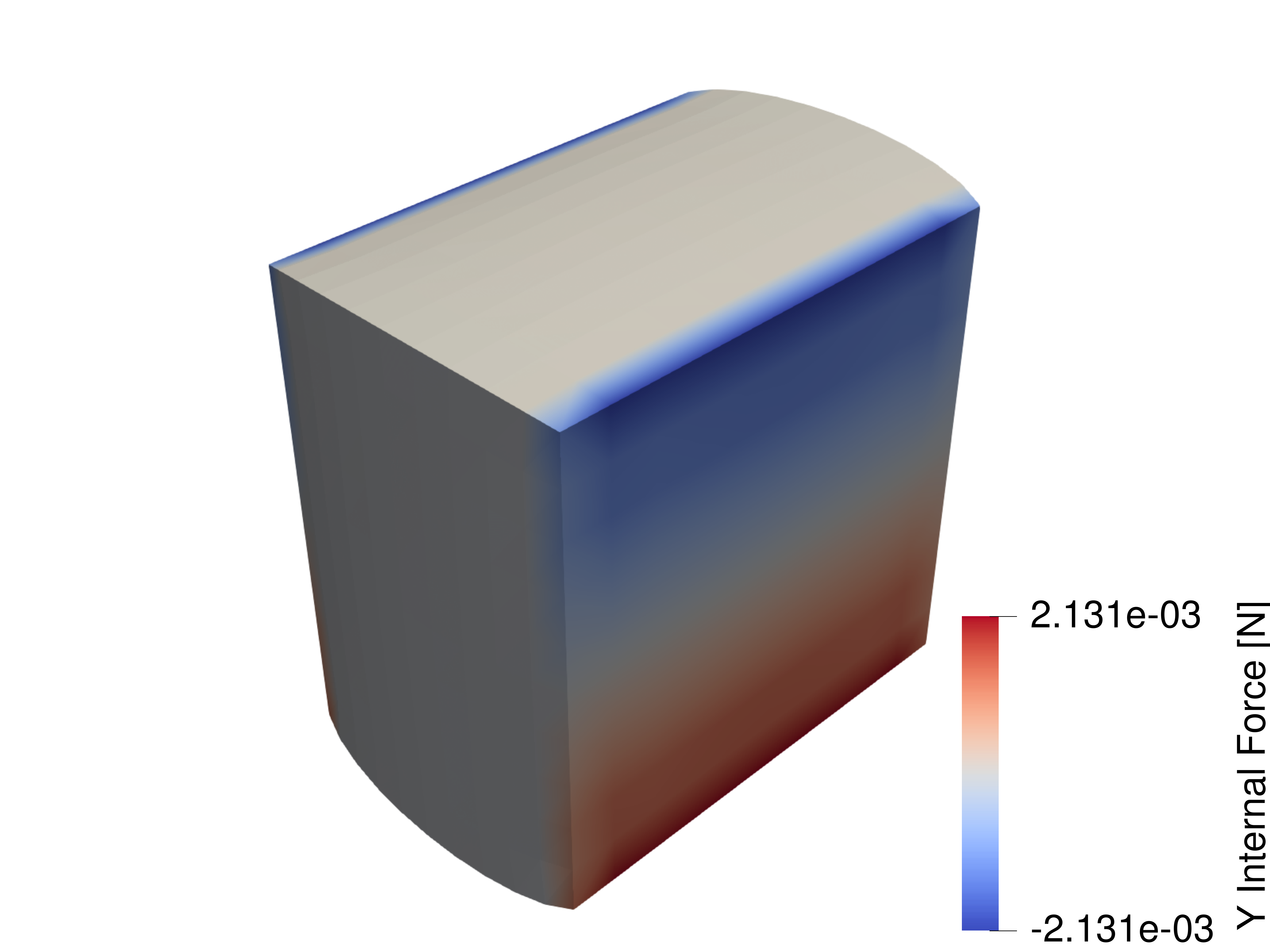}
		\caption{FEM $f_y$}
	\end{subfigure}
	\begin{subfigure}{0.3\textwidth}
		\centering
		\includegraphics[width=\textwidth]{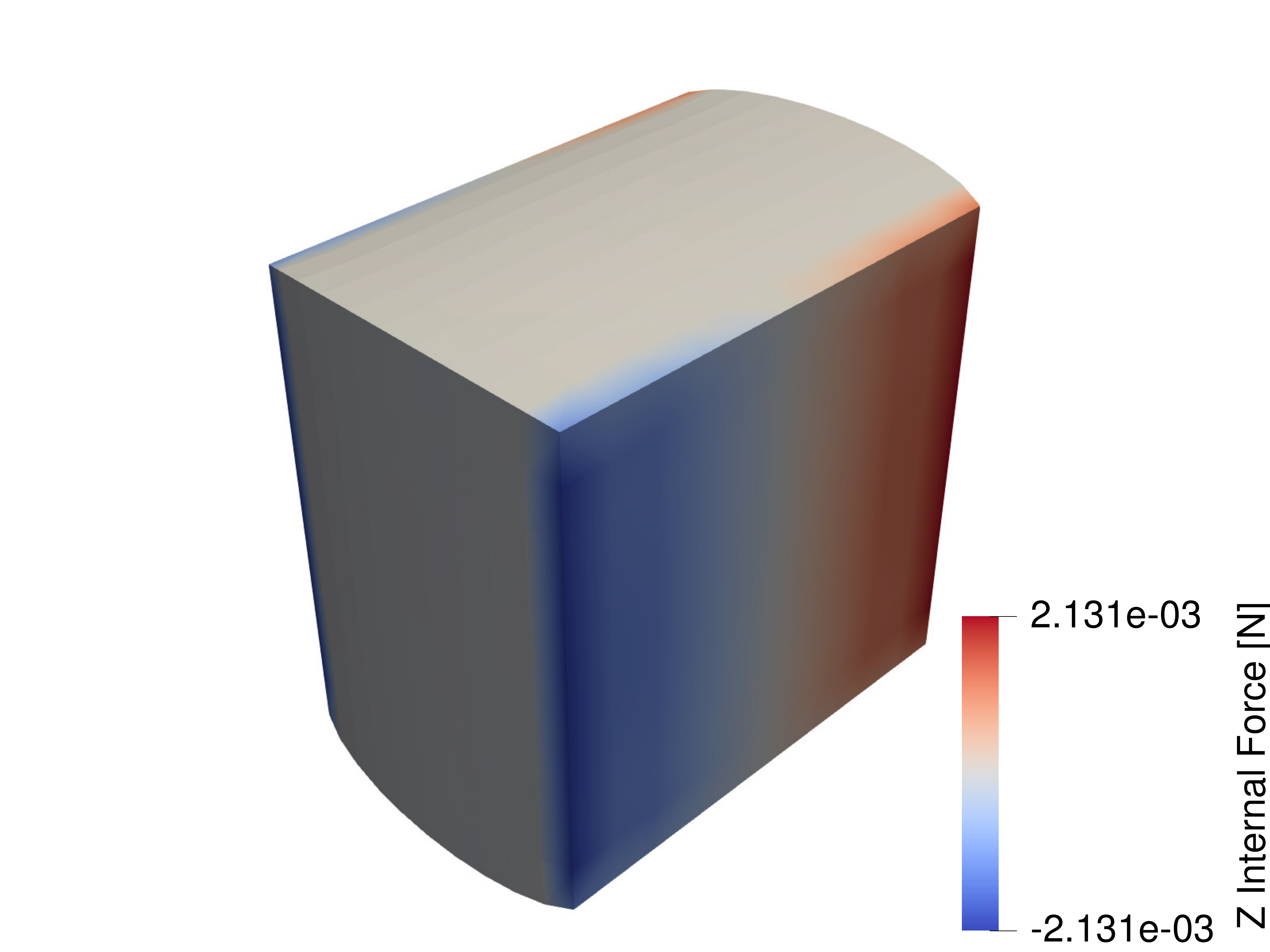}
		\caption{FEM $f_z$}
	\end{subfigure}
	\begin{subfigure}{0.3\textwidth}
		\centering
		\includegraphics[width=\textwidth]{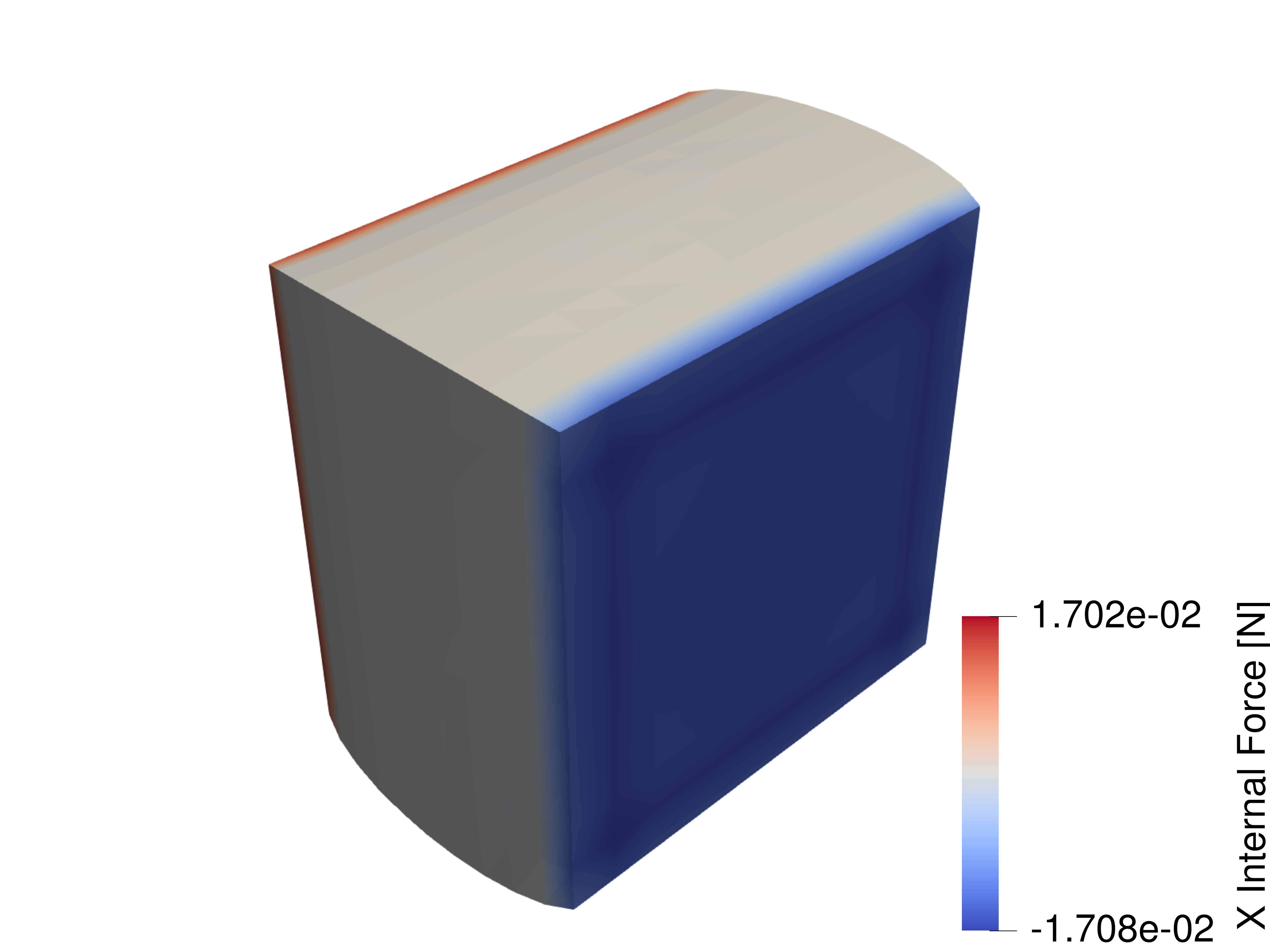}
		\caption{PINN $f_x$}
	\end{subfigure}
	\begin{subfigure}{0.3\textwidth}
		\centering
		\includegraphics[width=\textwidth]{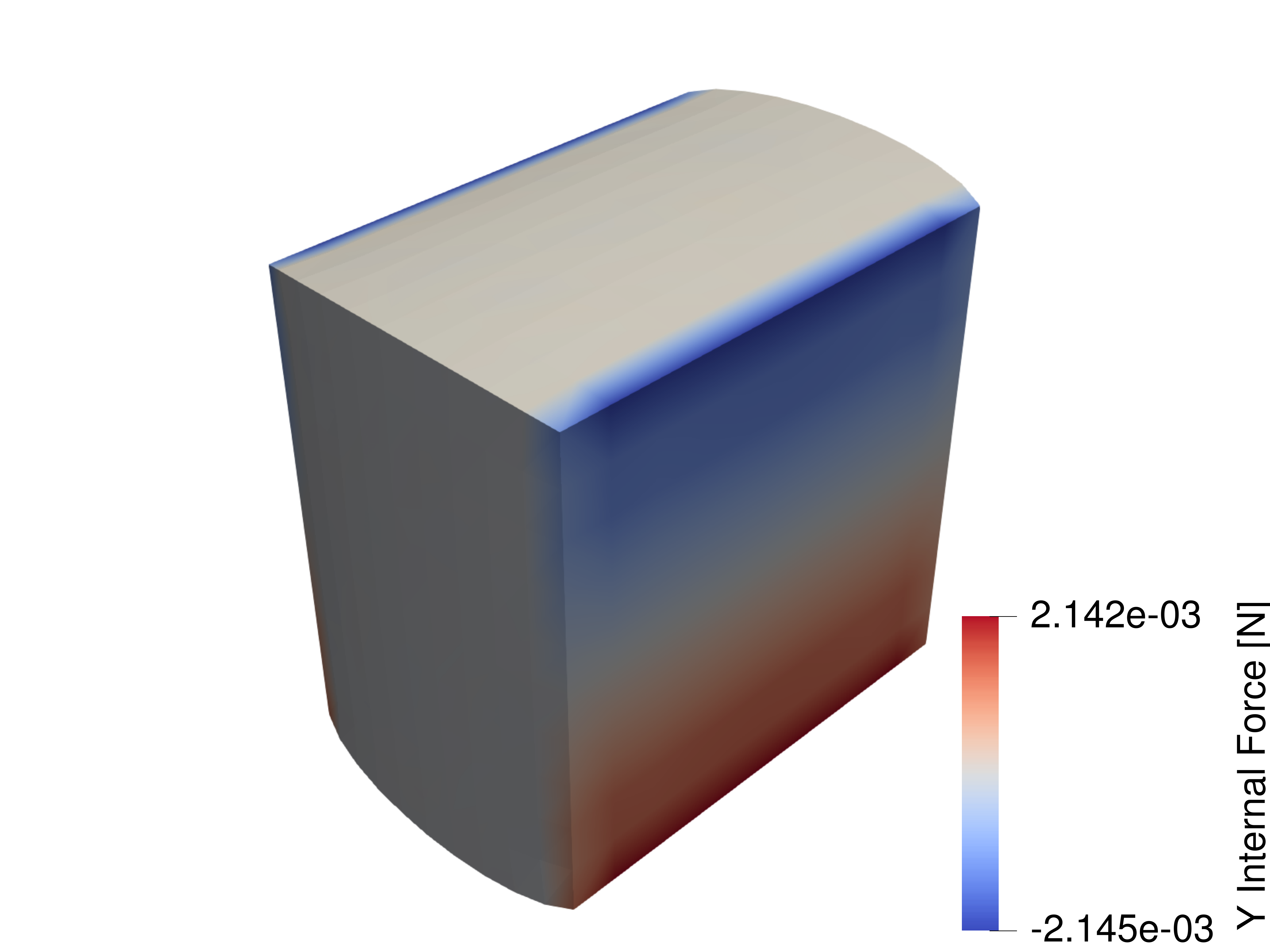}
		\caption{PINN $f_y$}
	\end{subfigure}
	\begin{subfigure}{0.3\textwidth}
		\centering
		\includegraphics[width=\textwidth]{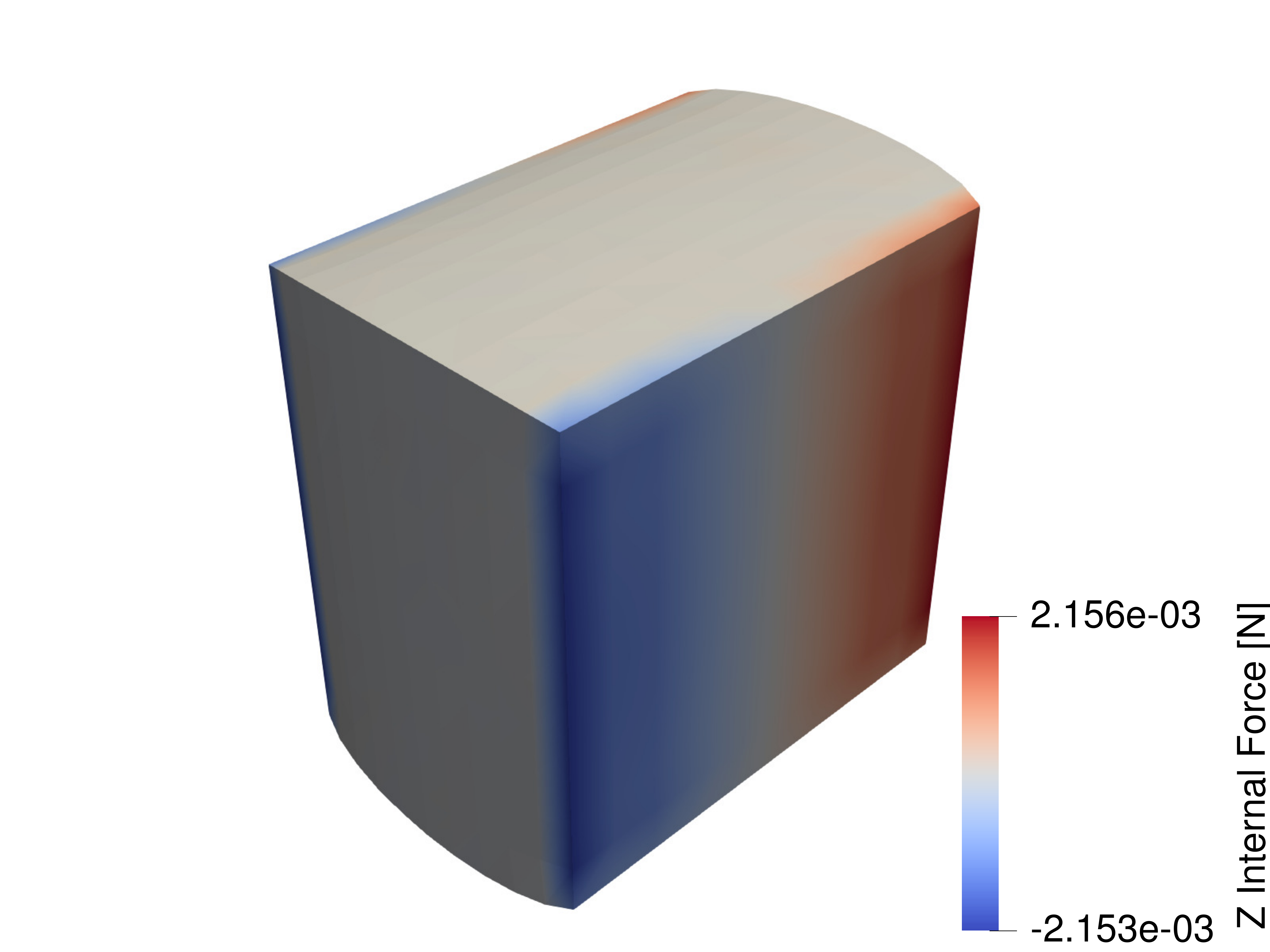}
		\caption{PINN $f_z$}
	\end{subfigure}
	\caption{Comparison of internal force field components between FEM simulations and PINN training for analogous BVPs for a Blatz-Ko model with a shear modulus of $\mu = 1.0$MPa}
\label{fig:blatz_ko_internal_force}
\end{figure}

Our next benchmark problem consists of a simple cube of length 1mm in uniaxial compression with a Blatz-Ko constitutive model with a shear modulus of $\mu = 1$ MPa driving the problem. We utilized ten time steps to achieve a total global displacement of -0.3mm or -30\% nominal strain. The geometry was meshed with 1000 Hex8 finite elements which corresponds to 80,000 quadrature point evaluations per epoch of training. Eight separate PINNs were trained to show the repeatability of the method due to the stochasticity of the NN optimizer and each PINN was trained for 100,000 epochs. The evolution of the PINN residual during training and the comparison of the PINN global force-displacement to FEM are shown in Figure~\ref{fig:blatz_ko_global_results}.
Similar to the first benchmark problem, our method is also very repeatable in terms of the energy and residual convergence and global force-displacement prediction when compared to FEM. Like the other cases, the total energy functional plateaus very quickly during training, but many more epochs are needed to properly balance the internal force vector. To further see the performance of our method for this particular model, the displacement field at maximum global strain for both the FEM solution and one of the trained PINNs after 100,000 epochs are shown in Figure~\ref{fig:blatz_ko_displacement}.
As can be seen from the comparisons of the displacement fields, our PINN does a reasonably accurate job of predicting the full-field displacement field when compared to FEM with the largest deviation interestingly being seen in the $y$ component of the displacement. To show that our PINN solution's forces are in balance, we compare the three components of the internal force field of the FEM solve and one of the trained PINNs after 100,000 epochs at maximum global strain in Figure~\ref{fig:blatz_ko_internal_force}.
Although very small deviations of the reaction forces are seen between the trained PINN and the FEM solution, the PINN is doing a quite reasonable job of predicting the reactions forces up to a few decimal places of the FEM solution.
 
\begin{figure}
	\centering
	\begin{subfigure}{0.4\textwidth}
		\centering
		\includegraphics[width=\textwidth]{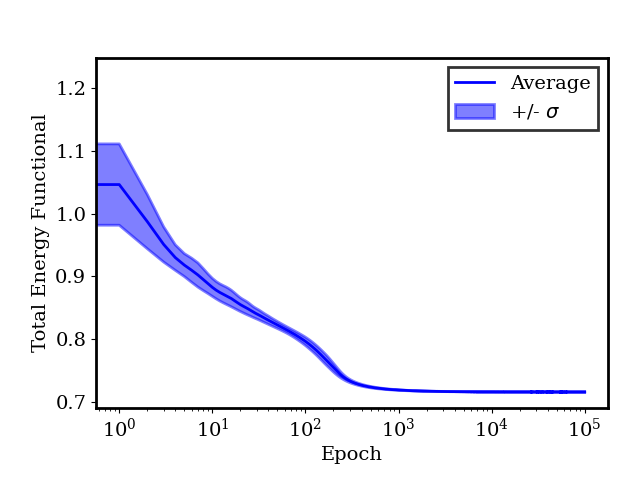}
		\caption{}
	\end{subfigure}
	\begin{subfigure}{0.4\textwidth}
		\centering
		\includegraphics[width=\textwidth]{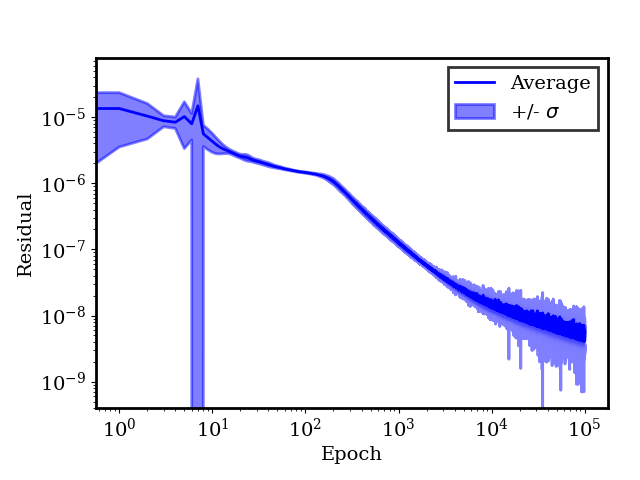}
		\caption{}
	\end{subfigure}
	\begin{subfigure}{0.4\textwidth}
		\centering
		\includegraphics[width=\textwidth]{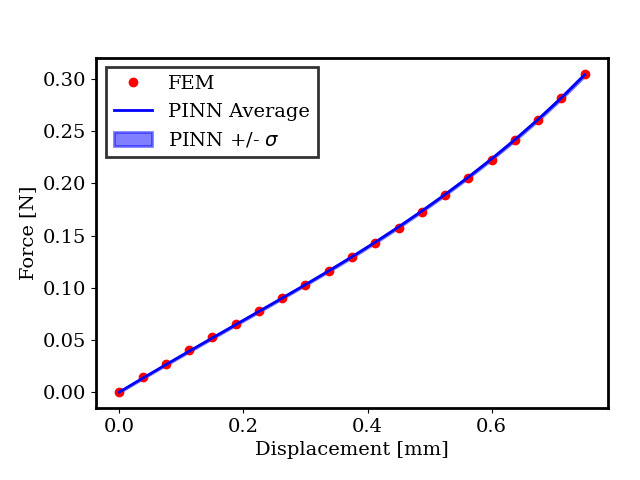}
		\caption{}
	\end{subfigure}
	\caption{Results for 8 separate training runs for a Gent model with a Poisson's ratio of $\nu = 0.3$. (a) Total energy functional convergence over epochs. (b) Total residual convergence over epochs. (c) Comparison of global force-displacement curve between FEM results and the trained PINN after 100,000 epochs.}
\label{fig:gent_poissons_ratio_0_3_global_results}
\end{figure}

\begin{figure}
	\centering
	\begin{subfigure}{0.3\textwidth}
		\centering
		\includegraphics[width=\textwidth]{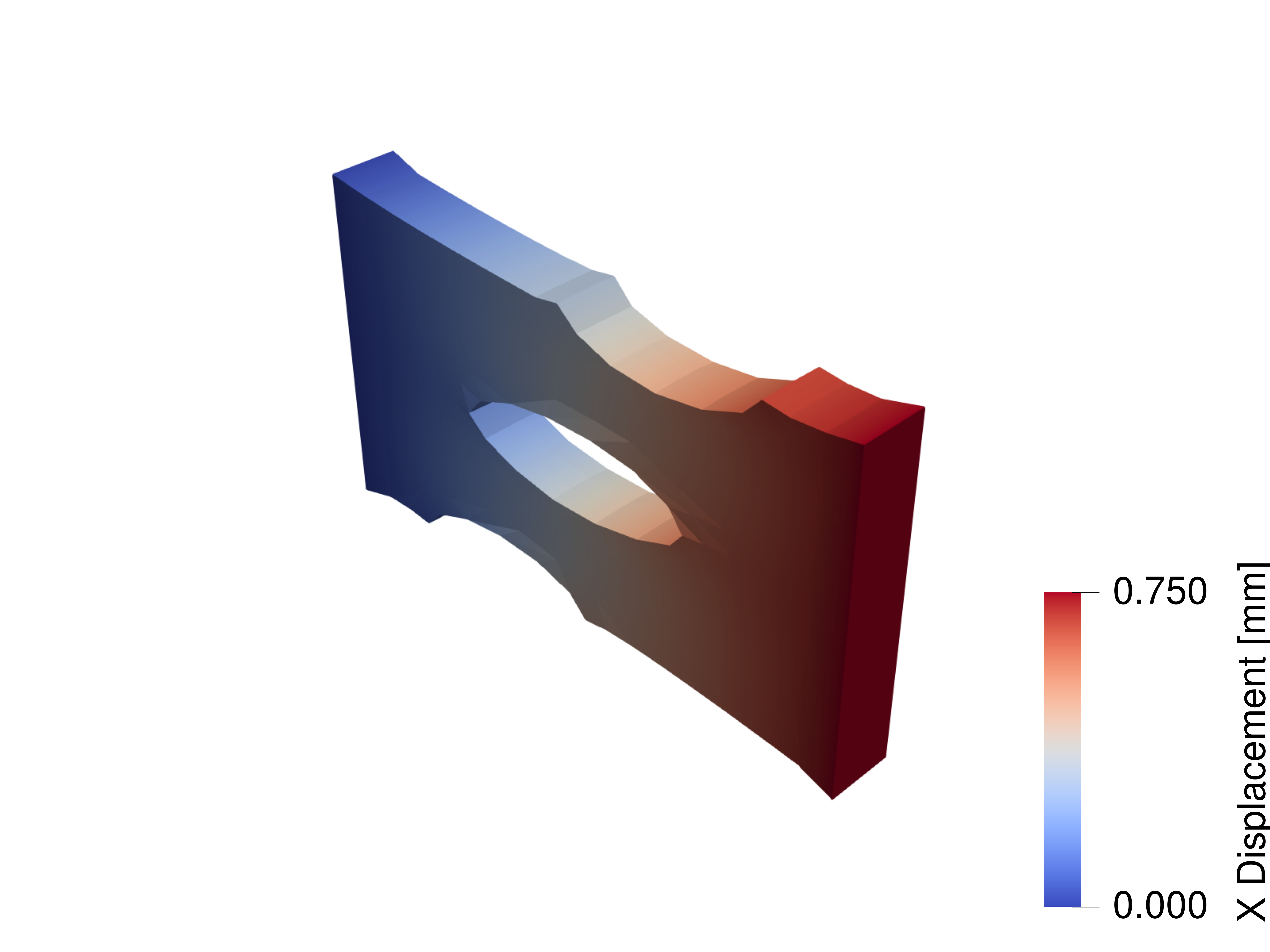}
		\caption{FEM $u_x$}
	\end{subfigure}
	\begin{subfigure}{0.3\textwidth}
		\centering
		\includegraphics[width=\textwidth]{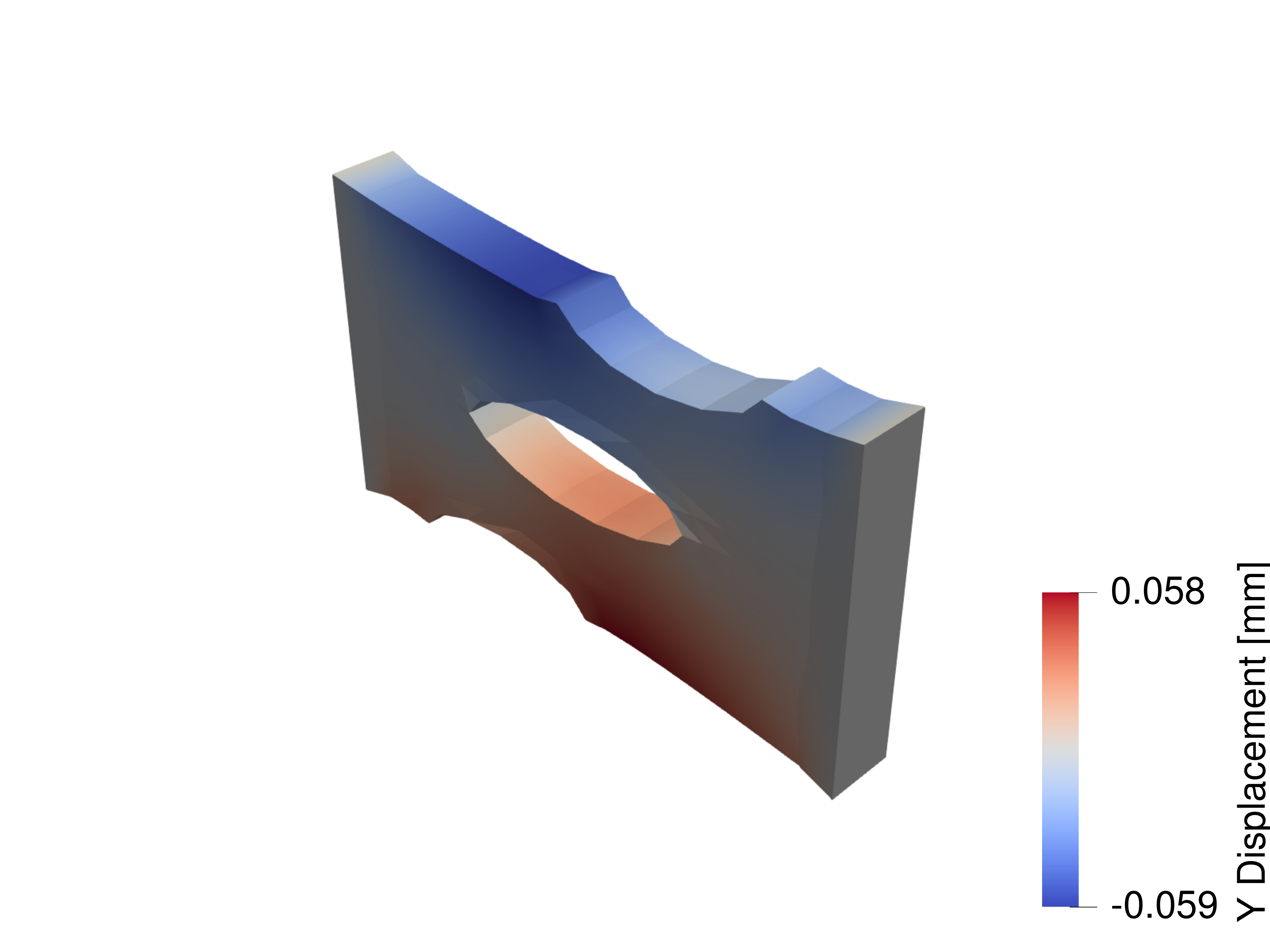}
		\caption{FEM $u_y$}
	\end{subfigure}
	\begin{subfigure}{0.3\textwidth}
		\centering
		\includegraphics[width=\textwidth]{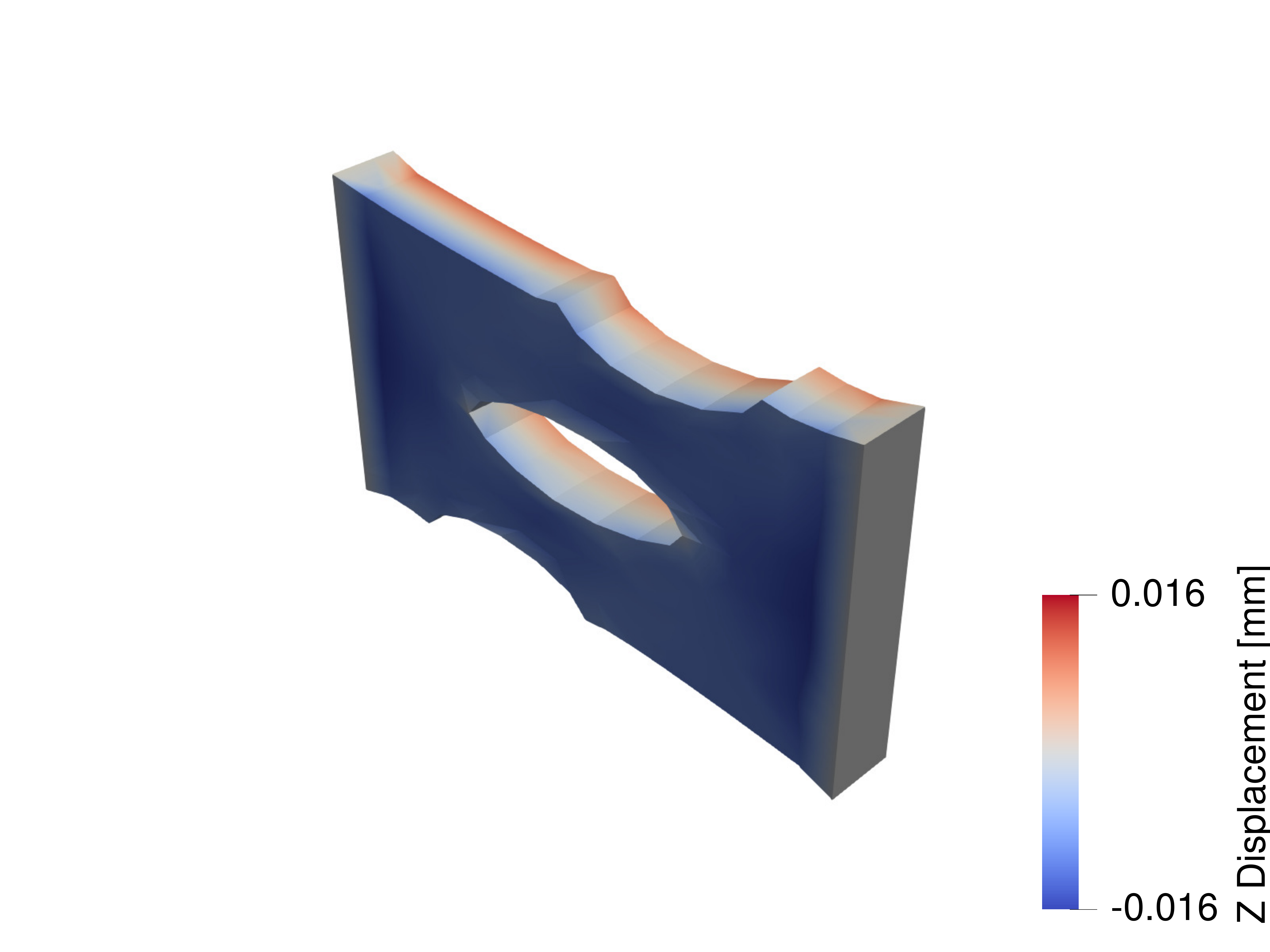}
		\caption{FEM $u_z$}
	\end{subfigure}
	\begin{subfigure}{0.3\textwidth}
		\centering
		\includegraphics[width=\textwidth]{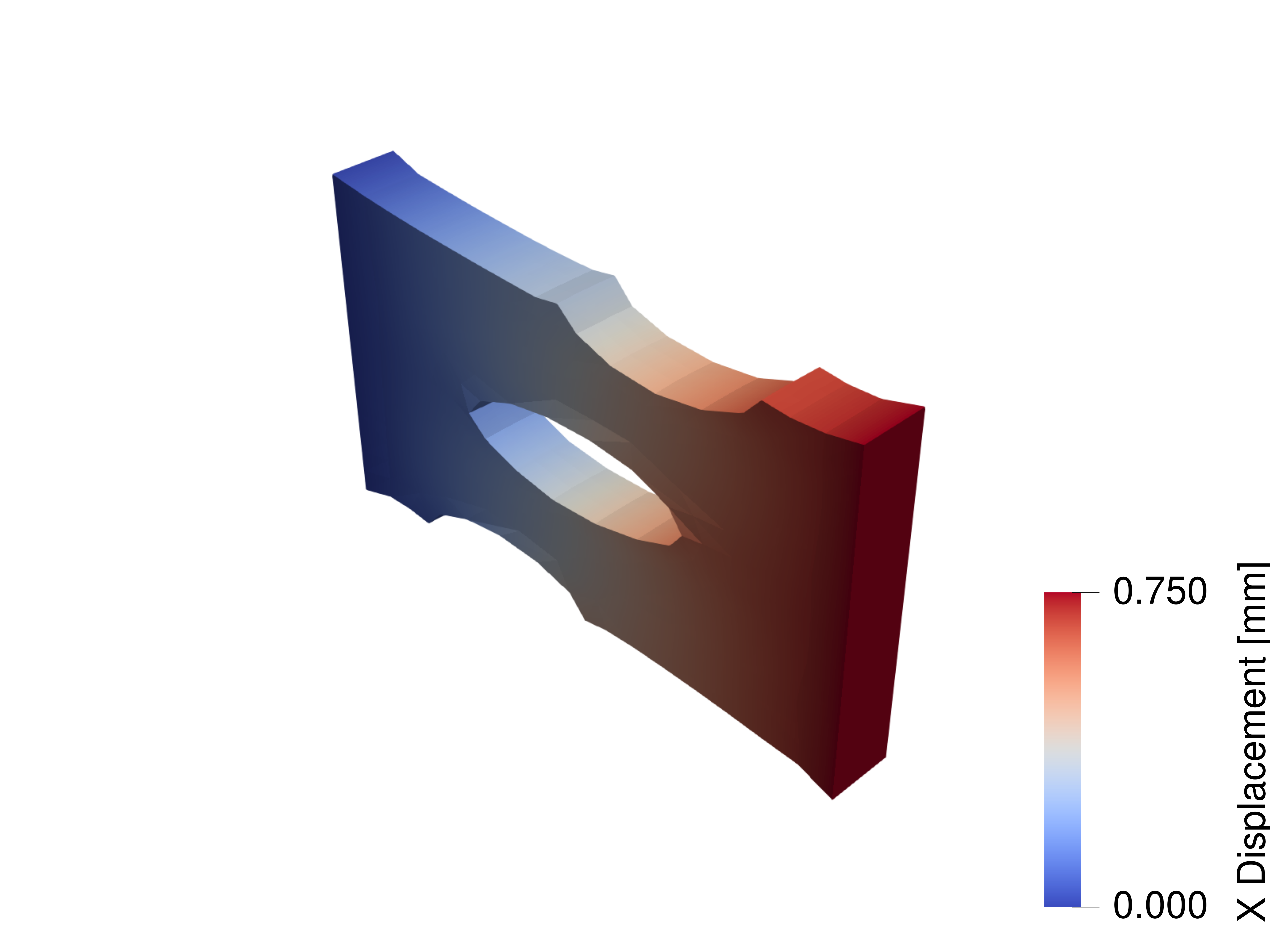}
		\caption{PINN $u_x$}
	\end{subfigure}
	\begin{subfigure}{0.3\textwidth}
		\centering
		\includegraphics[width=\textwidth]{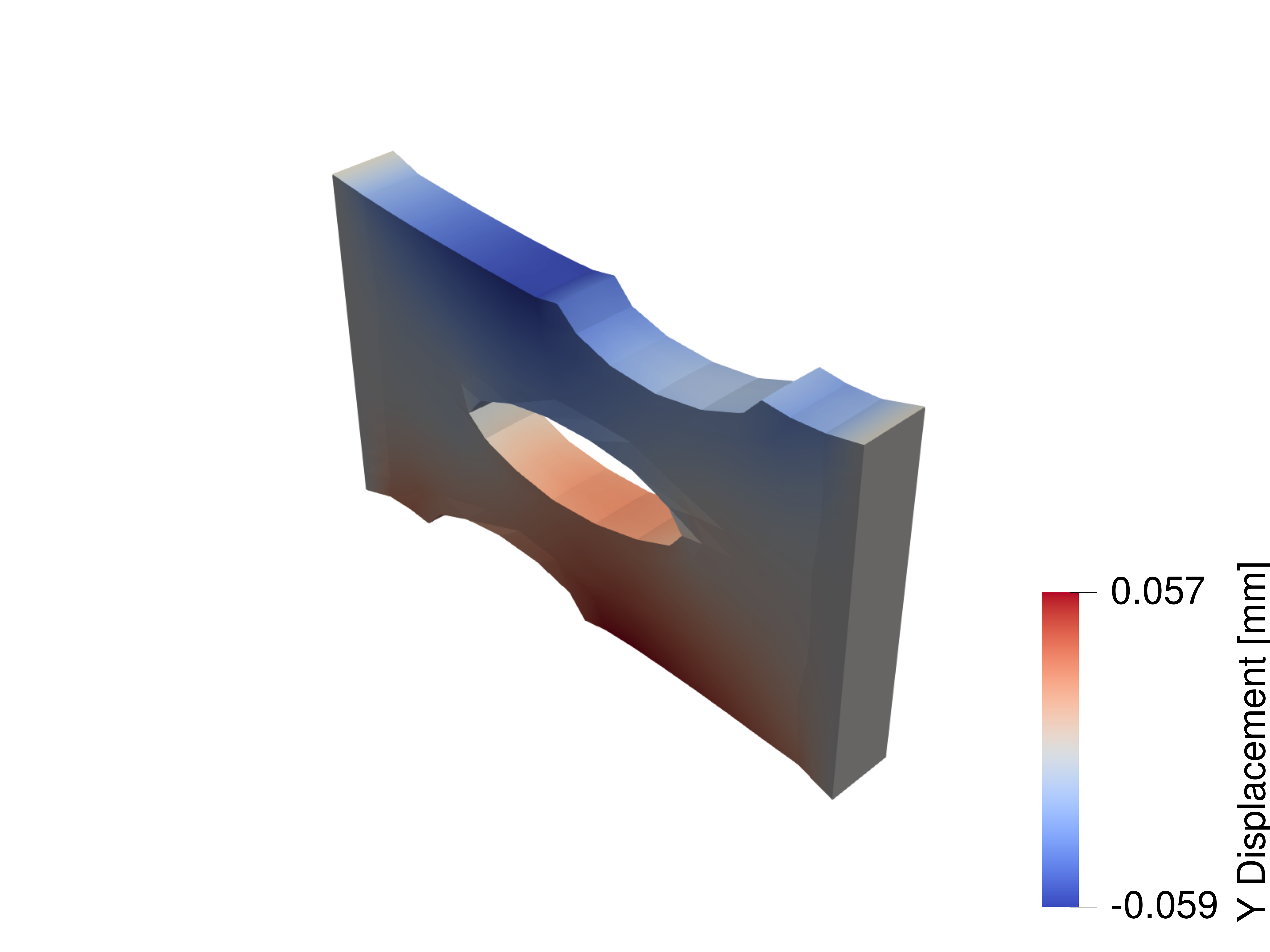}
		\caption{PINN $u_y$}
	\end{subfigure}
	\begin{subfigure}{0.3\textwidth}
		\centering
		\includegraphics[width=\textwidth]{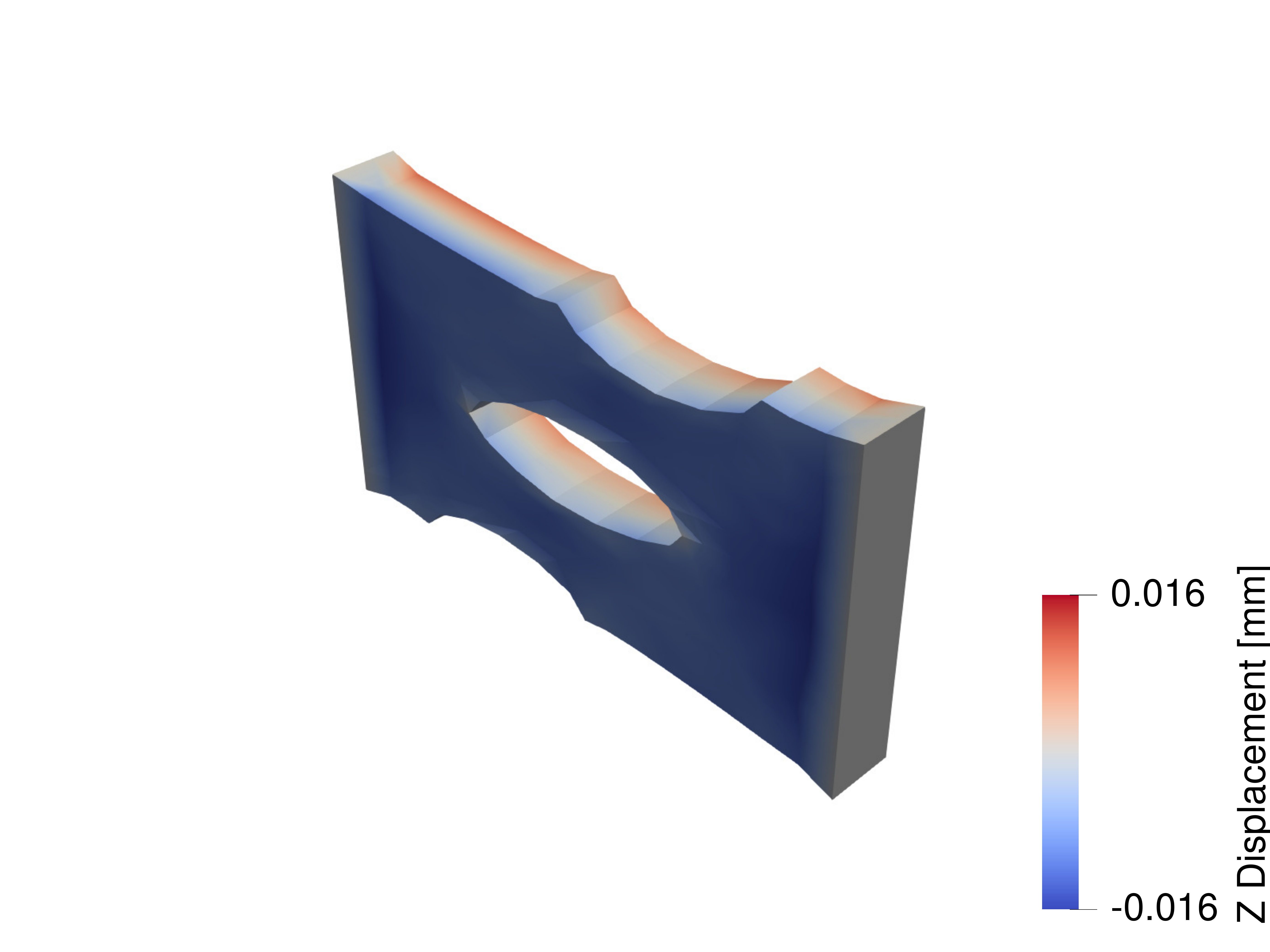}
		\caption{PINN $u_z$}
	\end{subfigure}
	\caption{Comparison of displacement field components between FEM simulations and PINN training for analogous BVPs for a Neo-Hookean model with a Poisson's ratio of $\nu = 0.3$}
\label{fig:gent_poissons_ratio_0_3_displacement}
\end{figure}

\begin{figure}
	\centering
	\begin{subfigure}{0.3\textwidth}
		\centering
		\includegraphics[width=\textwidth]{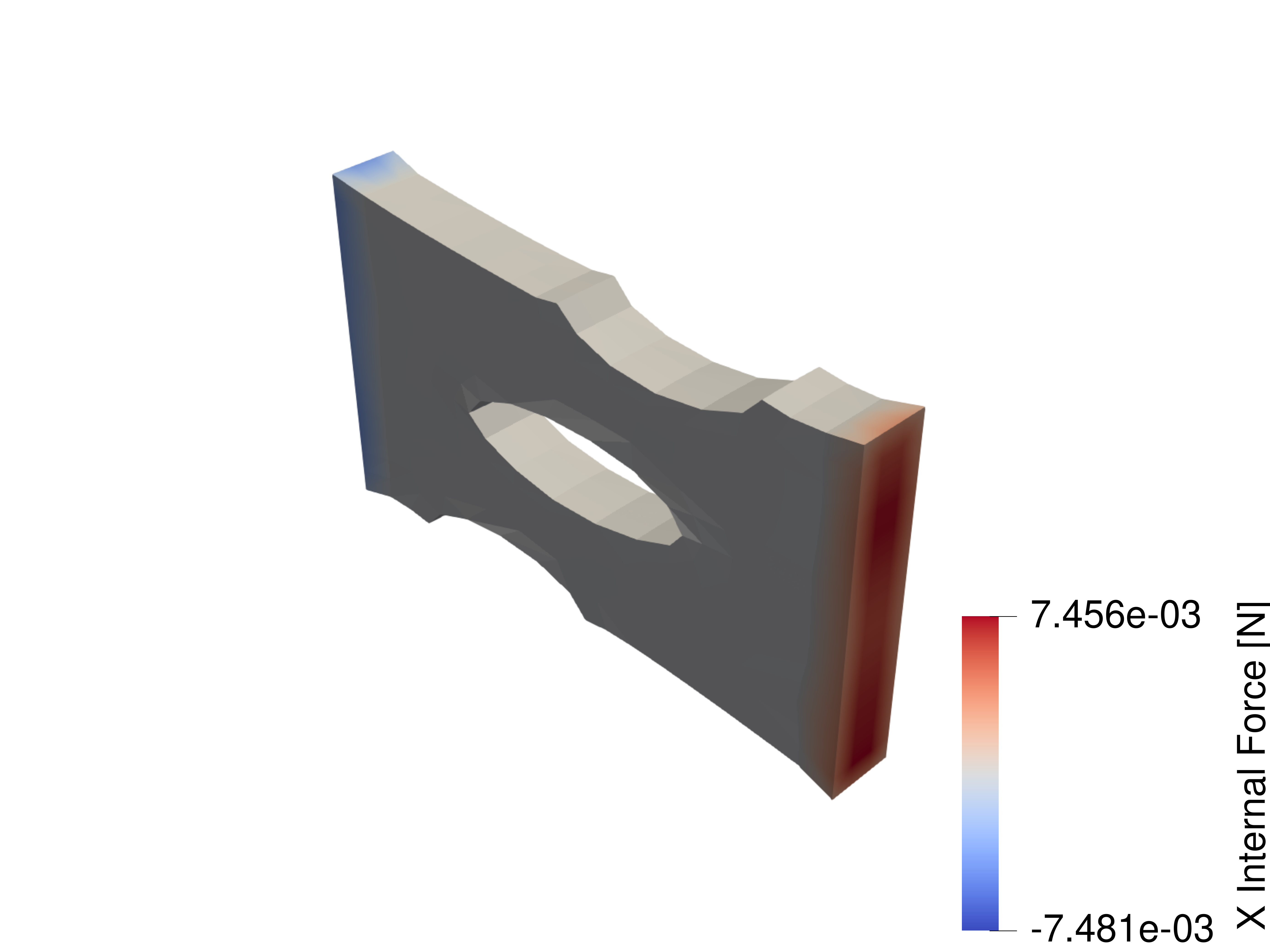}
		\caption{FEM $f_x$}
	\end{subfigure}
	\begin{subfigure}{0.3\textwidth}
		\centering
		\includegraphics[width=\textwidth]{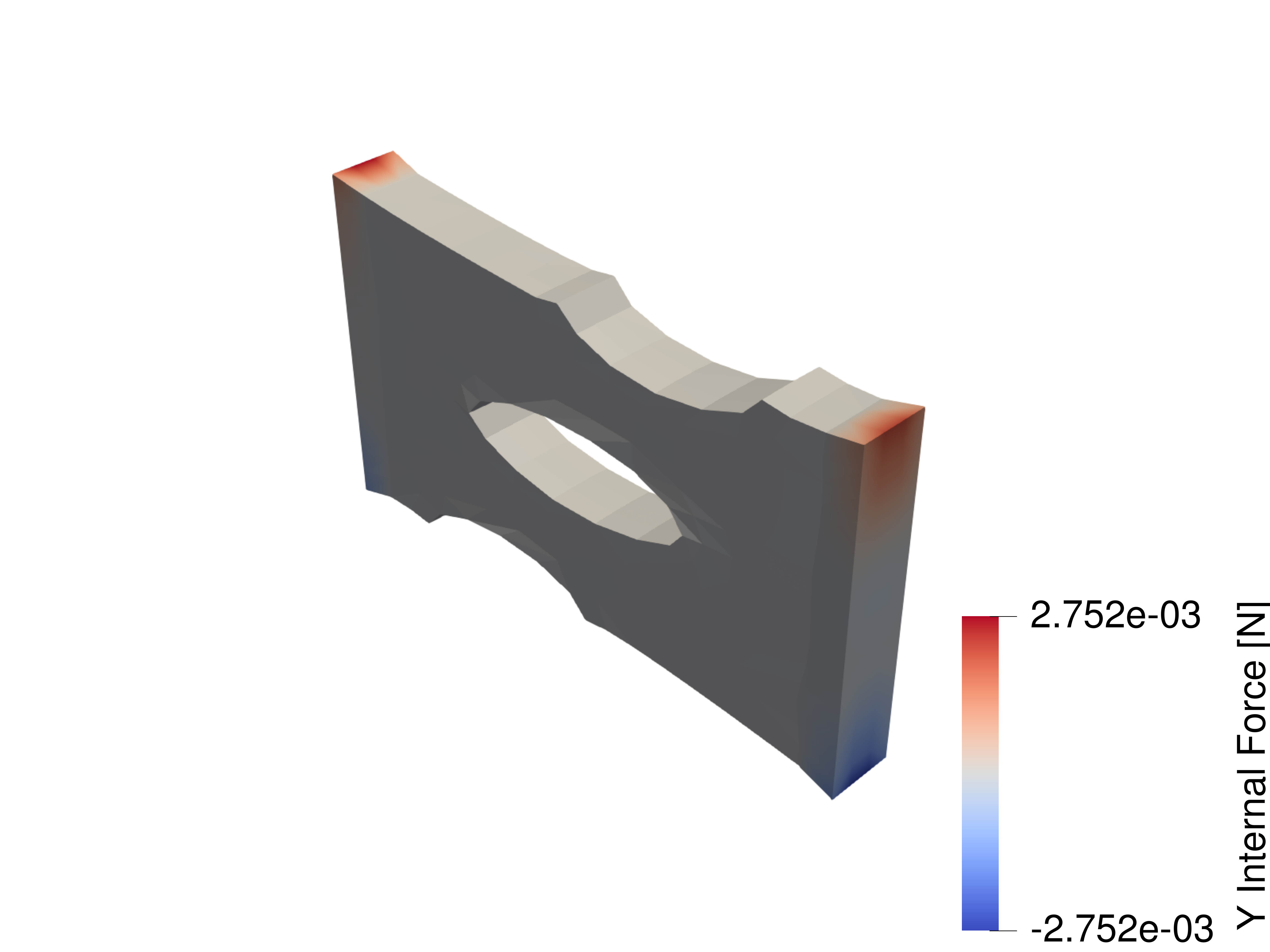}
		\caption{FEM $f_y$}
	\end{subfigure}
	\begin{subfigure}{0.3\textwidth}
		\centering
		\includegraphics[width=\textwidth]{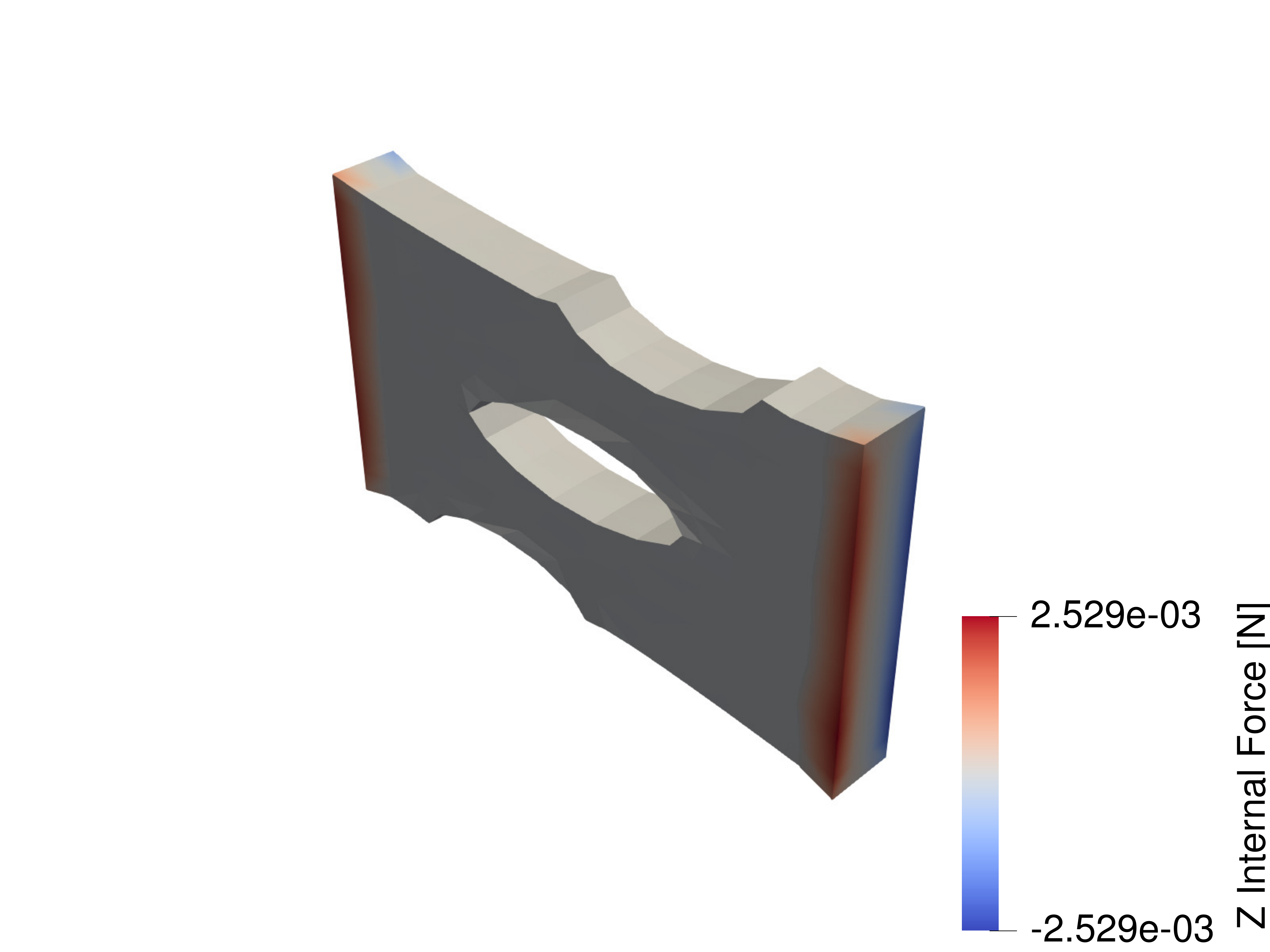}
		\caption{FEM $f_z$}
	\end{subfigure}
	\begin{subfigure}{0.3\textwidth}
		\centering
		\includegraphics[width=\textwidth]{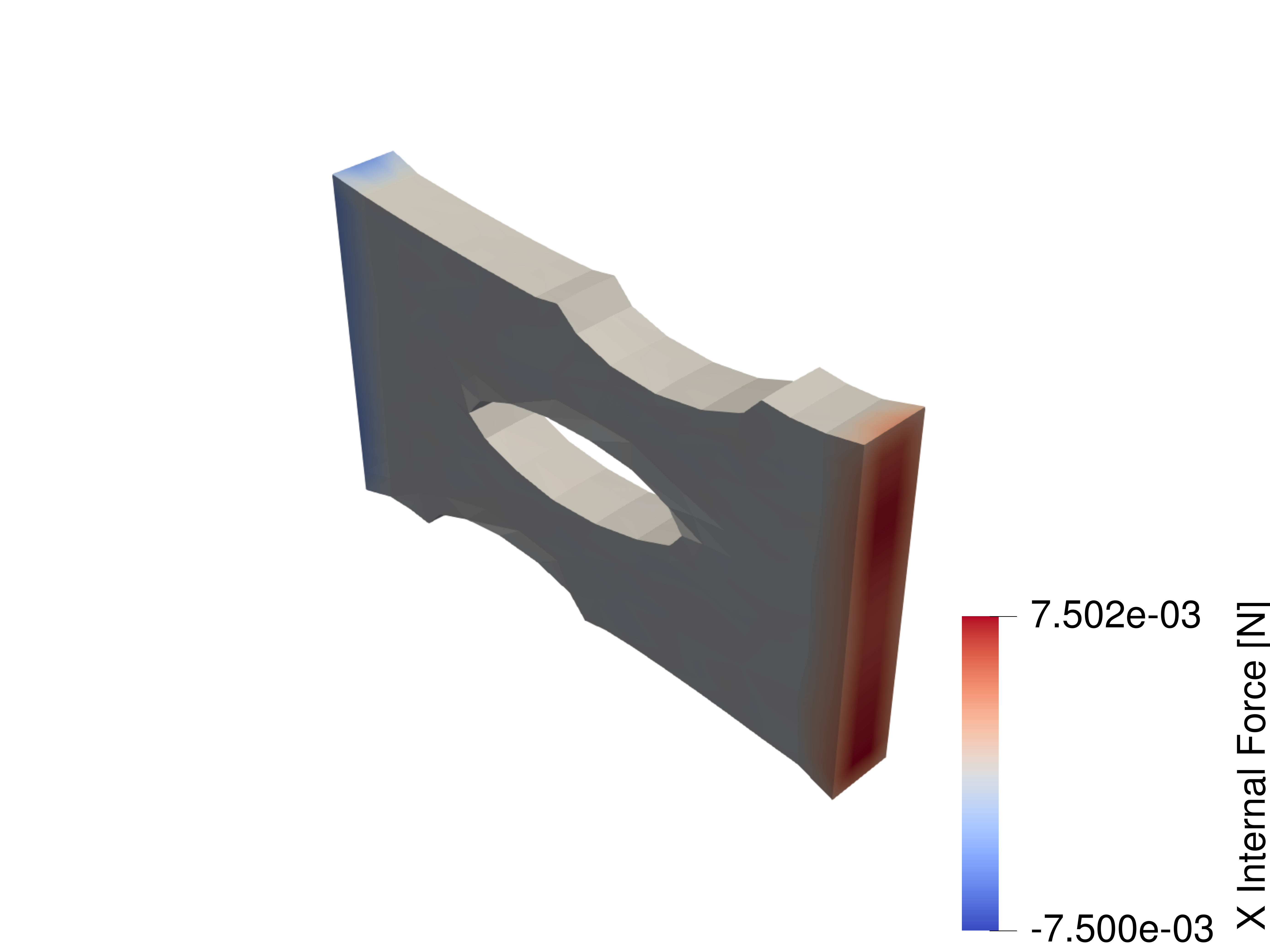}
		\caption{PINN $f_x$}
	\end{subfigure}
	\begin{subfigure}{0.3\textwidth}
		\centering
		\includegraphics[width=\textwidth]{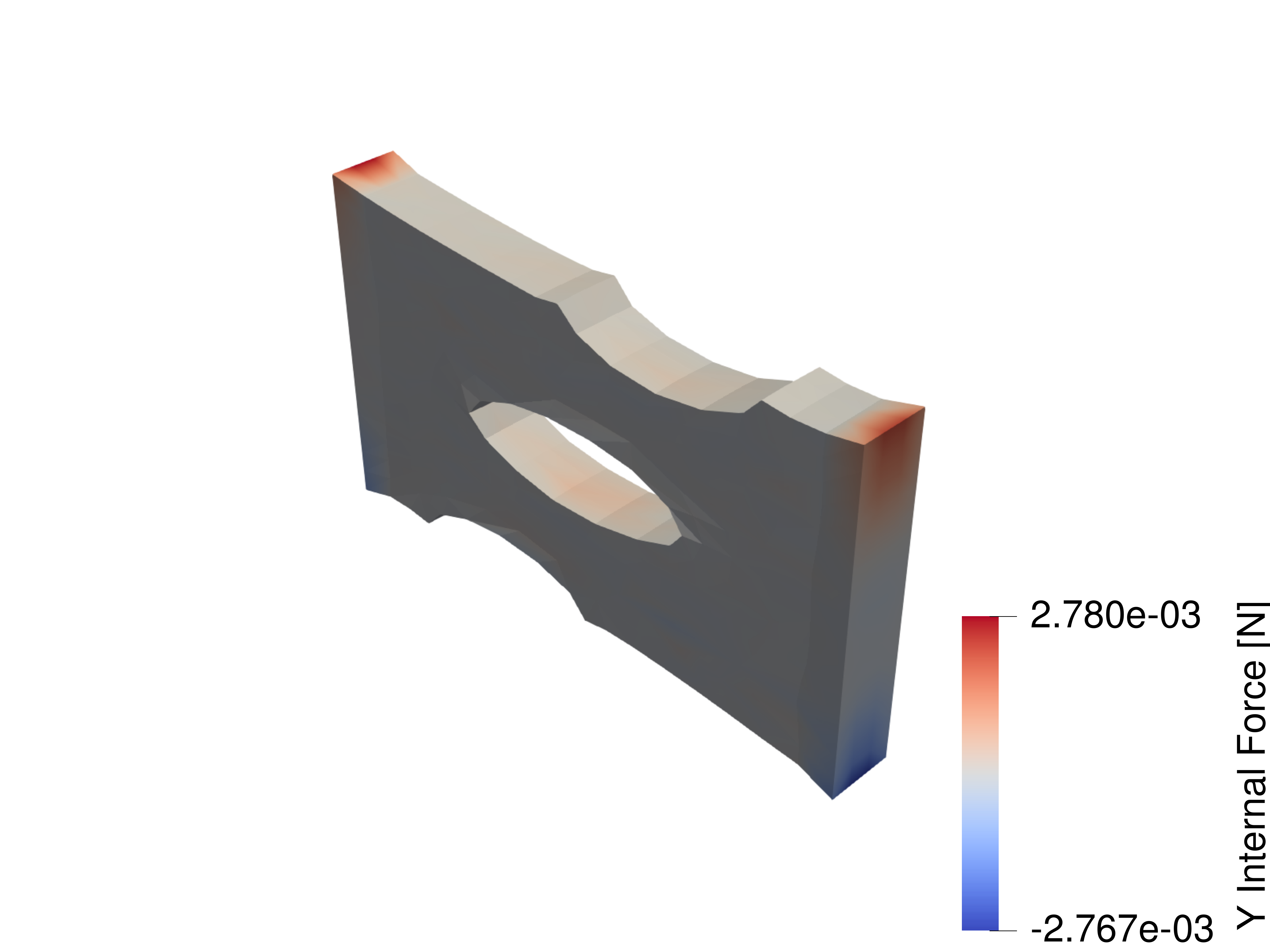}
		\caption{PINN $f_y$}
	\end{subfigure}
	\begin{subfigure}{0.3\textwidth}
		\centering
		\includegraphics[width=\textwidth]{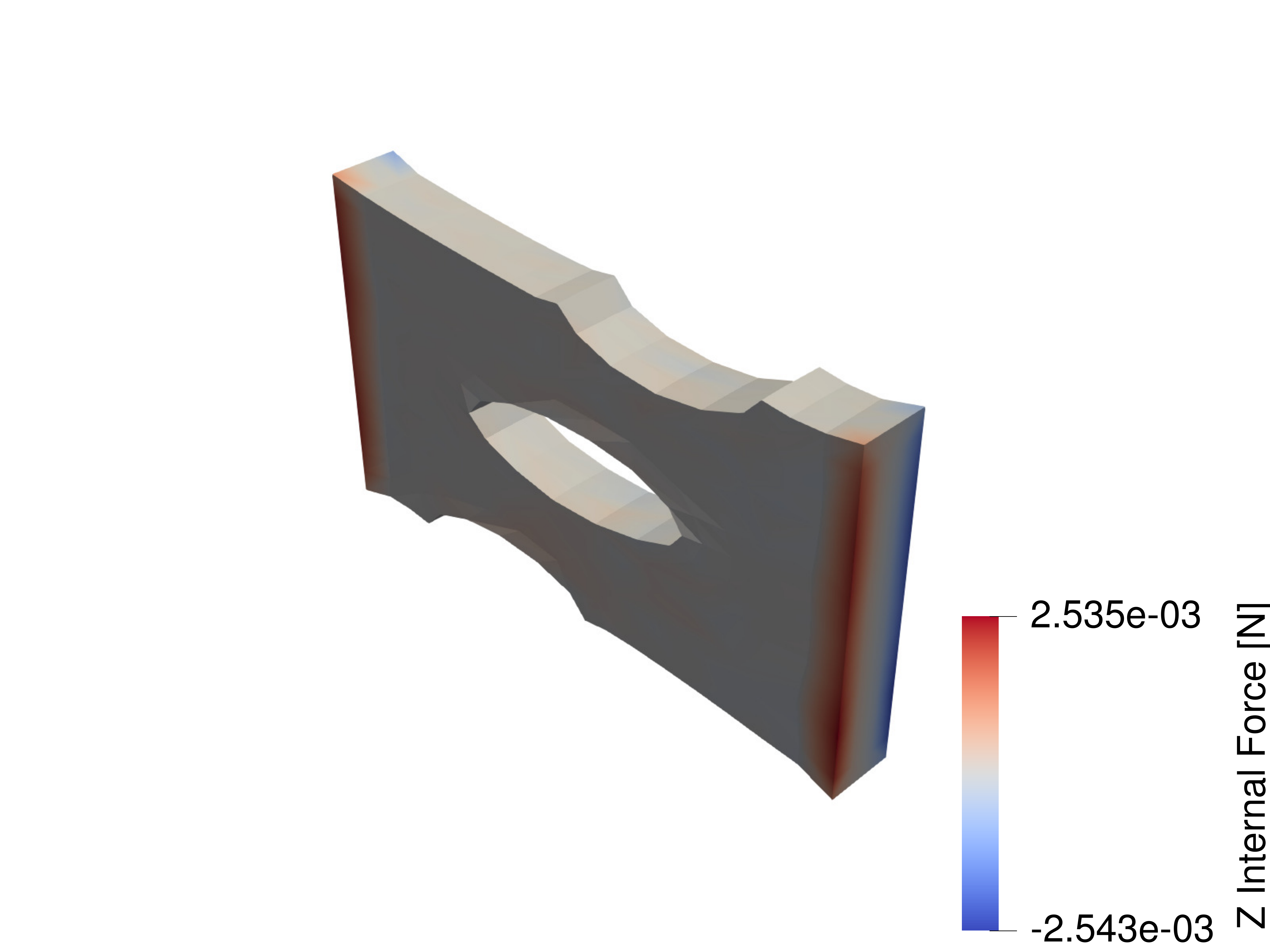}
		\caption{PINN $f_z$}
	\end{subfigure}
	\caption{Comparison of internal force field components between FEM simulations and PINN training for analogous BVPs for a Gent model with a Poisson's ratio of $\nu = 0.3$}
\label{fig:gent_poissons_ratio_0_3_internal_force}
\end{figure}

Our final benchmark problem is similar to our first but with a slightly more complicated geometry and constitutive model. Again we use a slab of material that is 1mm x 1mm x 0.2mm with a circle of radius 0.15mm punched out of the center. We further cut out two semi-circles from the top and bottom of the structure that are 0.125mm in radius and placed in a slight asymmetric fashion. The constitutive model utilized is the Gent constitutive model with a bulk modulus $K = 2.167$ MPa, a shear modulus $\mu = 1$ MPa, and a locking parameter $J_m = 1.5$, which is unitless. We utilize twenty time steps in this case to achieve a final displacement of 0.75mm or 75\% nominal strain. The geometry was meshed with 600 Hex8 finite elements which corresponds to 96,000 quadrature point evaluations per epoch of training. Eight separate PINNs were trained to show the repeatability of the method due to the stochasticity of the NN optimizer and each PINN was trained for 100,000 epochs. The evolution of the PINN residual during training and the comparison of the PINN global force-displacement to FEM are shown in Figure~\ref{fig:gent_poissons_ratio_0_3_global_results}.
Similar to the previous two benchmark problems, our method is very repeatable for this BVP in terms of the energy and residual convergence and global force-displacement prediction when compared to the FEM solution. Like the other cases, the total energy functional plateaus very quickly during training, but many more epochs are needed to properly balance the internal force vector. As before, we now plot the components of the displacement field at the maximum global strain for both the FEM solution and one of the trained PINNs after 100,000 epochs in Figures~\ref{fig:gent_poissons_ratio_0_3_displacement}.
As can be seen from the comparisons of the displacement fields, our PINN does a reasonably accurate job of predicting the full-field displacement field when compared to FEM with the largest deviation interestingly being seen in the $y$ component of the displacement. To show that our PINN solution's forces are in balance, we compare the three components of the internal force field of the FEM solve and one of the trained PINNs after 100,000 epochs at maximum global strain in Figure~\ref{fig:gent_poissons_ratio_0_3_internal_force}.
Although very small deviations of the reaction forces are seen between the trained PINN and the FEM solution, the PINN is doing a quite reasonable job of predicting the reactions forces up to three decimal places of the FEM solution. However in the case of the $y$ internal force field the PINN prediction has some locations that appear to be out of balance. This could either be due to the increased non-linearity in the materiel model or due to increased complexity in the BVP in the form of more complicated geometry and larger global strain. 

These three benchmark problems provide good evidence that our PINN does a reasonably good job for complex geometries and a wide set of different material non-linearities. This builds confidence that the PINN will properly enforce physical constraints in regions of the geometry that may not have available displacement data when solving inverse problems. A typical case for this would be the interior of the domain since DIC displacement data is available only on exterior surfaces visible to imaging systems.

\subsection{Inverse Problems} \label{sec:inverse_problems}
In this section we move towards the main focus of this paper: using our PINNs method to learn the model parameters for a constitutive model given full-field surface displacement data and global force-displacement data. To demonstrate this capability, we will consider BVPs identical to those in the previous section to generate synthetic DIC data. We do this by running an FEM problem and extracting the displacement components on a given surface (in all cases the front surface) as well as the sum of the forces in the direction of displacement on an essential boundary. To help show the ease of use of the method, we will purposely generate synthetic DIC data on meshes with a finer resolution than the computational domain used for PINN training. We do this to show the lack of a need for interpolation of the displacement data onto a computational grid as would be the case in FEMU and VFM and also to explore the effect of data dropout on our methods ability to resolve material property values accurately.

\begin{figure}
	\centering
	\begin{subfigure}{0.4\textwidth}
		\centering
		\includegraphics[width=\textwidth]{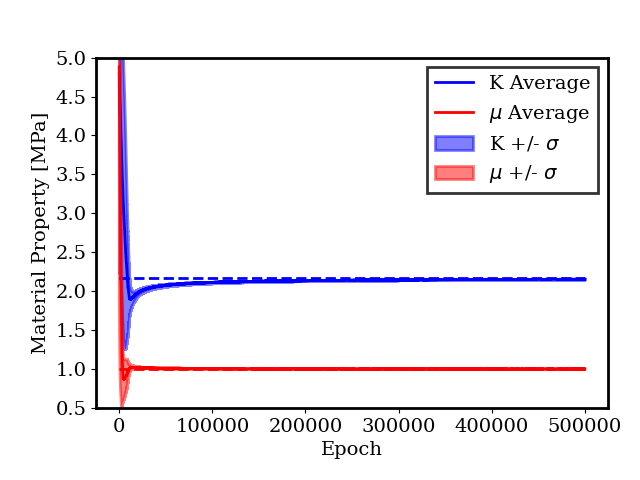}
		\caption{}
	\end{subfigure}
	\begin{subfigure}{0.4\textwidth}
		\centering
		\includegraphics[width=\textwidth]{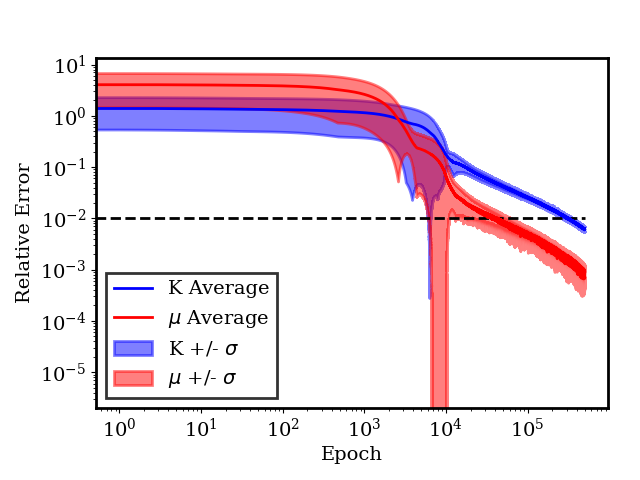}
		\caption{}
	\end{subfigure}
	\caption{Training results for an inverse problem using the Neo-Hookean constitutive model with synthetic DIC data points located at the same locations as the computational mesh nodes. (a) Material property value convergence for eight separately trained PINNs with different random initial values of material properties. (b) Material property relative error convergence for eight separately trained PINNs with different random initial values of material properties.}
\label{fig:neohookean_poissons_ratio_0_3_inverse_problem}
\end{figure}

Before we explore the effect of data dropout, we first consider a few benchmark problems for inverting the material properties. The first we will consider is analogous to the first benchmark problem considered in the previous section, but now the material parameters are unknown and added to the neural network optimizer. We use the same mesh for the generation of synthetic DIC data and the computational grid used in PINN training to serve as a benchmark for how well our method can learn material properties without deviation between the two grids. Again, we train eight separate PINNs to show the repeatability of the method due to the stochasticity of the NN optimizer. Each PINN was trained for 500,000 epochs, and material properties were initialized at random with values between 0 and 100 MPa. The convergence of the material properties for this problem can be seen in Figure~\ref{fig:neohookean_poissons_ratio_0_3_inverse_problem}.
As can be seen from the convergence plots, our method is very repeatable in terms of discovering the correct material property values to less than 1\% relative error in both the shear and bulk modulus. Overall the method finds a 5-10\% relative error solution in relatively few epochs, about 10,000. However, to get accurate values suitable for model deployment, many more epochs are needed but is still repeatable nonetheless. Interestingly the shear and bulk modulus appear to have different rates of convergence, which could be due to the sparsity of the available displacement data.

\begin{figure}
	\centering
	\begin{subfigure}{0.4\textwidth}
		\centering
		\includegraphics[width=\textwidth]{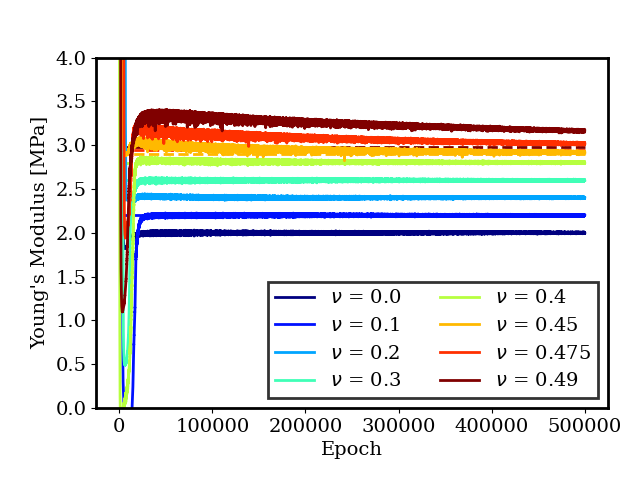}
		\caption{}
	\end{subfigure}
	\begin{subfigure}{0.4\textwidth}
		\centering
		\includegraphics[width=\textwidth]{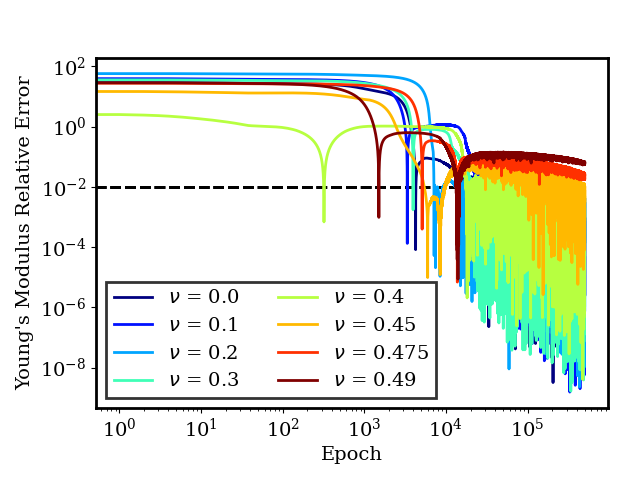}
		\caption{}
	\end{subfigure}
	\begin{subfigure}{0.4\textwidth}
		\centering
		\includegraphics[width=\textwidth]{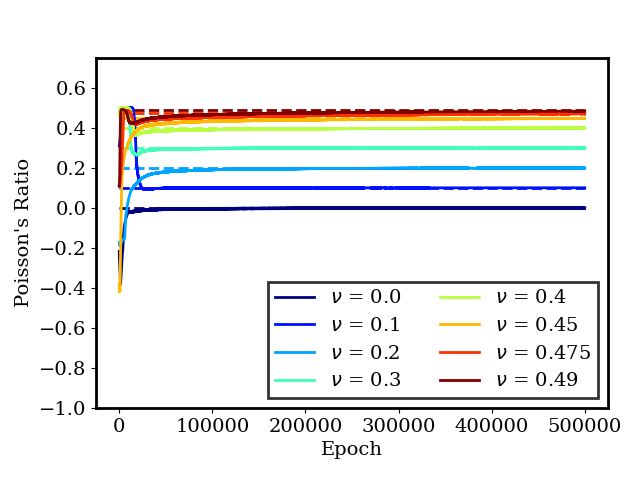}
		\caption{}
	\end{subfigure}
	\begin{subfigure}{0.4\textwidth}
		\centering
		\includegraphics[width=\textwidth]{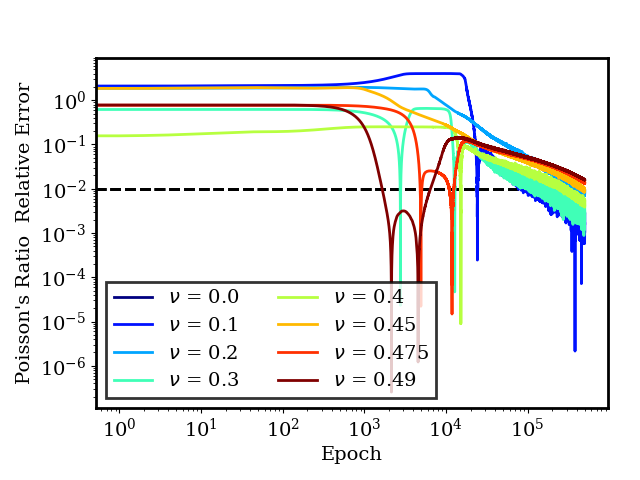}
		\caption{}
	\end{subfigure}
	\caption{Convergence of material property values for PINNs trained with synthetic DIC data generated from FEM models with different Poisson's ratios. (a) Young's modulus evolution during training. (b) Young's modulus relative error evolution during training. (c) Poisson's ratio evolution during training. (d) Poisson's ratio relative error evolution during training.}
\label{fig:neohookean_different_poissons_ratios_inverse_problem_values}
\end{figure}

We further explore our methods ability to invert the material properties for the Neo-Hookean model from synthetic DIC data by considering synthetic data that was generated from FEM models with different Poisson's ratios. The same approach for material property initialization and number of training epochs as was used in the previous inverse problem example is used here. The training results for different PINNs trained on different Poisson's ratio synthetic DIC data is shown in terms of different elastic constants in Figure~\ref{fig:neohookean_different_poissons_ratios_inverse_problem_values} where the material properties have been converted to the Young's modulus and Poisson's ratio for easier visualization.
Similar to the case of the varying Poisson's ratio forward problem benchmark, our method does quite well for obtaining the two elastic constants in this large deformation region for Poisson's ratio $\nu\le0.45$.

\begin{figure}
	\centering
	\begin{subfigure}{0.4\textwidth}
		\centering
		\includegraphics[width=\textwidth]{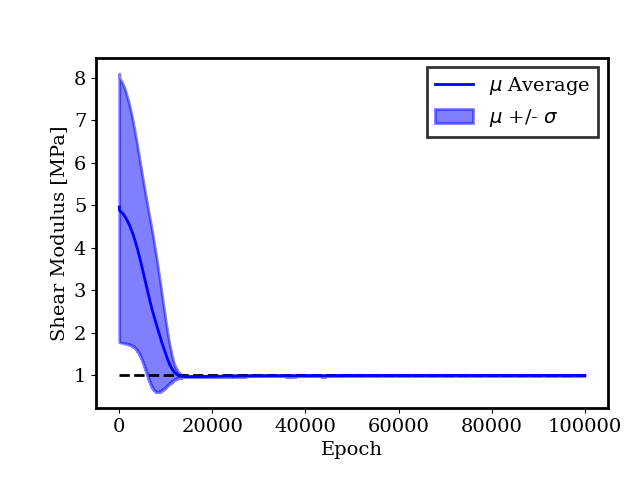}
		\caption{}
	\end{subfigure}
	\begin{subfigure}{0.4\textwidth}
		\centering
		\includegraphics[width=\textwidth]{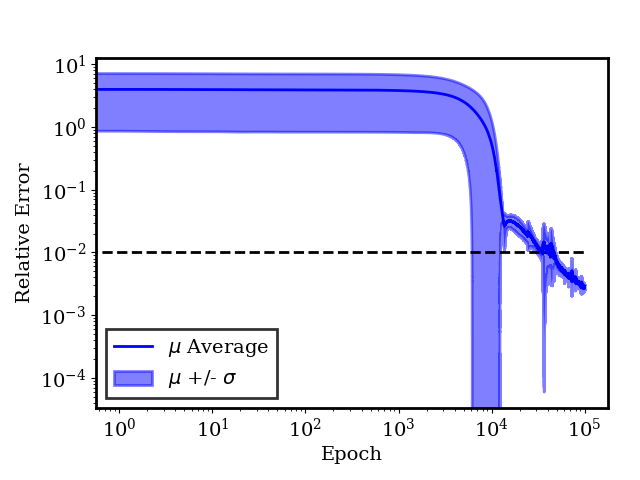}
		\caption{}
	\end{subfigure}
	\caption{Training results for an inverse problem using the Blatz-Ko constitutive model with various amounts of the synthetic DIC data. (a) Material property value for different amounts of displacement data. (b) Material property relative error for different amounts of displacement data.}
\label{fig:blatz_ko_inverse_problem}
\end{figure}

The next benchmark problem that will be considered is an analogous BVP to the Blatz-Ko forward problem considered in the previous section, but now with the shear modulus is unknown, and we supply the network with surface displacement data and global force data during training. We utilize the same mesh to generate synthetic DIC data that we use as the computational grid during PINN training to serve as a benchmark for how well our method can learn the material properties without deviation between the two grids. We train 8 separate PINNs to show the repeatability and randomly initialize the shear modulus between 0 and 100 MPa. Each of the 8 PINNs are trained for 100,000 epochs and the evolution of the material property values and the accompanying relative errors are shown in Figure~\ref{fig:blatz_ko_inverse_problem}.
From the figures we can see that the convergence of the single material property in this model is much faster than the case of the Neo-Hookean model where there two material properties to learn. Our method can calibrate this model up to 0.1\% error very efficiently and repeatably.

\begin{figure}
	\centering
	\begin{subfigure}{0.4\textwidth}
		\centering
		\includegraphics[width=\textwidth]{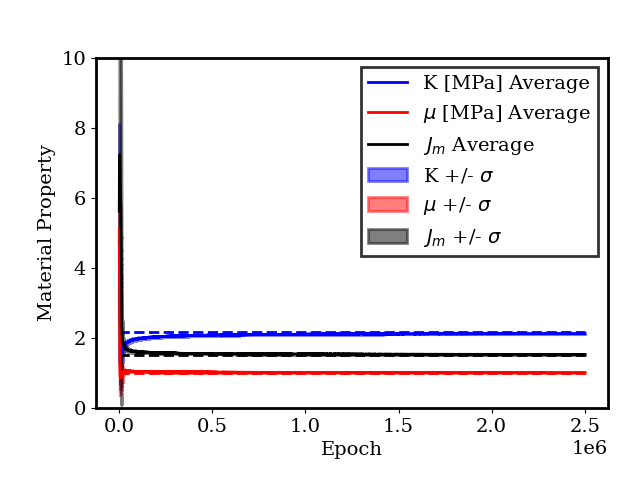}
		\caption{}
	\end{subfigure}
	\begin{subfigure}{0.4\textwidth}
		\centering
		\includegraphics[width=\textwidth]{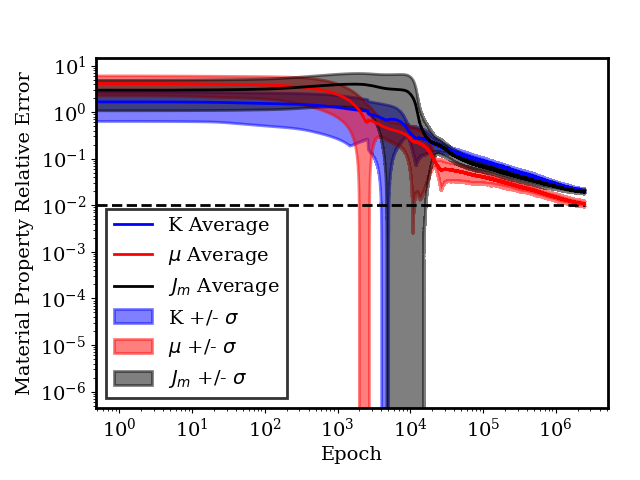}
		\caption{}
	\end{subfigure}
	\caption{Training results for an inverse problem using the Gent constitutive model with various amounts of the synthetic DIC data. (a) Material property value. (b) Material property relative error.}
\label{fig:gent_poissons_ratio_0_3_inverse_problem}
\end{figure}

Our final benchmark problem before moving on to the effect of data dropout on our method is to recover the material properties from surface displacement data and global force data from the Gent BVP in the previous section. Again, we utilize the same mesh to generate synthetic DIC data from FEM that we use as the computational mesh during PINN training to eliminate the effect of grid deviation. In this example we only train 4 separate PINNs to show repeatability due to increased computational cost of recovering the material parameters for this model and BVP. We trained each PINN for 2.5 million epochs and randomly initialize the shear and bulk modulus between 0 and 100 MPa and the locking parameters between 1 and 10. The convergence of the material property values and their accompanying relative errors are shown in Figure~\ref{fig:gent_poissons_ratio_0_3_inverse_problem}.
This inverse problem proved to be more difficult in terms of training for our PINN architecture, but the properties were still resolved to a relative error of around 2.5\%. The increased difficulty could be due to several reasons. The first could be the more complicated geometry in this BVP compared to the previous two inverse problem benchmarks. The second could be due to the large amount of displacement data due to the larger number of time steps in this problem. A third reason could be the increased material non-linearity in this particular strain energy density. Finally, our method could require longer training time with an increasing number of material properties for the model being calibrated. One way to improve this could be to use staged calibration approaches. For example, when calibrating Gent we could first calibrate the shear and bulk modulus to a Neo-Hookean model with displacement data from relatively low levels of deformation. We could then fix those parameters and then learn only the $J_m$ parameter with small and large deformation displacement data. Similar to typical constitutive model calibration techniques, there may not be a single strategy that can be applied to all constitutive models with our method. Overall the three benchmark problems show that our method can learn the material parameters for a wide set of different hyperelastic behavior with only surface displacement and global force data for complex geometries.

We now move on to the effect of data dropout and grid deviation on our method's ability to learn the material parameters. By grid deviation we mean that the grid that the synthetic DIC data was generated does not line up with the computational grid we use during PINN training. In FEMU or VFM approaches, the displacement data would typically have to be interpolated onto the computational grid. Since neural networks are naturally built for interpolation we can avoid this issue all together. Data dropout can be a concern when utilizing subset-based DIC data since that algorithm cannot measure displacements at the edges of specimens and can lose correlation in areas of large local deformation or poor speckling. To mimic both grid deviation and data dropout, we generate our synthetic DIC with a finer mesh than our computational grid. Nodal displacements on edge nodes are removed to mimic DIC's inability to resolve displacements accurately around edges. We then train 5 separate PINNs with different amounts of the synthetic data (100\%, 75\%, 50\%, 25\%, and 10\%) chosen at random for each model considered. We chose not to explore this effect on the Gent constitutive model because of the much longer run times required and only consider the Blatz-Ko and Neo-Hookean constitutive models for brevity.

\begin{figure}
	\centering
	\begin{subfigure}{0.4\textwidth}
		\centering
		\includegraphics[width=\textwidth]{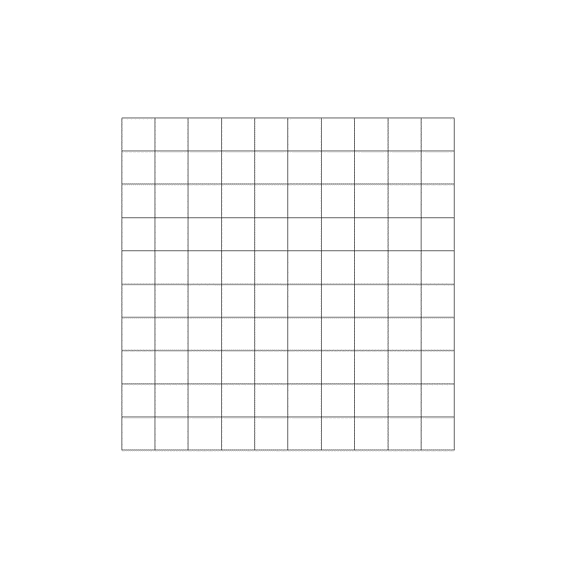}
		\caption{}
	\end{subfigure}
	\begin{subfigure}{0.4\textwidth}
		\centering
		\includegraphics[width=\textwidth]{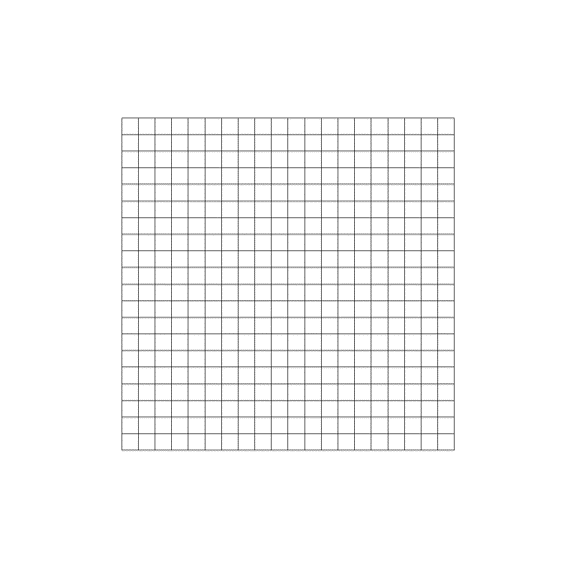}
		\caption{}
	\end{subfigure}
	\caption{Different meshes used to explore the effect of data dropout on learning the shear modulus for the Blatz-Ko constitutive model. Nodal displacements located on the exterior edges are removed from the data set. (a) Computational mesh used during PINN training. (b) Mesh used to generate DIC data with FEM.}
\label{fig:blatz_ko_data_dropout_meshes}
\end{figure}

\begin{figure}
	\centering
	\begin{subfigure}{0.4\textwidth}
		\centering
		\includegraphics[width=\textwidth]{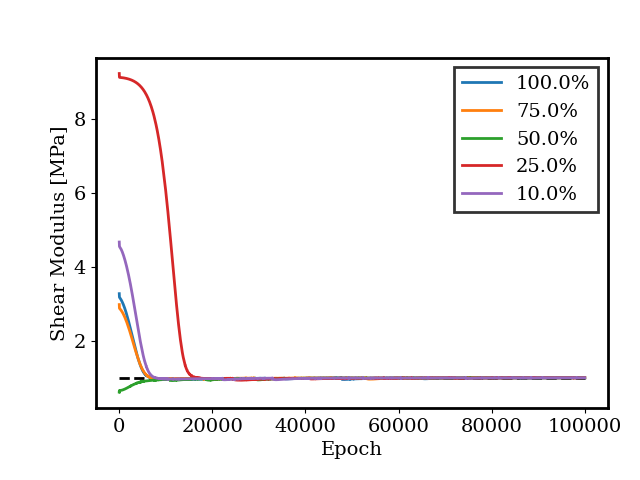}
		\caption{}
	\end{subfigure}
	\begin{subfigure}{0.4\textwidth}
		\centering
		\includegraphics[width=\textwidth]{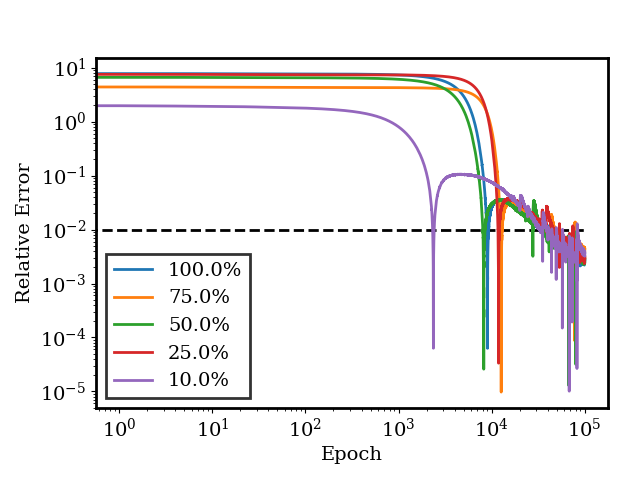}
		\caption{}
	\end{subfigure}
	\caption{Training results for an inverse problem using the Blatz-Ko constitutive model with various amounts of synthetic DIC data. (a) Material property value for different amounts of displacement data. (b) Material property relative error for different amounts of displacement data.}
\label{fig:blatz_ko_data_dropout}
\end{figure}

The first data dropout example we will consider is an inverse problem for a Blatz-Ko constitutive model using the cube compression BVP considered previously. A front view of the computational mesh and the mesh used to generate synthetic DIC data are shown in Figure~\ref{fig:blatz_ko_data_dropout_meshes}. 
For each of the 5 cases with different amounts of DIC data we train for 100,000 epochs with the shear modulus randomly initialized between 0 and 100 MPa. The evolution of the shear modulus and the relative error for the 5 different cases are shown in Figure~\ref{fig:blatz_ko_data_dropout}.
As can be seen from the figures, our method is relatively insensitive to data dropout for this particular BVP and constitutive model. For all 5 cases the shear modulus relative error is below 1\% after 100,000 epochs.

\begin{figure}
	\centering
	\begin{subfigure}{0.4\textwidth}
		\centering
		\includegraphics[width=\textwidth]{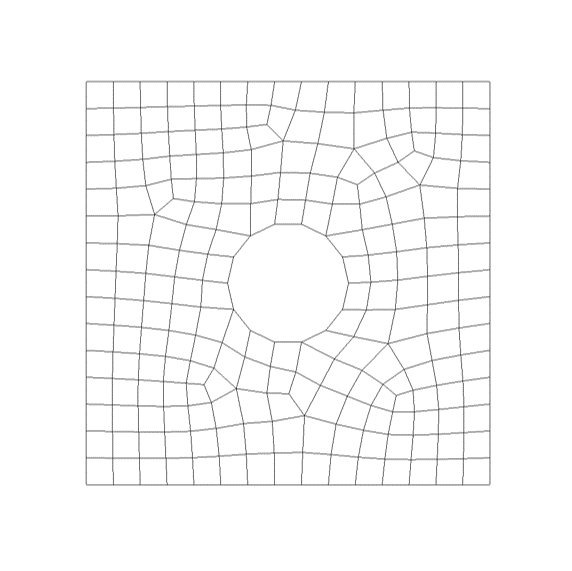}
		\caption{}
	\end{subfigure}
	\begin{subfigure}{0.4\textwidth}
		\centering
		\includegraphics[width=\textwidth]{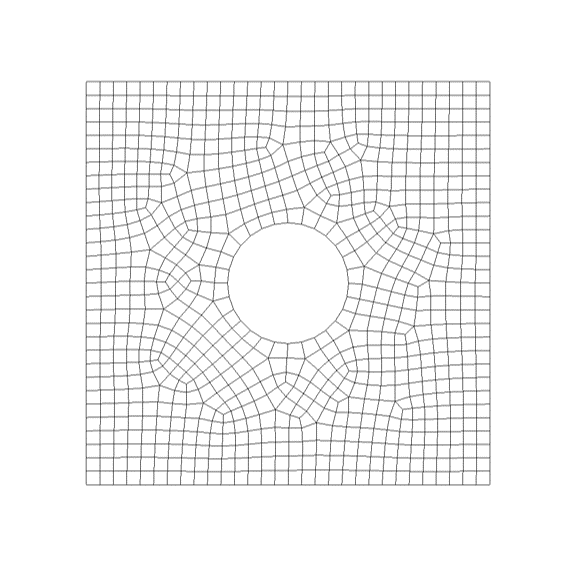}
		\caption{}
	\end{subfigure}
	\caption{Different meshes used to explore the effect of data dropout on learning the bulk and shear modulus for the Neo-Hookean constitutive model. Nodal displacement located on the exterior edges and the interior circular curve are removed from the data set. (a) Computational mesh used during PINN training. (b) Mesh used to generate DIC data with FEM.}
\label{fig:neohookean_data_dropout_meshes}
\end{figure}

\begin{figure}
	\centering
	\begin{subfigure}{0.4\textwidth}
		\centering
		\includegraphics[width=\textwidth]{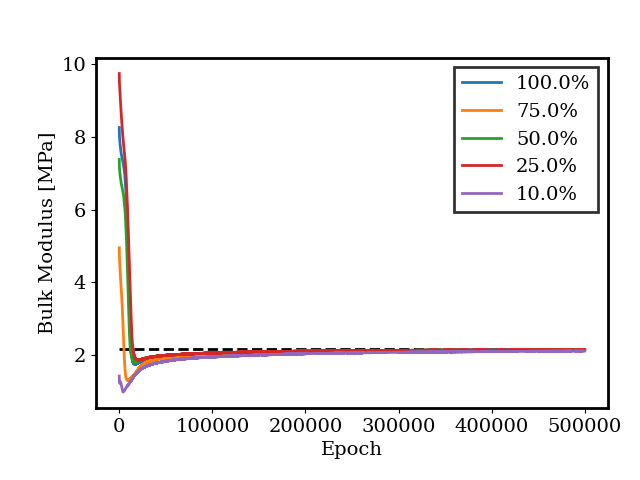}
	\end{subfigure}
	\begin{subfigure}{0.4\textwidth}
		\centering
		\includegraphics[width=\textwidth]{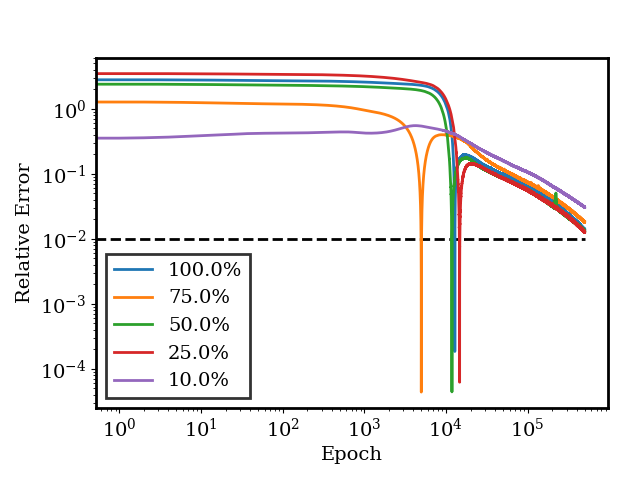}
	\end{subfigure}
	\begin{subfigure}{0.4\textwidth}
		\centering
		\includegraphics[width=\textwidth]{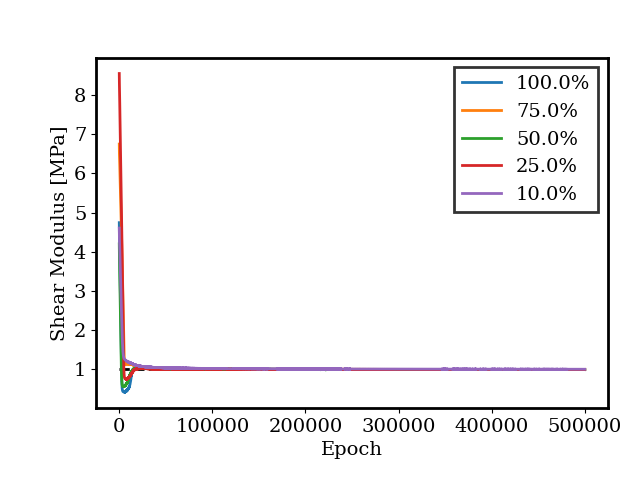}
	\end{subfigure}
	\begin{subfigure}{0.4\textwidth}
		\centering
		\includegraphics[width=\textwidth]{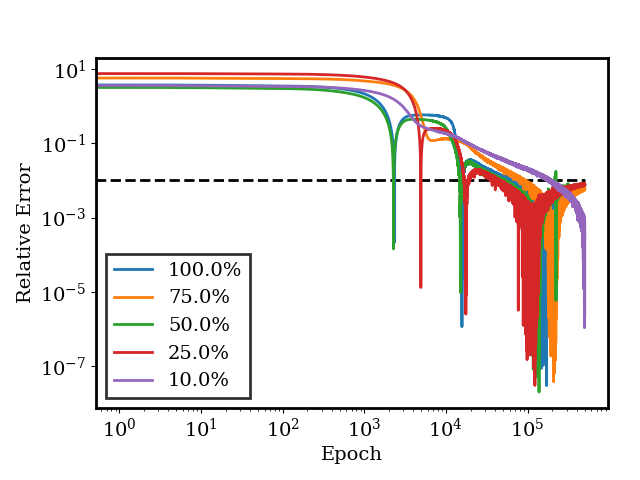}
	\end{subfigure}
	\caption{Training results for an inverse problem using the Neo-Hookean constitutive model with various amounts of synthetic DIC data. (a) Bulk modulus value evolution during training for different amounts of displacement data. (b) Bulk modulus relative error evolution during training for different amounts of displacement data. (c) Shear modulus value evolution during training for different amounts of displacement data. (d) Shear modulus relative error evolution during training for different amounts of displacement data.}
\label{fig:neohookean_data_dropout}
\end{figure}

We now move on to the harder problem of learning the two elastic constants for the Neo-Hookean model while considering the effect of data dropout. We utilize the same BVP of a slab of material with a hole punched through that center as we did for our previous examples but generate the synthetic DIC data via FEM on a finer mesh than the computational mesh used during PINN training. A front view of the meshes that we utilized for data dropout inverse problems are shown in Figure~\ref{fig:neohookean_data_dropout_meshes}. 
For each of the 5 cases with different amounts of DIC data we train for 500,000 epochs with the bulk and shear modulus initialized randomly between 0 and 100 MPa. The evolution of the shear modulus and the relative error for the 5 different cases are shown in Figure~\ref{fig:neohookean_data_dropout}.
As can be seen from the figures, our method does an excellent job of learning the shear modulus for different amounts of DIC data to below 1\% relative error. On the other hand, with decreasing amounts of available displacement data, the PINNs have a slower convergence rate in terms of the relative error of the bulk modulus. For the case where 100\% of the data is available (except for the unrealistic edge data) the bulk modulus is learned up to 1\% relative error after 500,000 epochs. For the other extreme case where only 10\% of the data is available the bulk modulus is only learned up to 5\% relative error after 500,000 epochs. For data percentages greater than 10\%, the convergence rates of the bulk modulus are almost indistinguishable showing that the method is relatively insensitive to reasonable amounts of data dropout. For the Neo-Hookean inverse problem benchmark shown previously in Figure~\ref{fig:neohookean_poissons_ratio_0_3_inverse_problem}, the bulk modulus was learned to well below 1\% relative error after 500,000 epochs. These two sets of results are potentially pointing to a slight increase in training convergence rate due to grid deviation and a more profound effect of training convergence rate on material properties due to data dropout. 

\section{Conclusion} \label{sec:conclusion}
In this paper a new approach for solving inverse problems in solid mechanics with PINNs was developed with an emphasis on formulating the problem in terms of what experimental data can actually be measured using existing trusted paradigms such as DIC. Apart from inverse problems, we showed that the new forward problem formulation was relatively robust and insensitive to geometric complexity and constitutive model non-linearities in terms of displacement predictions and force balance. We showed that our method was able to use only surface displacement data, that does not necessarily correspond with the computational mesh nodal locations, and global force-displacement data to calibrate constitutive models of increasing material non-linearities on geometries of expanding complexity, which is a key novelty to this work. The method also proved to be relatively insensitive to reasonable amounts of displacement data dropout, which can be a concern when using any method with real experimental DIC data. Although the method has shown great success for these hyperelastic constitutive models, more work is needed to expand the method into the more interesting and relevant domain of history-dependent constitutive models such as plasticity and viscoelasticity and to test the method on actual experimental data. We view this work as a stepping stone towards these applications.

\section*{Acknowledgments}
This work was supported by the Advanced Simulation and Computing (ASC) program at Sandia National Laboratories through investments in the Advanced Machine Learning Initiative (AMLI). This work utilized the institutionally funded Synapse resources provided by Sandia's Common Engineering Environment (CEE) in department 9321. These resources are greatly appreciated. Sandia National Laboratories is  a multi-mission laboratory managed and operated by National Technology \& Engineering Solutions of Sandia, LLC, a wholly owned subsidiary of Honeywell International Inc., for the U.S. Department of Energy's National Nuclear Security Administration under contract DE-NA0003525.

\bibliographystyle{unsrtnat}
\bibliography{bibliography_v2}  






\end{document}